\documentclass[letterpaper]{article} 
\usepackage[]{aaai2026}  
\usepackage{times}  
\usepackage{helvet}  
\usepackage{courier}  
\usepackage[hyphens]{url}  
\usepackage{graphicx} 
\usepackage{natbib}  
\usepackage{caption} 
\usepackage{hyperref}
\usepackage{algorithm}
\usepackage{algorithmic}

\usepackage{graphicx}
\usepackage{amsmath}
\usepackage{xcolor}
\usepackage{amsthm}
\usepackage{amssymb}
\usepackage{pifont}
\usepackage{xspace}
\usepackage{misc} 
\usepackage{thm-restate}
\usepackage{tikz}
\usepackage{microtype}
\usepackage{makecell}

\newtheorem{definition}{Definition}
\newtheorem{corollary}{Corollary}
\newtheorem{theorem}{Theorem}
\newtheorem{lemma}{Lemma}

\usepackage{acro}
\newcommand{\acro}[3]{\DeclareAcronym{#1}{short = {#2}, long= {#3}}}
\acro{gnn}{GNN}{graph neural network}
\acro{afml}{AFML}{alternation free modal logic}
\acro{rml}{RML}{ratio modal logic}
\acro{gml}{GML}{graded modal logic}
\acro{fo}{FO}{first-order logic}
\acro{ml}{ML}{modal logic}
\acro{ef}{EF}{Ehrenfeucht-Fra\"iss\'e}
\acro{mso}{MSO}{monadic second order logic}

\usepackage{colonequals}
\usepackage{enumitem}
\usepackage{mathtools}

\newcommand{\meanGNN}{\text{\textnormal{Mean-GNN}}\xspace}
\newcommand{\conmeanGNN}{\text{\textnormal{Mean$^c$-GNN}}\xspace}
\newcommand{\contresmeanGNN}{\text{\textnormal{Mean$^{c,t}$-GNN}}\xspace}

\newcommand{\simple}[1]{\text{\textnormal{simple} #1}\xspace}

\newcommand{\simpleGNN}{\simple{\meanGNN}\xspace}

\newcommand{\simplecontresGNN}{\simple{\contresmeanGNN}\xspace}
\newcommand{\sumGNN}{\text{\textnormal{Sum-GNN}}\xspace}
\newcommand{\maxGNN}{\text{\textnormal{Max-GNN}}\xspace}

\newcommand{\AFML}{\text{\textnormal{AFML}}\xspace}
\newcommand{\RML}{\text{\textnormal{RML}}\xspace}
\newcommand{\GML}{\text{\textnormal{GML}}\xspace}
\newcommand{\MoL}{\text{\textnormal{ML}}\xspace}

\newcommand{\MSO}{\text{\textnormal{MSO}}\xspace}
\newcommand{\AGG}{\mn{AGG}}
\newcommand{\COM}{\mn{COM}}
\newcommand{\CLS}{\mn{CLS}}
\newcommand{\APR}{\mn{APR}}
\newcommand{\MEAN}{\mn{MEAN}}
\newcommand{\SUM}{\mn{SUM}}
\newcommand{\MAX}{\mn{MAX}}
\newcommand{\RELU}{\mn{ReLU}}
\newcommand{\trRELU}{\mn{ReLU}^*}
\newcommand{\Unr}{\mn{Unr}}
\newcommand{\ws}{\mn{ws}}

\newcommand{\aff}{\mn{aff}}
\newcommand{\ri}{\mn{ri}}
\newcommand{\clo}{\mn{cl}}
\newcommand{\cvh}{\mn{cvh}}

\newcommand{\EF}[4]{
    \def\param{#1}%
    \ifx\param\empty
    \mathcal{E}_{#2}^{#3}(#4)%
    \else
    \mathcal{E}_{#1,#2}^{#3}(#4)%
    \fi
}

\title{Logical Characterizations of  GNNs with Mean Aggregation}
\author{
    Moritz Schönherr,
    Carsten Lutz
}
\affiliations{
    Department of Computer Science, Leipzig University, Germany\\
    Center for Scalable Data Analytics and Artificial Intelligence (ScaDS.AI), Dresden/Leipzig, Germany\\
    \{schoenherr, clu\}@informatik.uni-leipzig.de

}

\begin{document}

\maketitle

\begin{abstract}
  We study the expressive power of graph neural networks (GNNs) with
  mean as the aggregation function, with the following results. In the non-uniform setting, such GNNs have exactly the same expressive power as ratio modal
  logic, which has modal operators expressing that at least a
  certain
  ratio of the successors of a vertex satisfies a specified property.
  In the uniform setting, the expressive power relative to MSO is exactly that of modal logic, and thus identical to the (absolute) expressive power of GNNs with max aggregation. The proof, however, depends on constructions that are not satisfactory from a practical perspective. This leads us to making the natural assumptions that combination functions are continuous and classification functions are thresholds.
  The resulting class of GNNs with mean 
  aggregation turns out to be much less 
  expressive: relative to MSO and in the
  uniform setting, it has the same expressive power as alternation-free modal logic. This is in contrast to the expressive power of GNNs with max and sum 
  aggregation, which is not affected by these assumptions. 
\end{abstract}

%

\section{Introduction}





    

Graph neural networks (GNNs) are a family of deep-learning
architectures that act directly on graphs, removing the need for prior
serialization or encoding, in this way ensuring
isomorphism invariance \cite{Scarselli2009,Wu2021,Zhou2020}.  GNNs
have been applied successfully in domains ranging from molecular
property prediction and drug discovery \cite{Bongini2021} to fraud
detection \cite{Deng2021} and e-commerce recommendation \cite{Wu2022},
traffic forecasting \cite{Jiang2022}, and physics simulation
\cite{Shlomi2020}. Numerous GNN variants exist, all sharing central
features such as message passing and the iterative update of vertex
embeddings, and an important
 foundational question is which 
 properties a given GNN architecture can represent.
 Beginning with \citet{Barcelo2020}
 and \citet{DBLP:conf/lics/Grohe21}, an expanding body of work has
 addressed this question by associating the expressive power of GNNs
 with that of various logical formalisms. Such logical characterizations may
 be used to guide the choice of a GNN model for a specific task at hand
 and to reveal potential gaps between a model’s representational capacity and the task’s requirements.

\citet{Barcelo2020} focus on GNNs 
with constant iteration depth, sum as the aggregation function, and truncated ReLU as the activation function. One main result is that, relative to first-order logic (FO),
the expressive power of GNNs as a vertex classifier is exactly that of graded modal logic (GML). 
In other articles,
such as \cite{TenaCucala2023,TenaCucala2024}, sum aggregation is replaced with max aggregation, and GML is replaced with certain restrictions of non-recursive datalog. The main purpose of the
current paper is to characterize the  expressive power of GNNs as a vertex classifier when the aggregation function is arithmetic mean. This is in fact a very natural and important case. Influential graph learning systems such as GraphSAGE
\cite{Hamilton2017} and Pinterest's web-scale recommender PinSAGE \cite{Ying2018}
use mean aggregation, and  the popular Graph Convolutional Networks (GCNs) use a weighted version of mean \cite{Kipf2017}.
Moreover, mean aggregation is  amenable to 
random sampling and thus a good choice for graphs in which vertices may have very large degree.  Our characterizations are in terms of suitable versions of modal logic. We also provide several new observations regarding GNNs with 
max and sum aggregation. 

\begin{table}[t!]
\centering
\begin{tabular}{|c||c|c|c|}
\hline
 & \makecell{Non-\\Uniform} & \makecell{Uniform\\wrt. $\text{MSO}$ }& \makecell{Uniform\\absolute} \\
\hline
\hline
Mean$^{c,t}$ & RML {\scriptsize Th. \ref{lemma: rml-to-gnn-translation-nonuniform}, \ref{theorem: rml-to-gnn-translation-nonuniform-univ-approx}, \ref{lemma: mean-gnn-to-rml-non-uniform-translation}}& AFML {\scriptsize Th. \ref{lem:conmeangnnsubseteqafml}, \ref{lemma: afml-to-cmgnn-translation}, \ref{thm:sfjlherogahfga}} &   $>$ \text{AFML} {\scriptsize Th. \ref{thm:summarymeanabsoluteuniform}} \\
\hline
Mean & RML {\scriptsize Th. \ref{lemma: rml-to-gnn-translation-nonuniform}, \ref{theorem: rml-to-gnn-translation-nonuniform-univ-approx}, \ref{lemma: mean-gnn-to-rml-non-uniform-translation}}& ML {\scriptsize Cor. \ref{corollary: meangnn-to-ml-translation}, Th. \ref{lemma: rml-to-gnn-translation}, \ref{thm:crazyclass}}&  $>$ \text{ML} {\scriptsize Th. \ref{thm:summarymeanabsoluteuniform}}\\
\hline
Sum & GML {\scriptsize Th. \ref{theorem: rml-to-gnn-translation-nonuniform-univ-approx}, \ref{lemma: mean-gnn-to-rml-non-uniform-translation}}& GML$^\dagger$ & $>$ \text{GML}$^\ddagger$ \\
\hline
Max & ML {\scriptsize Th. \ref{lem:MLtomaxGNN}, Th.~\ref{lem:frommaxgnntoML}}& ML {\scriptsize Th. \ref{lem:MLtomaxGNN}, Th.~\ref{lem:frommaxgnntoML}}& ML {\scriptsize Th. \ref{lem:MLtomaxGNN}, Th.~\ref{lem:frommaxgnntoML}}\\
\hline
\end{tabular}
\caption{Overview of results, with $\cdot^\dagger$ from
\cite{Barcelo2020}
and $\cdot^\ddagger$ from \cite{DBLP:conf/icalp/BenediktLMT24}.}
\label{tab:overview}
\end{table}
\newcommand{\markDiamond}{%
  \mathbin{%
    \ooalign{%
      $\Diamond$\cr \hidewidth$\cdot$\hidewidth }%
  }%
} We consider the expressive power of GNNs in two different settings
that have both received attention in the literature.  In the uniform
setting, a GNN model \Mmc has the same expressive power as a logic
\Lmc if \Mmc and \Lmc can express exactly the same vertex classifiers,
across all graphs.  In the non-uniform setting, we only require that
for every $n \geq 1$, \Mmc and \Lmc can express exactly the same
vertex classifiers across all graphs with $n$ vertices.  In the
uniform case, we follow \citet{Barcelo2020} in studying the expressive
power relative to FO, that is, we restrict our attention to GNNs that
express a vertex property definable by an FO formula. In fact
we use monadic second-order logic (MSO) in place of FO, based on the observation from
\cite{Ahvonen2024} that every GNN that expresses an MSO-definable
property actually expresses an FO-definable property.  In the
non-uniform case, we seek absolute characterizations that are
independent of any background logic.

In the non-uniform case, we prove that
GNNs with mean aggregation have the same
expressive power as ratio modal logic (RML) which provides modal operators 
$\markDiamond^{\geq r} \varphi$ and $\markDiamond^{>r} \varphi$ expressing
that the fraction of successors that satisfy $\varphi$ is at least $r$ (resp.\ exceeds $r$). This should be contrasted with GNNs
based on sum aggregation and max aggregation,
which have the same expressive power as 
GML and modal logic (ML), respectively.
While these latter results are not surprising given existing work, we are not aware that they have
been proved anywhere in this form, and we
provide proof details here. 
All results are summarized in Table~\ref{tab:overview}. The expressive equivalence between \maxGNN{}s and ML even
holds in the uniform case.
In the non-uniform setting, GML is strictly more
expressive than RML, which in turn is strictly more expressive than
ML. Our results thus reflect the known fact
that \sumGNN{}s are (non-uniformly) strictly more expressive than \meanGNN{}s, which are in turn strictly more expressive than \maxGNN{}s \cite{DBLP:conf/iclr/XuHLJ19}.

The uniform setting turns out to be significantly more subtle and interesting.
The fragment of RML that is expressible
in MSO is exactly~ML. Given our results
from the non-uniform case, one may thus expect
that in the uniform case and relative
to MSO, \meanGNN{}s have the same expressive
power as ML. We prove that this is indeed the case. It follows that 
(uniformly and relative to MSO) \meanGNN{}s have the same expressive power as \maxGNN{}s, but are strictly less expressive than \sumGNN{}s. However,
the translation of ML formulas into \meanGNN{}s 
is unsatisfactory from a practical standpoint: one may either
use a combination function that is not differentiable, and in fact not
even continuous, or a rather unnatural
classification function (in our translation, the rational
numbers are classified as 0 and the irrational numbers as 1).
This leads us to study \meanGNN{}s under
the natural assumptions that 
(i)~the combination functions are
continuous and (ii)~the classification function is a threshold function. This case is denoted Mean$^{c,t}$ in Table~\ref{tab:overview}. 

We prove that
assumptions~(i) and~(ii) result in a
significant drop of expressive power:
relative to MSO, GNNs with mean aggregation now have the
same expressive power as \emph{alternation-free} modal logic (AFML) in which modal diamonds and boxes cannot be mixed. We believe that from a practical perspective, AFML provides a more
 realistic characterization of the expressive power of \meanGNN{}s than ML (relative to MSO). 
It is also interesting to note that assumptions~(i) and~(ii) have no impact on expressive power in the cases of sum aggregation
and max aggregation.

Throughout the paper, we also consider a `simple' version of GNNs in which the combination function is 
a feedforward neural network without hidden layers. We then pay special attention to the activation function.
Notably, we prove that the
non-uniform results for mean and sum
aggregation stated above hold for  all continuous non-polynomial activation functions. We also prove that our result
relating \meanGNN{}s and AFML in the uniform
case also holds for ReLU, truncated ReLU, and sigmoid activation. This is in contrast to \sumGNN{}s where transitioning from (truncated) ReLU to sigmoid reduces the expressive power  in the uniform case \cite{DBLP:conf/iclr/KhalifeT25}.

\smallskip

This is the extended version of \cite{schoenherr2025AAAI}.

\paragraph{Related Work.} We start with the
uniform setting. The link between GNNs and modal logic was  established in
\cite{Barcelo2020}, relative to FO.
The expressive power of {\sumGNN}s beyond FO was studied in
\cite{DBLP:conf/icalp/BenediktLMT24}, and linked
to modal logic with Presburger quantifiers. 
In \cite{TenaCucala2024},
{\maxGNN}s are translated into non-recursive datalog programs with negation-as-failure that adhere to a certain tree-shape. This formalism is closely related to modal logic.
Recently, GNNs with transformer layers have been characterized
in terms of modal logic \cite{AAAITransformers}.
Such layers are closely related to mean aggregation. Sound logical explanations for 
a certain kind of monotone \meanGNN
have been studied in \cite{MorrisHorrocks25}.
In the non-uniform setting, \citet{DBLP:journals/theoretics/Grohe24} 
shows that GNNs have the same expressive power
as an extension of GFO+C, the guarded fragment of FO with counting capabilities, with built-in relations.
Some related results are in \cite{DBLP:conf/lics/GroheR24}. Without reference
to logic, the expressive power of different aggregation functions is studied in \cite{DBLP:conf/iclr/XuHLJ19,Rosenbluth2023}.

\section{Preliminaries}
\label{sect:prelims}

We use $\mathbb{N}$ and $\mathbb{N}^+$ to denote the
set of all non-negative and positive integers, respectively. For $n \geq 1$, we write $[n]$ for the set $\{1,\dots,n\}$. With $\Mmc(X)$ 
we mean the set of all finite multisets over the set $X$, that is, the set of functions $X \to \mathbb{N}$ where all but finitely many elements of $X$ are mapped to~$0$. 
For a vector $\bar{x}\in\mathbb{R}^{\delta}$ we use $x_1,\ldots,x_{\delta}$ 
or, when more readable, $(\bar x)_1,\ldots,(\bar x)_{\delta}$ 
to refer to its
components. 


\paragraph{Graph Neural Networks.}
Let $\Pi = \{P_1,\ldots, P_n\}$ be a finite set of \emph{vertex labels}.
A \emph{($\Pi$-labeled directed) graph} is a tuple $G=(V,E, \pi)$ that consists of a set of vertices $V$, a set of edges $E\subseteq V\times V$, and a vertex labeling function $\pi: V\to 2^\Pi$.
Unless noted otherwise, all graphs in this article are finite.
For easier reference, we may write $V^G$ for $V$,
$E^G$ for $E$, and $\pi^G$ for $\pi$.
The \emph{neighborhood} of a vertex $v$ in a graph $G$ is the set of its successors, formally $\mathcal{N}(G,v) = \{u\mid (v,u)\in E\}$.
A \emph{pointed graph} is a pair $(G,v)$
with $G$ a graph and $v \in V^G$ a distinguished  vertex. A \emph{(vertex) property} is  a class of pointed graphs, over a common set of labels $\Pi$, that is closed under isomorphism. 


%
A \emph{\ac{gnn}} on $\Pi$-labeled graphs is a tuple
$$\mathcal{G}=(L,\{\AGG^{\ell}\}_{\ell \in [L]},\{\COM^{\ell}\}_{\ell \in [L]},\CLS)$$
where $L \geq 1$ is the number of layers and
for each $\ell \in [L]$,
$\AGG^{\ell}:\ \mathcal{M}(\mathbb{R}^{\delta^{\ell-1}})\to \mathbb{R}^{\delta^{\ell-1}}$
is an
\emph{aggregation function}, $\COM^{\ell}:\ \mathbb{R}^{\delta^{\ell-1}}\times \mathbb{R}^{\delta^{\ell-1}}\to \mathbb{R}^{\delta^{\ell}}$ is a
\emph{combination function}, and $\CLS:\ \mathbb{R}^{\delta^L} \to \{0,1\}$ is a \emph{classification function}. We call $\delta^{\ell-1}$
the \emph{input dimension} of layer~$\ell$ and $\delta^\ell$
the
 \emph{output dimension}, with 
 the input dimension $\delta^0$
 of Layer~1 being at least $|\Pi|$.
Typical aggregation functions are sum, max, and mean,  applied
component-wise. We only consider GNNs in which every layer uses the
same aggregation function. If we want to highlight the aggregation
function, we may
speak of a \sumGNN, a \maxGNN, or a \meanGNN.

We now make precise the semantics of GNNs. 
For $1\leq \ell\leq L$, the $\ell$-th layer assigns to each $\Pi$-labeled  graph  $G$ and vertex $v \in V^G$ a feature vector \mbox{$\bar{x}^{\ell}_{G,v} \in \mathbb{R}^{\delta^\ell}$}. 
%
The \emph{initial feature vector} 
$\bar{x}^{0}_{G,v} \in \mathbb{R}^{\delta^0}$ of vertex $v$ 
in graph $G$ is defined as follows: for all $i \in [|\Pi|]$, the $i$-th
value of $\bar{x}^{0}_{G,v}$ is $1$ if $P_i\in\pi^G(v)$, and all other values are $0$.
%
%
%
%
The feature vector $\bar{x}^{\ell}_{G,v}$ assigned by the $\ell$-th layer  is 
%
\begin{equation*}
\bar{x}^{\ell}_{G,v} = \COM^{\ell}(\bar{x}^{\ell-1}_{G,v}, \AGG^{\ell}(\{\!\{\bar{x}^{\ell-1}_{G,u}\mid u\in\mathcal{N}(v)\}\!\}))
\tag{$*$}
\end{equation*}
where $\{\!\{\cdots\!\,\}\!\}$ denotes multiset; we define the mean of the empty multiset to be~0.
  We write $\bar{x}^\ell_v$ instead of $\bar{x}^\ell_{G,v}$ if the graph $G$ is clear from the context.
%
%
The output of the last layer is then  passed to the classification
function.
%
The property \emph{defined} by a GNN $\mathcal{G}$ is the set 
of pointed $\Pi$-labeled graphs $$\{(G,v)\mid \CLS(\bar{x}_{G,v}^{L}) = 1\}.$$
We also say that \Gmc \emph{accepts} the graphs in this set.

%
%

In a \emph{simple} \ac{gnn}, as considered
for example in \cite{Barcelo2020,Ahvonen2024,TenaCucala2023},  the combination function is restricted to the form
$$\COM(\bar{x}_v,\bar{x}_a) = f(\bar{x}_v\cdot C + \bar{x}_a\cdot A +\bar{b}),$$
where
$f: \mathbb{R} \rightarrow \mathbb{R}$ is a (typically non-linear) \emph{activation function} applied component-wise, $A,C \in \mathbb{R}^{\delta^{\ell-1}\times \delta^{\ell}}$ are matrices,  $\bar{b} \in \mathbb{R}^{\delta^\ell}$ is a bias vector, and $\bar{x}_a$ is the result of the aggregation function,
applied as in ($*$). Relevant choices for the activation function $f$ include the \emph{truncated ReLU} $\trRELU(x) = \min(\max(0,x),1)$,  the \emph{ReLU} $\RELU(x) = \max(0,x)$
and the \emph{sigmoid function}  $\sigma(x) = \frac{1}{1+e^{-x}}$.
The following is an easy consequence of the observation that
$\trRELU(x) = \RELU(x) - \RELU(x-1)$ for all $x \in \mathbb{R}$.
\begin{restatable}{lemma}{lemtrRELUtoRELU}
\label{lem:trRELUtoRELU}
   Let $\AGG \in \{ \mn{Max},\mn{Sum},\mn{Mean}\}$.
   If a property is definable by a \simple{$\AGG$-\ac{gnn}}
   with $\trRELU$ activation, then it is definable
   by a \simple{$\AGG$-\ac{gnn}} with $\RELU$ activation.
\end{restatable}

\paragraph*{Modal Logic}

Formulas of \emph{modal logic (ML)} over a set of vertex labels $\Pi$ are defined by
the grammar rule
$$\varphi \coloncolonequals P\mid \lnot \varphi\mid 
\varphi\lor\varphi\mid \Diamond\varphi,
$$
where $P$ ranges over $\Pi$ \cite{DBLP:books/cu/BlackburnRV01}. As usual, we use $\varphi \wedge \psi$ as abbreviation for $\neg(\neg \varphi
\vee \neg \psi)$, $\Box \varphi$ as abbreviation for
$\neg \Diamond \neg \varphi$, $\top$ as abbreviation
for $P \vee \neg P$ with $P \in \Pi$ chosen arbitrarily, and $\bot$ as abbreviation for $\neg \top$.
Satisfaction of a formula $\varphi$ by a vertex $v$
in a $\Pi$-labeled graph $G$ is defined inductively as follows:
$$
  \begin{array}{rcll}
    G,v &\models& P & \text{if } P \in \pi(v) \\[1mm]
    G,v &\models& \neg \varphi &\text{if not }G,v\models\varphi\\[1mm] 
    G,v &\models& \varphi \vee \psi & \text{if }
 G,v \models \varphi 
    \text{ or } G,v \models \psi \\[1mm]
    G,v &\models& \Diamond \varphi & \text{if }
     G,u \models \varphi \text{ for some } u \in \Nmc(G,v).
  \end{array}
$$

%
%

\emph{\Ac{gml}} is a well-known extension of ML in which
the diamond is replaced
with a counting version $\Diamond^{\geq n}$, $n\in\mathbb{N}$, where
$$
  G,v \models \Diamond^{\geq n}\varphi \text{ if }
  |\{ u \in \Nmc(G,v) \mid G,u \models \varphi\}| \geq n.
$$
As a shortcut, we may use $\Diamond^{=k} \psi$ 
to mean
     $\Diamond^{\geq k} \psi \wedge \neg \Diamond^{\geq k+1} \psi$. For more details on GML, see for instance \cite{Goble1970,DeRijke2000}. 

We next introduce \emph{ratio modal logic (RML)} in which  the 
standard ML diamond is also replaced with a counting version, but here the counting is
relative rather than absolute. Diamonds take the form
$\markDiamond^{\geq r}$ and $\markDiamond^{>r}$ with $r \in [0,1]$.
We have $G,v \models \markDiamond^{\geq r}\varphi$ (resp.\ $G,v \models \markDiamond^{>r}$) if the fraction of successors of $v$ that satisfy $\varphi$ is at least $r$ (resp.\ exceeds $r$). If a vertex $v$ has no successors, then we define $G,v \models \markDiamond^{\geq t} \varphi$
and $G,v \not\models \markDiamond^{> t} \varphi$
for all $\varphi$. As a useful shortcut, we may
write $\markDiamond^{=f_i}$ to mean $\markDiamond^{\geq f_i}\varphi\land\lnot\markDiamond^{>f_i}\varphi$. Modal operators of this
kind have occasionally been 
considered in the literature, see for instance~\cite{DBLP:conf/kr/PacuitS04}. RML is a fragment of modal logic with Presburger constraints \cite{DBLP:journals/japll/DemriL10}. We remark that while ML and GML are fragments of first-order logic (FO), the diamond operators of RML
cannot even be expressed in MSO.


The \emph{modal depth} of a modal formula $\varphi$,
no matter whether \GML, \RML, or \MoL, 
is the nesting depth of diamonds in $\varphi$.

\paragraph{Notions of Expressive Power.}
Both GNNs and modal logic formulas may be viewed
as vertex classifiers. We say that a property $P$
is \emph{(uniformly) expressible} by a class
of vertex classifiers \Cmc, such as $\Cmc = \GML$,
if there is a $C \in \Cmc$ such that the
pointed graphs accepted by $C$ are exactly
those in~$P$.
For classes of classifiers $\Cmc_1,\Cmc_2$, we
write $\Cmc_1 \subseteq \Cmc_2$ to mean that every
property expressible by a classifier from $\Cmc_1$ 
is also expressible by a classifier from $\Cmc_2$.
We then also say that $\Cmc_2$ is \emph{at least as expressive} as $\Cmc_1$. 
This notion of expressive power 
is commonly referred to as the \emph{uniform setting}.

This is in contrast to the \emph{non-uniform setting} where a property $P$ is \emph{(non-uniformly) expressible} by a class
of classifiers \Cmc
if for every graph size $n\geq 0$, there is a $C \in \Cmc$ such that the
pointed graphs of size $n$ accepted by $C$ are exactly
those in~$P$. Also in this setting, we may write $\Cmc_1 \subseteq \Cmc_2$ to mean that every
property (non-uniformly) expressible by a classifier from $\Cmc_1$ 
is also (non-uniformly) expressible by a classifier from $\Cmc_2$.
We shall always make clear whether we refer
to the uniform or non-uniform setting. 


   In both the uniform and the non-uniform setting, we may write $\Cmc_1 = \Cmc_2$ as an abbreviation for
$\Cmc_1 \subseteq \Cmc_2 \subseteq \Cmc_1$.
%
The following is clear from the definitions.
\begin{lemma}
\label{lem:relatingexpressiveness}
    Let $\Cmc_1$ and $\Cmc_2$ be two classes of classifiers.
    If $\Cmc_1\subseteq \Cmc_2$ in the uniform setting, then $\Cmc_1\subseteq \Cmc_2$ in the non-uniform setting.
\end{lemma}
We now clarify the relative expressive power of 
the modal logics introduced above, both in the
uniform and in the non-uniform setting.
\begin{restatable}{lemma}{lemMLcomparison}
\label{lem:MLcomparison}
    In the uniform setting,
    \begin{enumerate}
        \item $\MoL \subsetneq \RML$ and $\MoL \subsetneq \GML$;
        \item $\RML \not\subseteq \GML$ and $\GML \not\subseteq \RML$.
    \end{enumerate}
    %
%
    In the non-uniform setting, 
    $\MoL \subsetneq \RML \subsetneq \GML$.
\end{restatable}

\section{Non-Uniform Setting}
\label{sect:nonuniform}

We give logical characterizations of 
GNNs in the non-uniform setting. 
The characterizations 
are absolute, that is, they are not 
relative to FO, MSO, or any other 
background logic. We start with a summary
of all results in this section.
%
\begin{theorem}
\label{thm:nonuniformsummary}
    In the non-uniform setting,
    \begin{enumerate}
      \item\label{itm: mean-nu} $\text{\textnormal{\meanGNN}} \subseteq    \RML \subseteq \simple{\text{\textnormal{\meanGNN}}}$

        \item\label{itm: sum-nu} $\sumGNN \subseteq \GML
        \subseteq \simple{\sumGNN}$
        
        \item $\maxGNN \subseteq \MoL \subseteq \simple{\maxGNN}$.
      
    \end{enumerate}
      This holds for truncated ReLU and ReLU, and for Points~\ref{itm: mean-nu} and~\ref{itm: sum-nu} for every continuous non-polynomial activation function.
\end{theorem}
Point~3 of Theorem~\ref{thm:nonuniformsummary} even holds in the uniform setting, also there without relativization to a background logic.
The same is true
for the second inclusion in Point~2 as per \cite{Barcelo2020}, but as we shall see,
not for any of the other inclusions. 
Together with Lemma~\ref{lem:MLcomparison},
Theorem~\ref{thm:nonuniformsummary} also reproves the following result
from \cite{DBLP:conf/iclr/XuHLJ19}.
\begin{corollary}
    In the non-uniform setting, \begin{align*}
        \maxGNN\subsetneq \text{\textnormal{\meanGNN}}\subsetneq \sumGNN.
    \end{align*}
     %
\end{corollary}
To prove  Theorem~\ref{thm:nonuniformsummary}, we first give
the translations from logic to GNNs, starting with Point~1.
The general strategy is as in \cite{Barcelo2020},
that is, we translate a \ac{rml} formula $\varphi$ with 
$L$ subformulas into a simple GNN with $L$ layers, each
of output dimension~$L$. We first observe
that we may assume w.l.o.g.\ that $\varphi$ contains no  \ac{rml} diamonds of the form $\markDiamond^{\geq t}$. This is because on graphs with at most $n$ vertices such a diamond can be replaced by $\markDiamond^{>t'}$ where
%
$t'$ is the largest rational number that is smaller than $t$ and  can occur as a fraction in a graph where every vertex has at most $n$ successors.

For the GNN translation, we choose a suitable enumeration $\varphi_1,\dots,\varphi_L$
of the subformulas of $\varphi$ and design the
GNN to compute the truth value (that is, 0 or 1) of each subformula $\varphi_i$ at every vertex~$v$ and store it in the $i$-th component of the feature vector for $v$.
The only interesting case are subformulas of the form $\markDiamond^{>t}\varphi_i$. If such a diamond is
satisfied at a node $v$ in a graph $G$ with at most
$n$ vertices, then at least a fraction of 
$$t' = \min\left\{\frac{\ell}{m}\mid 0\leq \ell\leq m\leq n, \frac{\ell}{m}> t\right\}$$
successors of $v$ must satisfy $\varphi_i$.
Note that 
$t'$ is the smallest rational number that is larger than $t$ and  can occur as a fraction in a graph where every vertex has at most $n$ successors.
If $\markDiamond^{>t}\varphi_i$ is violated, then at most a fraction of $t''$ successors
of $v$ can satisfy $\varphi_i$ where
$t''$ is defined like $t'$ except that $\min$ is now $\max$ and  `$>$' is~`$\leq$'. The gap between these
fractions is at least $\frac{1}{n^2}$ and can be amplified using the
matrix $A$ from simple combination functions, which through the bias
vector allows us to compute the desired truth value. 
\begin{restatable}{theorem}{lemmarmltognntranslationnonuniformrelu}\label{lemma: rml-to-gnn-translation-nonuniform}
    $\RML\subseteq \simpleGNN$
    in the non-uniform setting.
    This holds for truncated ReLU and ReLU  activation.
%
\end{restatable}
%
%
We next strengthen Theorem~\ref{lemma:
  rml-to-gnn-translation-nonuniform} to all continuous non-polynomial
activation functions. This is based on universal approximation
theorems from machine learning 
\cite{Pinkus1999}.  Intuitively, approximation suffices because the constant bound
on the size of 
graphs imposed in the non-uniform setting ensures that a \ac{gnn} can generate only a
constant number of different feature vectors, across all input graphs,
and we are good as long as we can distinguish these.
 A variation of the proof also
works for sum aggregation, delivering 
the  second inclusion in Point~2 
of Theorem~\ref{thm:nonuniformsummary}.
\begin{restatable}{theorem}{lemmarmltognntranslationnonuniform}\label{theorem: rml-to-gnn-translation-nonuniform-univ-approx}
In the non-uniform setting and for all continuous non-polynomial activation functions:
\begin{enumerate}

   \item $\RML\subseteq \simpleGNN$;

    \item     $\GML\subseteq \simple{\sumGNN}$.

\end{enumerate}
%
\end{restatable}
Point~2 was proved in  \cite{Barcelo2020} in the uniform setting,
but only for the special case of truncated ReLU activation.

%
%
We next treat the second inclusion in 
Point~3 of Theorem~\ref{thm:nonuniformsummary}, using a minor
variation of the proof in \cite{Barcelo2020}. Like
that proof, our proof even works in the uniform setting.
\begin{restatable}{theorem}{lemMLtomaxGNN}
\label{lem:MLtomaxGNN}
     $\MoL\subseteq \simple{\maxGNN}$
    in the uniform setting.    This holds for truncated ReLU and ReLU activation.
\end{restatable}
We do not know whether this can be strengthened to all continuous non-polynomial activation functions. 

\smallskip

We now turn to the translations from GNNs to logic.
These also rely on the fact that in the non-uniform setting, a \ac{gnn} can generate only a constant number of different feature vectors, across all input graphs. For every feature vector $\bar x$ and every layer $\ell$ of the 
GNN, we can construct a modal logic formula $\varphi^\ell_{\bar x}$ such that
the GNN assigns $\bar{x}$ to a vertex $v$ in layer $\ell$ if and only if $G,v \models \varphi_{\bar{x}}^{\ell}$. Depending on the aggregation function of
the GNN, the formula $\varphi^\ell_{\bar x}$ needs to describe the distribution of feature vectors at the successors of $v$ computed by level $\ell-1$ in varying degrees of detail. For mean, we only need to know the
fraction of successors at which each feature vector was computed, and thus
the formula can be formulated in \RML. For sum,
we need exact multiplicities and thus require a \GML formula.
\begin{restatable}{theorem}{lemmameangnntormlnonuniformtranslation}\label{lemma: mean-gnn-to-rml-non-uniform-translation}
  In the non-uniform setting,
  \begin{enumerate}

      \item $\text{\textnormal{\meanGNN}} \subseteq \RML$;

            \item  $\text{\textnormal{\sumGNN}} \subseteq \GML$.

  \end{enumerate}
  %
\end{restatable}
As observed already in \cite{TenaCucala2024}, with max aggregation the set
of feature vectors ever generated by a GNN, across all input graphs, is finite even without 
bounding the graph size. Using similar arguments
as for Theorem~\ref{lemma: mean-gnn-to-rml-non-uniform-translation}, we can thus show the following.
\begin{restatable}{theorem}{lemfrommaxgnntoML}
\label{lem:frommaxgnntoML}
In the uniform setting,
  $\text{\textnormal{\maxGNN}} \subseteq \MoL$.
\end{restatable}
We remark that all our results obtained in the non-uniform setting also hold in the uniform setting when a constant bound is imposed on the outdegree of vertices.



\section{Uniform Setting}
\label{sect:uniform}

We again start with a summary of our
results. Note that these are now relative
to MSO, except for the case of {\maxGNN}s.
\begin{theorem}
\label{thm:uniformsummary}
    In the uniform setting,
    \begin{enumerate}
      \item $\text{\textnormal{\meanGNN}}  \cap \text{\textnormal{MSO}} \subseteq    \MoL \subseteq \simple{\text{\textnormal{\meanGNN}}}$

        \item $\sumGNN  \cap \text{\textnormal{MSO}}  \subseteq \GML
        \subseteq \simple{\sumGNN}$
        
        \item $\maxGNN \subseteq \MoL \subseteq \simple{\maxGNN}$.
      
    \end{enumerate}
      This holds for truncated ReLU and ReLU activation.
\end{theorem}
Also note that, compared to Theorem~\ref{thm:nonuniformsummary},
the logic associated with \meanGNN{}s
has changed from \RML to \MoL, because \RML diamonds are
not expressible in MSO. With  Lemma~\ref{lem:MLcomparison}, we obtain the following corollary.
\begin{corollary}
\label{cor:unifrelexpower}
    In the uniform setting, \begin{align*}
        \maxGNN =  \text{\textnormal{\meanGNN}} \cap \MSO \subsetneq \sumGNN \cap \MSO.
    \end{align*}
     %
\end{corollary}
We now prove Theorem~\ref{thm:uniformsummary}.
We have already shown Point~3 as Theorems~\ref{lem:MLtomaxGNN} and~\ref{lem:frommaxgnntoML}.
The result in Point~2 is from \cite{Barcelo2020}, stated there for truncated
ReLU and for FO in place of MSO. We may invoke Lemma~\ref{lem:trRELUtoRELU}
for ReLU. 
Regarding the replacement of FO with MSO,
it was observed in \cite{Ahvonen2024} that the 
following is a consequence of results by \cite{Elberfeld2016}.
\begin{lemma}
\label{lem:FromFOtoMSO}
  Any property expressible in MSO and by a GNN is also FO-expressible. This
  only depends on invariance under unraveling and thus holds for
all choices of aggregation, activation, and classification function. 
\end{lemma}
In the remainder of this section, we
prove Point~1 of Theorem~\ref{thm:uniformsummary}.
For the first inclusion, we need suitable versions
of \ac{ef} games. Such games are played by two players, \emph{Spoiler ($S$)} and \emph{Duplicator ($D$)}, who play on two potentially infinite pointed graphs $(G_1,v_1),(G_2,v_2)$. Spoiler's aim is to show that the
graphs are
dissimilar while $D$ wishes to show that they are similar. The game is played in rounds. In the  \emph{GML game}, which in addition is parameterized by a \emph{number of rounds $\ell\in\mathbb{N}$} and a \emph{counting bound $c\in\mathbb{N}^+$}, each round
consists of the following steps~\cite{Otto2019}:
\begin{enumerate}
    \item $S$ chooses $i \in \{1,2\}$ and a set  $U_i\subseteq \mathcal{N}(G_i,v_i)$ with $0 < |U_i| \leq c$;
    \item $D$ selects a set $U_{3-i} \subseteq \mathcal{N}(G_{3-i},v_{3-i})$ with $|U_1| = |U_2|$;
    \item $S$ selects a vertex $u_{3-i}\in U_{3-i}$;
    \item $D$ selects a vertex $u_i\in U_i$.
\end{enumerate}
The game proceeds on the graphs $(G_1,u_1)$
and $(G_2,u_2)$.  Spoiler wins as soon as one of the following conditions hold, possibly at the very beginning of the game:
\begin{itemize}
    \item $\pi^{G_1}(v_1) \neq \pi^{G_2}(v_2)$; 
    \item $D$ fails in Step~2 because $|U_i| > |\mathcal{N}(G_{3-i},v_{3-i})|$.
\end{itemize}
Duplicator wins if one of the following conditions hold:
\begin{itemize}
    \item $S$ cannot choose a non-empty set $U_i$ in Step~1,
    \item after $\ell$ rounds, $S$ has not won.
\end{itemize}
We write $\Emc_{\ell}^{\GML[c]}(G_1,v_1,G_2,v_2)$ to denote the
$\ell$-round \GML game with counting bound~$c$ on pointed graphs $(G_1,v_1)$ and $(G_2,v_2)$.  We may vary
\GML games to obtain games for \MoL.
An \emph{ML game} is a \GML game with counting bound~1. 
Thus Spoiler selects a singleton set $U_i$ in Step~1
and consequently Steps~3 and~4 are trivialized. In
other words, a round consists of first $S$ choosing
$i \in \{1,2\}$ and a vertex $u_i \in  \mathcal{N}(G_i,v_i)$, and $D$ replying with a vertex $u_{3-i} \in  \mathcal{N}(G_{3-i},v_{3-i})$. We denote these games with $\EF{}{\ell}{\MoL}{G_1,v_1,G_2,v_2}$.

We use $\GML[c]$ to denote the fragment of \GML in which
in all diamonds $\Diamond^{\geq n}$ we have $n \leq c$.
%
%
%
\begin{theorem} \label{lemma: ef-gml-equivalence}
Let $\Lmc \in \{ \MoL \} \cup \{ \GML[c] \mid c \geq 0 \}$,
and let $P$ be a vertex property.
    The following  are equivalent for all $\ell\geq 0$:
    \begin{enumerate}
        \item there exists an \Lmc formula $\varphi$ of modal depth at most $\ell$ such that 
        for all pointed graphs  $(G,v)$: $G,v\models \varphi$ if and only if $(G,v) \in P$.
        \item Spoiler has a winning strategy in $\EF{}{\ell}{\Lmc}{G_1,v_1,G_2,v_2}$ for all
        pointed graphs $(G_1,v_1), (G_2,v_2)$ with
        $(G_1,v_1)\in P$ and $(G_2,v_2)\notin P$.
    \end{enumerate}
  \end{theorem}
   Proofs can be found
 in the literature, see for instance  \cite{Otto2019} and Chapter 3.2 in \cite{Goranko2007}. We next recall
 that the following was proved in \cite{DBLP:journals/sigmod/BarceloKMPRS20},
 not specifically for {\meanGNN}s, but 
 in fact independently of the aggregation function used.
\begin{restatable}{theorem}{lemtoAFMLstepone}
\label{lem:toAFMLstepone}
    $\text{\textnormal{\meanGNN}}\cap \MSO \subseteq \GML$ in the uniform setting. 
\end{restatable}
We improve this from \GML to \MoL, exploiting
the fact that properties definable by {\meanGNN}s
are invariant under scaling the graph, that is, 
choosing a $c \geq 1$ and multiplying each vertex
in the graph exactly $c$ times. To make this formal,
let $G = (V, E, \pi)$ be a $\Pi$-labeled graph and  $c \geq 1$.
The \emph{$c$-scaling of $G$} is the $\Pi$-labeled graph $c\cdot G := (V', E', \pi')$ where $V' = \{(v,i)\mid v\in V,\ 1\leq i\leq c\}$, $E' = \{((v,i), (u,j))\mid (v,u)\in E\}$, and $\pi'((v,i)) = \pi(v)$ for all $v \in V$
and $i \in [c]$. The following is immediate from
the fact that the mean of a multiset is invariant under multiplying all multiplicities by a constant $c \geq 1$.
\begin{lemma}\label{lemma: gnn-invariance-under-scaling} 
    Let \Gmc be a \meanGNN on $\Pi$-labeled graphs,
    $(G,v)$ a $\Pi$-labeled pointed graph, and $c \geq 1$. Then  for all $i \in [c]$: 
    $\bar x^L_{G,v} = \bar x^L_{c \cdot G,(v,i)}$.
\end{lemma}
The following relates EF-games for \MoL
to EF-games for $\GML[c]$ on the 
corresponding $c$-scaled graphs.
%
\begin{restatable}{lemma}
{lemmameangnntomlwinningstrategies}
\label{lemma: meangnn-to-ml-winning-strategies}
    Let $(G_1,v_1),\ (G_2,v_2)$ be pointed graphs and $\ell \geq 0$.
    If $D$ has a winning strategy in $\EF{}{\ell}{\MoL}{G_1,v_1,G_2,v_2}$, then $D$ also has a winning strategy in $$\EF{}{\ell}{\GML[c]}{c\cdot G_1, (v_1,k_1), c\cdot G_2, (v_2,k_2)},$$ for all $c\geq 1$ and $k_1,k_2\in [c]$.
\end{restatable}
With Lemma~\ref{lemma: meangnn-to-ml-winning-strategies}
 as the main ingredient, we can now show the following.
\begin{corollary}\label{corollary: meangnn-to-ml-translation}
    $\text{\textnormal{\meanGNN}}\cap\MSO\subseteq \MoL$ in the uniform setting. 
\end{corollary}
\noindent
\begin{proof}\ 
Let $P$ be a vertex property that is 
expressible by a $\text{\textnormal{\meanGNN}}$ and by an $\MSO$
formula. By Theorem~\ref{lem:toAFMLstepone},  $P$ is 
definable by 
a \GML formula  $\varphi$. Let $c$ be maximal such that  $\varphi$ contains a diamond $\Diamond^{\geq c}$.

Assume to the contrary of what we
have to show that $P$ cannot be expressed in \ac{ml}.
Then by Theorem~\ref{lemma: ef-gml-equivalence} for each \mbox{$\ell \geq 0$} there exist pointed graphs $(G_1,v_1) \in P$ and $(G_2,v_2) \notin P$ such that  Duplicator wins $\EF{}{\ell}{\MoL}{G_1,v_1,G_2,v_2}$.
By Lemma~\ref{lemma: meangnn-to-ml-winning-strategies}, Duplicator  also wins $\EF{}{\ell}{\GML[c]}{c\cdot G_1,(v_1,1),c\cdot G_2,(v_2,1)}$. Since $P$ is definable by a 
$\text{\textnormal{\meanGNN}}$, Lemma~\ref{lemma: gnn-invariance-under-scaling} yields $(c\cdot G_1, (v_1,1)) \in P$ and $(c\cdot G_2, (v_2,1))\notin P$.
Therefore, again by Theorem~\ref{lemma: ef-gml-equivalence}, $P$ cannot be defined by a $\GML[c]$ formula; 
a contradiction.
\end{proof}
Regarding the second inclusion of Point~1
of Theorem~\ref{thm:uniformsummary}, we actually prove something stronger.
\begin{restatable}{theorem}{lemmarmltognntranslation}
\label{lemma: rml-to-gnn-translation}
    $\RML\subseteq \simple{\meanGNN}$ in the uniform setting.
\end{restatable}
The proof is similar to that of Theorem~\ref{lemma: rml-to-gnn-translation-nonuniform}. In particular, truth and falsity of formulas is represented by the values~1 and~0 in feature vectors. Importantly, we use a step function as
the activation function. 
The  reason why we cannot use a continuous
activation function such as ReLU in our translation
is
the modal diamond $\Diamond \varphi$: if a vertex has a successor that satisfies $\varphi$, then the mean over all successors may still be arbitrarily close to $0$, inducing a discontinuity.

\section{Uniform Setting, Reloaded}\label{sect:unif_reloaded}

From a practical perspective, the use of 
a non-continuous activation function 
(resulting in a non-continuous combination function)
in the proof of Theorem~\ref{lemma: rml-to-gnn-translation} is unsatisfactory. The combination function of a GNN is often represented as a feed-forward neural network (FNN) with 
a continuous activation function such as
truncated ReLU, ReLU, or sigmoid, and 
is then guaranteed to be 
continuous. Importantly, the use of a non-continuous and thus non-differentiable combination
function precludes a direct use of backpropagation,
gradient descent, and related methods. It
is thus natural to ask whether the translation of ML to {\meanGNN}s can also
be realized using a continuous combination 
function. The answer turns out to be positive.
We use \conmeanGNN to denote the 
class of GNNs in which 
the combination functions $\COM^\ell$, viewed as functions
\mbox{$\COM^\ell:\mathbb{R}^{2\delta^{\ell-1}}\to
  \mathbb{R}^{\delta^\ell}$}, are continuous. 
\begin{theorem}
\label{thm:crazyclass}
    $\MoL \subseteq \conmeanGNN 
    $ in the uniform setting.
\end{theorem}
From a practical perspective, however,
the construction in the proof of Theorem~\ref{thm:crazyclass} is even more unsatisfactory. It uses a combination function
that is continuous, but very artificial: the
function is inspired by and derived from the proof of Cantor's isomorphism theorem. Moreover, there is a price to pay in terms of a rather unnatural classification function. We represent truth and falsity of logical formulas as 
irrational and rational numbers, and consequently the classification function  has to return~1 for all irrational numbers and 0 for all rational ones. We conjecture that our proof can be
improved to yield simple 
{\conmeanGNN}s, at the expense
of making it more technical.


In practice, classification functions are
often threshold functions.\footnote{Or they are represented
 by a feed-forward neural network
 (FNN) to which a threshold is applied. Our model 
  captures this case because we can include the FNN in the COM function of the final GNN layer.}
It is therefore relevant to ask about the expressive power of {\meanGNN}s that use
a continuous combination function and a threshold classification function. We denote
this class with Mean$^{c,t}$-GNN. To
be more precise, the classification function is
required to be of the form 
$$\CLS(\bar{x}) = \begin{cases}
    1&\text{ if }x_i \sim c,\\
    0&\text{ otherwise}
\end{cases}$$
where $i \in [\delta^L]$, ${\sim} \in \{ \geq, > \}$, and~\mbox{$c \in \mathbb{R}$}.
Note that adding the options
${\sim} \in\{\leq, <\}$ is syntactic
sugar because we can replace the last combination function $\COM^L$ with $(-1)\cdot \COM^L$ and compare $x_i$ against $-c$ in the classification function.
We remark that all translations from logic
to GNN given in this paper use threshold classification, except Theorem~\ref{thm:crazyclass}.

\smallskip


We characterize the expressive power
of Mean$^{c,t}$-GNNs, relative to MSO,
in terms of an alternation-free 
fragment of \MoL. Formally, \emph{alternation-free  modal logic (\acs{afml})} 
 is defined by the grammar rule
$$
\begin{array}{rcl}
\varphi &\coloncolonequals & \psi \mid \vartheta\\[1mm]
\psi &\coloncolonequals&  P\mid\lnot P\mid \Box\bot\mid\psi\land\psi\mid \psi\lor\psi\mid \Diamond \psi \\[1mm]
\vartheta  &\coloncolonequals&  P\mid\lnot P\mid \Diamond\top\mid \vartheta\land\vartheta\mid \vartheta\lor\vartheta\mid \Box \vartheta.
\end{array}
$$
We use $\AFML[1]$ to denote all AFML formulas formed
according to the grammar rule for $\psi$ in the
definition of \AFML, and likewise for $\AFML[2]$ 
and the grammar rule for $\vartheta$.
Our main result is  as follows.
\begin{theorem}\label{lemma: mean-gnn-afml-equivalence}
In the uniform setting,
    $$
        \contresmeanGNN\cap \MSO \subseteq \AFML \subseteq  \simplecontresGNN.
    $$
    This holds for truncated ReLU, ReLU, and sigmoid activation.
\end{theorem}
The proof of the first inclusion in 
Theorem~\ref{lemma: mean-gnn-afml-equivalence} relies on
\ac{ef} games for \AFML. For $k \in \{1,2\}$, an \emph{$\AFML[k]$ game} is an \MoL game
subject to the modification that Spoiler chooses the same
value $i=k$ in the first step of each round,  except that $S$
may choose $i = 3-k$ in case that $v_k$ has no
successors. 
 Note that in the latter case, Spoiler
immediately wins because Duplicator cannot respond with a successor of $v_k$ (we in fact have  $G_k,v_k\models\Box\bot$ and $G_{3-k},v_{3-k}\not\models \Box\bot$).
We denote these games with $\EF{}{\ell}{\AFML[k]}{G_1,v_1,G_2,v_2}$. A version
of Theorem~\ref{lemma: ef-gml-equivalence}
for $\AFML[1]$ and $\AFML[2]$ is proved
in the appendix. 
%
%
%
%
%

Using AFML games, it is easy to prove that
basic ML properties such as $\varphi = \Diamond P\land \Box Q$ are not expressible
in AFML, that is, $\AFML \subsetneq \MoL$
(both in the uniform and non-uniform setting). Details are in the appendix.
%
%

To prove Theorem~\ref{lemma: mean-gnn-afml-equivalence}, we start from Corollary~\ref{corollary: meangnn-to-ml-translation}. 
We need some preliminaries.
Let $G = (V,E,\pi)$ be a graph and  $v\in V$. A \emph{path}
in $G$ is a sequence $p=v_0,\dots,v_n$ of vertices from $V$ such that $(v_i,v_{i+1}) \in E$ for 
    all $i < n$. The path \emph{starts at $v_0$} and is of \emph{length} $n$, and we use $\mn{tail}(p)$ to denote $v_n$.
The \emph{unraveling of $G$ at $v$} is the potentially infinite tree-shaped graph  ${\Unr(G,v) = (V',E',\pi')}$ defined as follows:
\begin{itemize}
    
    \item $V'$ is the set of all paths in $G$ that start at $v$;

    \item $E'$ contains an edge $(p,pu)$ if $(\mn{tail}(p),u) \in E$;

    \item $\pi'(p)=\pi(\mn{tail}(p))$.
    
\end{itemize}
For $L \geq 0$, the \emph{unraveling of $G$ at $v$ up to depth $L$}, denoted $\Unr^L(G,v)$, is the (finite) subgraph of $\Unr(G,v)$ induced by all paths of length at most~$L$.

It is well-known that modal formulas are invariant under unraveling up to their modal depth.
%
A similar statement holds for GNNs. 
%
\begin{lemma}[\citet{Barcelo2020}]\label{lemma: gnn-invariance-under-unraveling}
Let $G$ be a graph, $v \in V^G$,  $\mathcal{G}=(L,\{\AGG^{\ell}\}_{\ell \in [L]},\{\COM^{\ell}\}_{\ell \in [L]},\CLS)$, and $1\leq \ell\leq L$. Then
$\bar{x}_{G,v}^\ell = \bar{x}_{\Unr^L(G,v), v}^\ell.$
\end{lemma}



We next show that slightly changing a highly scaled input graph to a \contresmeanGNN does not change the computed value in an unbounded way.
We first formalize what we mean by `slight change'.
\begin{definition}
    Let $G=(V,E,\pi)$ be a  graph and $n \geq 0$.
    A  graph $G' = (V',E',\pi')$ is  an \emph{$n$-extension} of $G$ if it satisfies the following conditions for all $v\in V$:
    \begin{enumerate}
        \item $V\subseteq V'$, $E\subseteq E'$,
        \item for all $v\in V$: $\pi(v) = \pi'(v)$
        \item if $\mathcal{N}(G,v)\neq \emptyset$, then $|\mathcal{N}(G',v)\setminus\mathcal{N}(G,v)| \leq n$, and
        \item if $\mathcal{N}(G,v) = \emptyset$, then $\mathcal{N}(G',v) = \emptyset$.
    \end{enumerate}

\end{definition}
Note that 
$n$-extensions can add up to $n$ fresh successors to any vertex that already has at least one successor.
Intuitively, the value of $n$ will be small compared to the scaling of the graph $G$ of which the $n$-extension is taken. 

\medskip

In what follows, we use the maximum metric and define the distance of
two vectors $\bar{x}$ and $\bar{y}$ of dimension $\delta$ to be ${||\bar{x}-\bar{y}||_{\infty}} = \max_{1\leq i\leq \delta} |x_i-y_i|$. The following
makes precise why we are interested
in $n$-extensions.

\begin{restatable}{lemma}{lemmagnninvarianceundersmallchanges}
\label{lemma: gnn-invariance-under-small-changes}
    Let $\mathcal{G}$ be a \conmeanGNN with $L$ layers. Then for all $\varepsilon>0$, $n \geq 1$, and $\ell \in [L]$, there exists a constant $c$ such that for all $c'\geq c$, graphs $H = c'\cdot G$, vertices $(v,i)$ in $H$, and  $n$-extensions $H'$ of $H$:
     $||\bar{x}_{H,(v,i)}^{\ell} - \bar{x}_{H',(v,i)}^\ell||_\infty < \varepsilon$.
\end{restatable}
One ingredient to the proof of Lemma~\ref{lemma:
  gnn-invariance-under-small-changes} is the observation that for
every \conmeanGNN, there are constant upper and lower bounds on the values that may occur in feature vectors, across all
input graphs. The reader might want to compare this with {\maxGNN}s
where even the number of such values is
bounded by a constant, see the discussion
before Theorem~\ref{lem:frommaxgnntoML}. 

\begin{restatable}{theorem}{lemconmeangnnsubseteqafml}
\label{lem:conmeangnnsubseteqafml}
    $\contresmeanGNN\cap\MSO \subseteq \AFML$ in the uniform setting.
\end{restatable}

\noindent
\begin{proof}\ {\bf (sketch)}
    Assume to the contrary that there exists a {\contresmeanGNN} $\mathcal{G}$ with $L$ layers that is equivalent to an MSO formula, but  not  to an \ac{afml} formula. By Corollary \ref{corollary: meangnn-to-ml-translation}, \Gmc is equivalent to an \MoL formula $\varphi$.
    Assume first that the classification function of \Gmc uses `$>$' rather than `$\geq$'. 
    
Because $\varphi$ is not expressible in \ac{afml}, for each $\ell\in\mathbb{N}$ there exist pointed graphs $(G,v)$ and $(G',v')$ with \mbox{$G,v\models \varphi$} and $G',v'\not\models\varphi$ such that $D$ has a winning strategy in $\EF{}{\ell}{\AFML[1]}{G,v,G',v'}$.
    Our aim is to transform
    $(G,v)$ into a pointed graph $(H,u)$ such that 
    \begin{enumerate}
    \item[(i)]
    $H,u\models\varphi$ and
    \item[(ii)] $D$ has a winning strategy in $\EF{}{\ell}{\MoL}{H,u,G',v'}$.
    \end{enumerate}
    Then, by Theorem~\ref{lemma: ef-gml-equivalence}, $\varphi$ is not expressible in \ac{ml}. A contradiction.

    It can be shown that since $D$ has a  winning strategy in
$\EF{}{\ell}{\AFML[1]}{G,v,G',v'}$, they also have one in $\EF{}{\ell}{\AFML[1]}{\Unr^K(G,v), v, G',v'}$ for all $K \geq \ell$.
Moreover, in AFML$[1]$-games in
which the first graph is tree-shaped, the existence of a winning strategy for $D$ implies the existence of a memoryless winning strategy. We may view such a strategy as a function $\ws:V^{\Unr^K(G,v)}\to V^{G'}$  such that if $S$ plays vertex $u$ in
$\Unr^K(G,v)$, then $D$ always
answers with $\ws(u)$. We also 
set $\ws(v)=v'$.

Let $m=|V^{G'}|$ and $K = \max(\ell, L)$.  Since classification is
based on `$>$',
we find an $\varepsilon>0$ such that
     all pointed graphs $(H,u)$ with $||\bar{x}^L_{G,v}-\bar{x}^L_{H,u}||_\infty < \varepsilon$ are accepted by \Gmc.
By Lemma \ref{lemma: gnn-invariance-under-small-changes}, there exists
a $c$ such that $||\bar{x}_{G'',v}^L - \bar{x}_{X,v}^L||_\infty<\varepsilon$
in each $m$-extension $X$ of $G''=c\cdot \Unr^K(G,v)$.
%
%
%
Now, $(H,u)$ is defined
as follows:
%
%
\begin{enumerate}

    \item start with $G''=c\cdot \Unr^K(G,v)$;

    \item take the disjoint union with all $\Unr^K(G',v')$,  $v'\in V^{G'}$;

    \item for each vertex $(u,i) \in V^{G''}$ that has at least one successor, let $\mathcal{N}(G',\ws(u)) =\{u_1',\ldots, u_m'\}$.
    Add to $(u,i)$ the fresh successors $u_1',\ldots,u_m'$;

    \item $u=(v,1)$.
\end{enumerate}
We show in the appendix that Conditions~(i) and~(ii) are satisfied. We also deal there with
the case where the classification function is based on `$\geq$', using AFML$[2]$.
\end{proof}

Next, we show that each formula in \ac{afml} can be realized by a simple \contresmeanGNN. 
As in the proof of Theorem~\ref{lemma: rml-to-gnn-translation},
we face the challenge that the mean over all
successors may diminish values. Here, we 
address this by representing
the truth of subformulas by values from the range $(0,1]$ and falsity
by value~$0$. With this encoding, we can
realize the modal diamonds of AFML$[1]$
using truncated ReLU activation, but we
cannot realize modal boxes. Still, it is
easy to also treat AFML$[2]$ using duality
arguments.
%
%
\begin{restatable}{theorem}{lemmaafmltocmgnntranslation}
\label{lemma: afml-to-cmgnn-translation}
    $\AFML\subseteq \simplecontresGNN$ in the uniform setting. This holds for truncated ReLU and ReLU as activation functions. 
\end{restatable}

The proof of Theorem~\ref{lemma: afml-to-cmgnn-translation} can be 
adapted to sigmoid activation.
This once more requires non-trivial 
modifications. Truth is now encoded
by values from the range $(\frac{1}{2},1)$ and falsity by
value~$\frac{1}{2}$.
%
%

\begin{restatable}{theorem}{lemAFMLinsimplemeanGNNwithsigmoid}
  \label{thm:sfjlherogahfga}
$\AFML\subseteq \simplecontresGNN$ in the uniform setting, with sigmoid as the activation function. 
\end{restatable}

\section{Conclusion}

We have identified logical characterizations of graph neural networks
with mean aggregation, in several different settings.  Some interesting
questions remain open. For instance, we would like to know whether
Theorem~\ref{lem:MLtomaxGNN} can be strengthened to all continuous
non-polynomial activation functions, in the spirit of
Theorem~\ref{theorem: rml-to-gnn-translation-nonuniform-univ-approx}.
It would also be interesting to consider broader classes of activation
functions in the uniform setting.
Finally, it would be interesting to find absolute logical characterization of
 {\meanGNN}s and {\contresmeanGNN}s, that is, characterizations that are
not relative to any background logic such as FO or MSO. In fact, we can already provide
two first observations on this subject.
\begin{theorem}~\\[-4mm]
\label{thm:summarymeanabsoluteuniform}
\begin{enumerate}
    
\item    The property `there exist more successors that satisfy~$P_1$ than successors that satisfy $P_2$' is not expressible in \ac{rml}, but  by a simple {\contresmeanGNN}.

\item  The \ac{rml} formula $\markDiamond^{>\frac{1}{2}}\markDiamond^{>\frac{1}{2}}P$  is not expressible by a {\contresmeanGNN}.

\end{enumerate}
\end{theorem}
%


\section*{Acknowledgements}

Carsten Lutz was supported by  DFG project LU 1417/4-1.

\bibliography{aaai25}
\cleardoublepage

\appendix



\section{Proofs for Section~\ref{sect:prelims}}

\lemtrRELUtoRELU*
\noindent
\begin{proof}\
      To convert a simple $\AGG$-\ac{gnn} \Gmc
   with $\trRELU$ activation
   into a simple $\AGG$-\ac{gnn} $\Gmc'$ with $\RELU$ activation, we can thus 
   proceed as follows. We double the number of layers.
   Each layer $\ell$ of \Gmc with output dimension $\delta^\ell$ is simulated by layers $(2\ell-1,2\ell)$ of $\Gmc'$ with output dimensions  $2\delta^\ell$ and $\delta^\ell$, respectively. If the result of layer $\ell$ of \Gmc is $\trRELU(\bar x)$, then the
   $2\ell-1$-st layer of $\Gmc'$ computes both 
   $\RELU(\bar x)$ and $\RELU(\bar x-\bar 1)$, where $\bar 1$ is the all-1 vector of appropriate dimension, and stores
   the result in the now twice as large output feature vectors. The $2\ell$-th layer of $\Gmc'$ then 
   computes $\RELU(\RELU(\bar x) - \RELU(\bar x-\bar 1))$, which is equal to $\RELU(\bar x) - \RELU(\bar x-\bar 1)$, in a straightforward way.
\end{proof}

\lemMLcomparison*
\noindent
\begin{proof}\ 
    We start with the uniform setting.
    It is clear that $\MoL \subseteq \GML$ and $\MoL \subseteq \RML$:
    \begin{itemize}
        \item in \ac{gml}, we have $\Diamond\varphi \equiv \Diamond^{\geq 1}\varphi$; 
        \item in \ac{rml}, we have $\Diamond \varphi \equiv \markDiamond^{>0}\varphi$. 
        \end{itemize}
    We prove below in the non-uniform setting that $\GML \not \subseteq \MoL$,
    $\RML \not\subseteq \MoL$, and $\GML \not\subseteq \RML$.
    By Lemma~\ref{lem:relatingexpressiveness}, these
    results carry over to the uniform setting.
        
It thus remains to argue that $\RML \not \subseteq \GML$. 
Consider
the \ac{rml} formula $\varphi = \markDiamond^{\geq\frac{1}{2}} P$. We argue that it is not expressible in \ac{gml}.
        To prove this, notice that a \ac{gml} formula $\varphi$ with maximal counting constant $c$ cannot distinguish between a vertex with $c$ successors that satisfy $P$ and $c$ successors that do not satisfy $P$, and a vertex with $c$ successors that satisfy $P$ and $c+1$ successors that do not satisfy $P$. This can in fact be proved by induction on the structure of $\varphi$.
        But these two graphs can be distinguished by $\markDiamond^{\geq \frac{1}{2}}P$.

    \smallskip
    In the non-uniform setting, we obtain $\MoL\subseteq \RML$ from the uniform setting and Lemma~\ref{lem:relatingexpressiveness}. To show $\RML\subseteq \GML$, we note that on graphs of size at most $n$, the \RML diamond $\markDiamond^{\geq r} \varphi$ can be expressed as 
   $$
    \Diamond^{=0} \top \vee \bigvee_{\substack{0 \leq p \leq q \leq n \\ q\neq 0,\ p/q \geq r}}
      \Diamond^{=q} \top \wedge \Diamond^{=p} \varphi.
   $$
    %
The  \RML diamond $\markDiamond^{> r} \varphi$ can be expressed similarly.

    To prove $\RML \not\subseteq \MoL$, consider the 
    \RML formula  $\markDiamond^{>\frac{1}{2}}P$. We 
    claim that it is not nonuniformly expressible in $\MoL$.
    We construct two graphs that can be distinguished by $\markDiamond^{>\frac{1}{2}}P$ but cannot be distinguished by any formula in \ac{ml}.
    In fact, $\MoL$ cannot count and thus cannot distinguish between a vertex with two successors, exactly one of
        which satisfies $P$, and a vertex with one successor that  satisfies $P$ and two successors that do not. Formally, this can be proved using
        bisimulations, see for instance \cite{DBLP:books/cu/BlackburnRV01}.
        The proof of $\GML\not\subseteq \MoL$ is very
        similar, using the
    \GML formula  $\Diamond^{\geq 2}P$. 

    It remains to prove $\GML \not\subseteq \RML$.
                Consider the \ac{gml} formula $\varphi = \Diamond^{\geq 2}P$. We claim that it is not expressible in \ac{rml}.
        In fact, it is easy to see that there does not exist an \ac{rml} formula $\psi$ that can distinguish between a vertex that has one successor that satisfies $P$ (and no other
        successors) and a vertex that has two successors
        that satisfy $P$. Formally, this can be proved by induction on the structure of $\psi$.%
    \end{proof}

\section{Proofs for Section~\ref{sect:nonuniform}}

\subsection{Proof of Theorem~\ref{lemma: rml-to-gnn-translation-nonuniform}}

\lemmarmltognntranslationnonuniformrelu*

\noindent
\begin{proof}\
     We start with truncated ReLU.
    Let $\varphi$ be a formula in \ac{rml} over
     a finite set $\Pi = \{P_1,\ldots, P_r\}$  of vertex labels and fix a maximum graph size $n \geq 1$. As observed in the main body of the paper,
     we may assume w.l.o.g.\ that $\varphi$ contains no  \ac{rml} diamonds of the form $\markDiamond^{\geq t}$, since they can be replaced by $\markDiamond^{>t'}$, where
     $$t' = \max\left\{\frac{\ell}{m}\mid 0\leq \ell\leq m\leq n, \frac{\ell}{m}< t\right\}.$$
     Note that \(t'\) is the largest rational number that is smaller than \(t\) and that can occur as a fraction in a graph where every vertex has at most \(n\) successors.

Let $\varphi_1,\ldots,\varphi_L$ be an enumeration of the subformulas of $\varphi$ such that (i)~$\varphi_i=P_i$ for $1 \leq i \leq r$ and
(ii)~if $\varphi_\ell$ is a subformula of $\varphi_k$,
then $\ell < k$.
As in \cite{Barcelo2020}, we construct a simple \ac{gnn} \Gmc with $L$ layers, all of input and output dimension $L$.
To encode satisfaction of a subformula $\varphi_i$ 
at a vertex~$v$, we store value $1$ in the $i$-th component of the feature vector of $v$; likewise, falsification
of $\varphi_i$ is encoded by value $0$.
The \ac{gnn}  evaluates one subformula in each layer
so that for every $k \in [L]$, from layer $k$ on all subformulas $\varphi_1,\ldots,\varphi_k$ are  encoded correctly.
All layers use the same combination function.\footnote{Such GNNs in are  called `homogeneous' in \cite{Barcelo2020}.}

We define the matrices $A,C\in \mathbb{R}^{L}\times \mathbb{R}^{L}$ and the bias vector $\bar{b}\in\mathbb{R}^L$ that define the combination function of the (simple) GNN in the following way, where
all entries that are not mentioned explicitly have value~$0$. Let $k \in [L]$. We make a case 
distinction to set certain entries:
\begin{enumerate}[label=\textit{Case }\arabic*:, ref=\arabic*, leftmargin=*]
    \item \label{enumerate: rml-to-gnn-case1}$\varphi_k = P_k$. Set $C_{k,k} = 1$.
    \item \label{enumerate: rml-to-gnn-case2}$\varphi_k = \lnot \varphi_i$. Set $C_{i,k} = -1$ and $b_k = 1$.
    \item \label{enumerate: rml-to-gnn-case3}$\varphi_k = \varphi_i\lor \varphi_j$. Set $C_{i,k} = C_{j,k} = 1$.
    \item \label{enumerate: rml-to-gnn-case4}$\varphi_k = \markDiamond^{>t}\varphi_i$. Set $A_{i,k} = n^2$ and  $$b_k = -n^2\cdot \max\left\{\frac{\ell}{m}\mid 0\leq \ell\leq m\leq n, \frac{\ell}{m}\leq t\right\}.$$
\end{enumerate}
As the classification function we use
    $$\CLS(\bar{x}) = \begin{cases}
        1 & \text{ if } x_L>0\\
        0 & \text{otherwise}.
    \end{cases}$$
We next show correctness of the translation.
\\[2mm]
{\bf Claim 1.} For all $\varphi_k$, $1 \leq k \leq L$, the following holds:
if $v \in V^G$ and $k\leq k' \leq L$, then $$(\bar{x}_{G,v}^{k'})_{k} = \begin{cases}
    1&\text{ if } G,v\models \varphi_k,\\
    0&\text{ otherwise.}
\end{cases}$$
The proof is by induction on $k$, distinguishing 
Cases~1 to~4.
For Cases \ref{enumerate: rml-to-gnn-case1}, \ref{enumerate: rml-to-gnn-case2}, \ref{enumerate: rml-to-gnn-case3}, the (straightforward) arguments can be found in \cite{Barcelo2020}.
For Case~\ref{enumerate: rml-to-gnn-case4}, we first observe that different fractions with denominators bounded by $n$ differ at least by a certain amount.
\\[2mm]
{\bf Claim 2.} For all $n, n_1, n_2\in\mathbb{N}^+$, $m_1,m_2\in\mathbb{N}$ with $n_1,n_2\leq n$ and $\frac{m_1}{n_1}\neq \frac{m_2}{n_2}$:
    $$\left|\frac{m_1}{n_1}-\frac{m_2}{n_2}\right| \geq \frac{1}{n^2}.$$
\\[2mm]
The claim holds because the least common multiple
between the denominators $n_1$ and $n_2$ is at most $n^2$ and  the smallest non-zero difference between any two such fractions is clearly  at least 
$\frac{1}{n^2}$.

\medskip

Now consider Case~4 with $\varphi_k = \markDiamond^{>t}\varphi_i$, assume that Claim~1
has already been shown for $\varphi_1,\dots,\varphi_{k-1}$, and let $k \leq k' \leq L$. Consider any vertex
$v \in V^G$. Since $|V^G| \leq n$, the fraction of successors of $v$ that satisfy $\varphi_i$
can clearly be represented by $\frac{\ell}{m}$ for some $\ell,m$ with $\ell\leq m\leq n$.
Since $i < k$, Claim~1 has already been shown for 
$\varphi_i$. Consequently, the $i$-th component of the vector computed by mean aggregation is $\frac{\ell}{m}$ and the 
$i$-th component of the new feature vector stored 
by level $k'$ of the GNN at node $v$ is $$(\bar{x}_v^{k'})_k = \trRELU(n^2\frac{\ell}{m}+ b_k).$$
First assume that $G,v\models \varphi_k$.
Then $\frac{\ell}{m} >t$ and by definition
of $\bar b_k$ and by Claim~2, this implies
$n^2\frac{\ell}{m}+ b_k \geq 1$.
Thus, by definition of the truncated ReLU the above value is~$1$, as required.

Now assume that  $G,v\not\models\varphi_k$. Then 
$\frac{\ell}{m} \leq t$ and by definition
of $\bar b_k$, this implies $n^2\cdot\frac{\ell}{m}+b_k \leq 0$.
Thus, by definition of the truncated ReLU the above value is~$0$, as required. This finishes the proof
of Claim~1.

\medskip
For the non-truncated ReLU, it suffices to invoke Lemma~\ref{lem:trRELUtoRELU}. 
\end{proof}

\subsection{Proof of Theorem~\ref{theorem: rml-to-gnn-translation-nonuniform-univ-approx}}

The proof uses universal approximation theorems which we introduce next.
\begin{definition}
    A function $f:\mathbb{R}\to\mathbb{R}$ has the universal approximation property if for every compact $K\subseteq \mathbb{R}^n$, every continuous $g:K\to\mathbb{R}^m$, and every $\varepsilon>0$ there exist $d\in\mathbb{N}$, matrices $W_1\in\mathbb{R}^{n\times d}$ and $W_2\in\mathbb{R}^{d\times m}$ and a vector $\bar{b}\in \mathbb{R}^d$ such that $$\sup_{\bar{x}\in K} ||g(\bar{x}) - f(\bar{x}W_1+\bar{b})W_2||_\infty < \varepsilon.$$
\end{definition}

There are several theorems characterizing functions which have the universal approximation property.
For example, \cite{Leshno1993} show that
\begin{lemma}\label{lemma: continuous-non-poly-univ-approx}
    A continuous function $f:\mathbb{R}\to\mathbb{R}$ has the universal approximation property if and only if $f$ is not polynomial.
\end{lemma}
\begin{corollary}
    Truncated ReLU, ReLU, and sigmoid have the universal approximation property.
\end{corollary}

\cite{Hornik1989} show the universal approximation property for all monotone and bounded, but possibly noncontinuous, functions.
See the survey \cite{Pinkus1999} for more information.

For the proof of Theorem~\ref{theorem: rml-to-gnn-translation-nonuniform-univ-approx} we will use \acp{gnn} which satisfy some continuity conditions for the combination and aggregation function.

We remind the reader that for $X\subseteq \mathbb{R}^\gamma$, 
a function \mbox{$f: X\to \mathbb{R}^{\gamma'}$} is 
\emph{continuous} if for all $\bar{x}\in X$ and all $\varepsilon>0$ there exists a $\delta>0$ such that for all $\bar{y}\in X$, $||\bar{x} - \bar{y}||_{\infty} < \delta$ implies $||f(\bar{x})-f(\bar{y})||_{\infty} < \varepsilon$.
The function $f$ is \emph{uniformly continuous} if $\delta$ can be chosen independently of $\bar{x}$.
That is, for all $\varepsilon>0$ there exists a $\delta>0$ such that for all $\bar x,\bar y\in X$, $||\bar x-\bar y||_{\infty} < \delta$ implies $||f(\bar x)-f(\bar y)||_{\infty} < \varepsilon$. 

\begin{definition}
  \label{def:boundedcontagg}
    We call an aggregation function $\AGG$ with input dimension
    $\gamma$ \emph{bounded continuous} if for all $\varepsilon > 0$ and \mbox{$\bar{x}_1,\ldots,\bar{x}_n\in\mathbb{R}^\gamma$,} there exists a $\delta > 0$ such that for all 
    \mbox{$\bar{y}_1,\ldots,\bar{y}_n\in\mathbb{R}^\gamma$} with $||\bar{x}_i-\bar{y}_i||_\infty < \delta$ for all $i\in[n]$, we have $||\AGG(\{\!\{\bar{x}_1,\ldots,\bar{x}_n\}\!\}) - \AGG(\{\!\{\bar{y}_1,\ldots,\bar{y}_n\}\!\}) ||_\infty < \varepsilon.$
\end{definition}
\noindent 
As we will see, the notion of bounded continuity guarantees that in the non-uniform setting the aggregation function behaves like a continuous function, since in graphs up to size $n$, the aggregation function is applied to multisets of size at most $n$.

\begin{lemma}
$\MEAN$, $\MAX$ and $\SUM$ are bounded continuous. 
\end{lemma}
\noindent
\begin{proof}\
 In case $\AGG\in \{\MEAN, \MAX\}$ it can be verified easily that if $||\bar{x}_i-\bar{y}_i||_\infty < \varepsilon$ for all \(i\in [n]\), then $||\AGG(\{\!\{\bar{x}_1,\ldots,\bar{x}_n\}\!\}) - \AGG(\{\!\{\bar{y}_1,\ldots,\bar{y}_n\}\!\}) ||_\infty < \varepsilon$.
    In these cases, the choice of $\delta$ thus only depends on $\varepsilon$, and therefore $\MEAN$ and $\MAX$ actually satisfy a very strong version of bounded continuity. In particular, it is bounded continuous in the sense of Definition~\ref{def:boundedcontagg}
    
     For $\AGG = \SUM$, $\delta$ additionally depends on the size of the multisets. If $||\bar{x}_i - \bar{y}_i||_\infty < \frac{\varepsilon}{n}$ for all \(i\in[n]\), then $||\AGG(\{\!\{\bar{x}_1,\ldots,\bar{x}_n\}\!\}) - \AGG(\{\!\{\bar{y}_1,\ldots,\bar{y}_n\}\!\}) ||_\infty < \varepsilon$.
    It is still bounded continuous in the sense of Definition~\ref{def:boundedcontagg}, even in a stronger sense since
  $\delta$ does not depend on the feature vectors but only on the size of the multisets. The reader should think of Definition~\ref{def:boundedcontagg} as a minimum requirement for bounded continuity.
\end{proof}

\smallskip
%

%
\begin{lemma}\label{lemma: continuous-layers-are-uniformly-continuous}
    Let $\AGG, \COM$ be a \ac{gnn} layer with input dimension $\gamma$ where $\AGG$ is  bounded continuous and $\COM$ is continuous.
    Then for all finite sets $\chi\subseteq\mathbb{R}^\gamma$, $n\in\mathbb{N}$, $\varepsilon > 0$ there exists a $\delta>0$ such that for all $m\in[n]$, $\bar{x}_0,\ldots, \bar{x}_m\in \chi$ and $\bar{y}_0,\ldots, \bar{y}_m\in\mathbb{R}^\gamma$ with $||\bar{x}_i-\bar{y}_i||_\infty < \delta$, we have \begin{align*}
        ||&\COM(\bar{x}_0, \AGG(\{\!\{ \bar{x}_1,\ldots,\bar{x}_m\}\!\})) \\
        &- \COM(\bar{y}_0, \AGG(\{\!\{ \bar{y}_1,\ldots,\bar{y}_m\}\!\}))||_\infty < \varepsilon.
    \end{align*}
\end{lemma}

\noindent
\begin{proof}\ 
It suffices to show that for each size of the multisets $m$ and each $\chi$ and $\varepsilon$ there exists such $\delta_m$.
For a given $n$ we then can choose $\delta = \min_{0\leq i\leq n} \delta_i$.

Since $m$ is fixed, we can view $\AGG$ as a continuous function $\AGG: \mathbb{R}^{m\gamma}\to \mathbb{R}^\gamma$.
Thus, we can view the layer as a function $L: \mathbb{R}^{(m+1)\gamma}\to \mathbb{R}^\gamma$.
$L$ is continuous, since continuous functions are closed under composition.

We define for each $\bar{x}\in\chi$ the set $I_{\bar{x}} = \{\bar{y}\in\mathbb{R}^\gamma\mid ||\bar{x}-\bar{y}||_\infty \leq 1\}$, which is closed and bounded.
We define $I = \bigcup_{\bar{x}\in\chi} I_{\bar{x}}$, which is a finite union of closed and bounded sets and thus also closed and bounded.
By the Heine-Borel theorem from real analysis, $I$ is compact and by the Heine-Cantor theorem, $L$ restricted to the domain $I^{m+1}$ is uniformly continuous.
Hence, we can find a $\delta'$ such that for all $\bar{x},\bar{y}\in I^{m+1}$ with $||\bar{x}-\bar{y}||_\infty < \delta'$ we have $||L(\bar{x})-L(\bar{y})||_\infty < \varepsilon$.
We can now choose $\delta_m = \min(1, \delta')$.
\end{proof}

We also observe that if the size of the input graphs is bounded by a constant, a \ac{gnn} can generate only a constant number of different feature vectors, across all input graphs. In fact, this is an immediate
consequence of the fact that, for every finite set of vertex labels $\Pi$ and every $n \geq 1$, there are only finitely many $\Pi$-labeled graphs of size at most~$n$.

\begin{lemma}\label{lemma: nonuniform-finitely-many-feature-vectors}
    Let $\mathcal{G}$ be a \ac{gnn} with $L$ layers over some finite set of vertex labels $\Pi$.
    For each layer $\ell$ of output dimension $\delta^\ell$ and for all $n \geq 1$, let 
    $\chi^\ell_n\subseteq \mathbb{R}^{\delta^\ell}$ denote the set of feature vectors $\bar x$ such that
    for some input graph $G$ of size at most $n$ and some vertex $v$
    in $G$, \Gmc generates $\bar x$ at $v$ in layer $\ell$.
    Then 
    $\chi^{\ell}_n$ is finite.    
\end{lemma}

We consider aggregation functions that commute with matrix multiplication.

\begin{lemma}
    For all $\AGG\in \{\SUM, \MEAN\}$, $x_1,\ldots,x_n\in\mathbb{R}^{\delta}$ and matrices $A\in  \mathbb{R}^{\delta\times \gamma}$ $$\AGG(\{\!\{\bar{x}_1,\ldots,\bar{x}_n\}\!\})A = \AGG(\{\!\{\bar{x}_1A,\ldots,\bar{x}_nA\}\!\}).$$
\end{lemma}
\noindent\begin{proof}\ 
    $\MEAN$ commutes with  matrix multiplication, since for all $\bar{x}_1,\ldots,\bar{x}_n\in\mathbb{R}^{\delta}$ and $A\in\mathbb{R}^{\delta\times\gamma}$ 
    \begin{align*}
        \MEAN(\{\!\{\bar{x}_1,\ldots,\bar{x}_n\}\!\})A &= \left(\frac{1}{n}\sum_{i = 1}^n \bar{x}_i\right)A \\
        = \frac{1}{n}\sum_{i=1}^n (\bar{x}_iA) &= \MEAN(\{\!\{\bar{x}_1A,\ldots, \bar{x}_nA\}\!\}).
    \end{align*}
    The proof that \SUM\  also commutes with matrix multiplication is analogous.
\end{proof}
We remark that $\MAX$ does not commute with matrix multiplication.
A counterexample is $\bar{x}_1 = 0, \bar{x}_2 = 1$ and $A = \begin{pmatrix}-1\end{pmatrix}$, since $\MAX(\{\!\{0,1\}\!\})A = -1 \neq 0 = \MAX(\{\!\{0,-1\}\!\}) = \MAX(\{\!\{ 0A, 1A\}\!\})$.

The following lemma is the main ingredient to the proof of Theorem~\ref{theorem: rml-to-gnn-translation-nonuniform-univ-approx}.

\begin{restatable}{lemma}{placeholder}\label{lemma: nonuniform-approximation-of-gnns}
    Let $\mathcal{G} = (L, \{\AGG^\ell\}_{\ell\in[L]}, \{\COM^\ell\}_{\ell\in[L]}, \CLS)$ be a \ac{gnn} where each $\AGG^\ell$ is bounded continuous and each $\COM^\ell$ is continuous.
    Let each $\AGG^\ell$ commute with matrix multiplication.
    Let $f:\mathbb{R}\to\mathbb{R}$ be a function with the following properties.
    \begin{enumerate}
        \item $f$ has the universal approximation property.
        \item There exist $a,b\in\mathbb{R}$ and an $\varepsilon > 0$ such that all $x\in (a-\varepsilon,a+\varepsilon)$ and $y\in (b-\varepsilon,b+\varepsilon)$ satisfy $f(x)<f(y)$.
    \end{enumerate}

    Then for all $n\in\mathbb{N}^+$ there exists a simple \ac{gnn} $\mathcal{G}'$ with activation function $f$ and $L+2$ layers such that $\mathcal{G}$ and $\mathcal{G}'$ are equivalent on graphs up to size $n$.
    $\mathcal{G}'$ uses for each layer $1\leq \ell\leq L$ the aggregation function $\AGG^\ell$ and $\AGG^{L+1}$ and $\AGG^{L+2}$ can be chosen arbitrarily.
\end{restatable}

\noindent
\begin{proof}\ Let $\Gmc$ and $f$ be as in the lemma, and let
  $n\in\mathbb{N}^+$ be a bound on the graph size.  Let
  $\delta^0,\ldots,\delta^L$ be the dimensions of the layers of
  $\mathcal{G}$. We will use the values $a$ and $b$ from Point~2 to
  define the classification function of $\Gmc'$, and the $\pm
  \varepsilon$-balls around $a$ and $b$ are needed because the layers of $\Gmc'$ will only approximate
  those of $\Gmc$. To prepare for this, we first introduce
  a modest extension $\Hmc$ of $\Gmc$ with one additional layer
   of output dimension $1$ that already introduces the values $a$ and $b$. 
    For each $\bar{x}_v\in \chi_n^L$ and $\bar{x}_a\in\mathbb{R}^{\delta^L}$, define
    $$\COM^{L+1}(\bar{x}_v, \bar{x}_a) = \begin{cases}
        b&\text{ if } \CLS(\bar{x}_v) = 1\\
        a&\text{ otherwise.}
      \end{cases}$$
    Since $\chi_n^L$ is a finite set and $\bar{x}_a$ is ignored by $\COM^{L+1}$, we can find a polynomial (which is continuous) that extends $\COM^{L+1}$ to the domain $\mathbb{R}^{\delta^L+ \delta^L}$.
    Let $\AGG^{L+1}$ be any bounded continuous  \footnote{In principle, there is no need for $\AGG^{L+1}$
      to be bounded continuous. We only assume this here
    so that the preconditions of Lemma~\ref{lemma:
        continuous-layers-are-uniformly-continuous} are satisfied.} aggregation
    function.
    For the classification function, we choose a threshold between $a$ and $b$.
    Let $\sim$ be $>$ if $b > a$ and let $\sim$ be $<$ otherwise.
    Then $$\CLS'(x) = \begin{cases}
        1&\text{ if } x \sim \frac{a+b}{2}\\
        0&\text{ otherwise.}
    \end{cases}$$
    The \ac{gnn} $$\mathcal{H} = (L+1, \{\COM^\ell\}_{\ell\in[L+1]},
    \{\AGG^\ell\}_{\ell\in[L+1]}, \CLS')$$ is equivalent to \Gmc on
    graphs of size at most $n$, since $\CLS(\bar{x}) =
    \CLS'(\COM^{L+1}(\bar{x}, \bar{y}))$ for all $\bar{x}\in \chi_n^L$
    and $\bar{y}\in\mathbb{R}^{\delta^L}$. 
    In addition, each aggregation function is bounded continuous and
    each combination function is continuous.
    For all $\ell \in [L+1]$, let $\chi_n^\ell$ be the set of feature vectors
    computed by $\mathcal{H}$ in layer $\ell$ on some pointed graph of size at most~$n$.
    By Lemma~\ref{lemma: nonuniform-finitely-many-feature-vectors},
    all these sets are finite.

    We now show how to approximate each layer in $\mathcal{H}$ 
    using $f$, in a way that shall allow us to construct the desired
    GNN~$\Gmc'$.
    We define the approximations of the layers inductively, starting with the last layer.
    The desired error bound of the last layer is $\varepsilon_{L+1} = \varepsilon$.
    
    Knowing the desired error bound $\varepsilon_{\ell}$ we define the
    approximation of the $\ell$-th layer in the following way.
    Since $\AGG^\ell$ is bounded continuous and $\COM^\ell$ is continuous, by Lemma~\ref{lemma: continuous-layers-are-uniformly-continuous} there exists an $\varepsilon_{\ell-1}$ such that for all $m\in [n]$, $\bar{x}_0,\ldots,\bar{x}_m\in\chi_n^{\ell-1}$ and $\bar{y}_0,\ldots, \bar{y}_m\in \mathbb{R}^{\delta^\ell}$ with $||\bar{x}_i-\bar{y}_i||_\infty < \varepsilon_{\ell-1}$, we have 
\begin{align*}
        ||&\COM^\ell(\bar{x}_0, \AGG^{\ell}(\{\!\{ \bar{x}_1,\ldots,\bar{x}_m\}\!\})) \\
        &- \COM^\ell(\bar{y}_0, \AGG^{\ell}(\{\!\{ \bar{y}_1,\ldots,\bar{y}_m\}\!\}))||_\infty < \frac{\varepsilon_\ell}{2}.
    \end{align*}
    We define $I_{\ell-1} = \bigcup_{\bar{x}\in \chi_n^{\ell-1}} \{\bar{y}\in \mathbb{R}^{\delta^{\ell-1}}\mid ||\bar{x}-\bar{y}||_\infty \leq \varepsilon_{\ell-1}\}.$
    It contains all vectors that have a distance of at most $\varepsilon_{\ell-1}$ to some vector in $\chi_n^{\ell-1}$.
    By the Heine-Borel theorem and since $I_{\ell-1}$ is bounded and closed, $I_{\ell-1}$ is compact.
    We define $J_{\ell-1}$ similarly, but with the aggregation function in mind.
    Let $$J_{\ell - 1} = \bigcup_{0\leq i \leq n}\AGG^\ell\{\!\{\bar{y}_1,\ldots,\bar{y}_i\mid \forall k\in[i]\!: \bar{y}_k\in I_{\ell-1}\}\!\}.$$
    $J_{\ell-1}$ is a compact set, because $(I_{\ell-1})^i$ is compact for all $i\in [n]$, $\AGG^\ell$ is a continuous function when restricted to multisets of size $i$ and because continuous functions preserve compact sets.

    Because $f$ has the universal approximation property, there exist a dimension $\gamma^\ell$ and matrices
    $W_1^\ell\in \mathbb{R}^{2\delta^{\ell-1}\times \gamma^\ell}$ and $W_2^\ell\in\mathbb{R}^{\gamma^\ell \times \delta^\ell}$ and a vector $\bar{b}^\ell\in\mathbb{R}^{\gamma^\ell}$  such that for all
    ${\bar{x}\in I_{\ell-1}}$ and $\bar{y}\in J_{\ell-1}$ we have
    \begin{equation*}
    ||\COM^\ell(\bar{x}, \bar{y}) 
    - f((\bar{x}\circ \bar{y})W_1^\ell +
    \bar{b}^\ell)W_2^\ell||_\infty < \frac{\varepsilon_\ell}{2},
  \tag{$*$}
  \end{equation*}
   where
    $\circ$ denotes vector concatenation.
    We choose $\varepsilon_{\ell-1}$ as the desired error bound of
    layer $\ell-1$.

    \smallskip
    
    We now use the approximations identified above to define a \ac{gnn} $\mathcal{G}'$.
    To obtain a simple GNN, we split each $W_1^\ell\in
    \mathbb{R}^{2\delta^{\ell-1}\times \gamma^\ell}$ into two halves
    $C^\ell,A^\ell\in\mathbb{R}^{\delta^{\ell-1}\times \gamma^\ell}$ such that
    $\bar{x}C^\ell + \bar{y}A^\ell = (\bar{x}\circ\bar{y})W_1^\ell$
    for all $\bar{x},\bar{y}\in\mathbb{R}^{\delta^{\ell-1}}$. Note that in ($*$),
    $W^\ell_2$
    is multiplied outside of $f$. We thus have to defer this
    multiplication
    to the subsequent layer.
    %
    For $1\leq \ell \leq L+1$, the $\ell$-th layer computes
    $$ {\COM'}^\ell(\bar{x}_v,\bar{x}_a) = f(\bar{x}_v\cdot W_2^{\ell-1}\cdot C^\ell+\bar{x}_a\cdot W_2^{\ell-1}\cdot A^\ell +\bar{b}^\ell),$$
    where $W_2^0$ is the identity matrix of appropriate size.
    The $L+2$-nd layer computes
    $${\COM'}^{L+2}(\bar{x}_v,\bar{x}_a) = f(\bar{x}_v W_2^{L+1}).$$
    Let $c = \inf f((b-\varepsilon, b+\varepsilon))$ and let $\sim$ be $>$ if ${c\in f((a-\varepsilon, a+\varepsilon))}$ and $\geq$ otherwise.
    As the classification function, we use
        $$\CLS_2(\bar{x}) = \begin{cases}
            1 & \text{ if } x_L\sim c\\
            0 & \text{otherwise}.
        \end{cases}$$
        We define $$\mathcal{G'} = (L+2, \{{\COM'}^\ell\}_{\ell\in[L+2]}, \{\AGG^\ell\}_{\ell\in[L+2]}, \CLS_2).$$
        Since the last layer ignores the aggregation vector, we can choose ${\AGG}^{L+2}$ arbitrarily.
%
%
\\[2mm]
{\bf Claim.} $\mathcal{G}'$ is equivalent to $\mathcal{H}$ (and thus
also to $\mathcal{G}$) on graphs of size at most $n$.
\\[2mm]
Let $G$ be a graph of size at most $n$.
For each vertex $v\in V^G$, we use $\bar{x}_{v}^\ell$ and
$\bar{y}_v^\ell$ to denote the feature vectors computed by
$\mathcal{H}$ and $\mathcal{G}'$ at $v$ in layer $\ell$, respectively.
By definition, $\bar{x}_v^\ell\in \chi_n^\ell$ for all $v \in V(G)$.
Let $$\APR^\ell(\bar{x}_v, \bar{x}_a) =f((\bar{x}_v\circ \bar{x}_a)W_1^\ell+\bar{b}^\ell)W_2^\ell$$ be the approximation of $\COM^\ell(\bar{x}_v, \bar{x}_a)$. 
We have to show that for each $0\leq \ell\leq L+1$, $$||\bar{x}_v^\ell - \bar{y}_v^\ell W_2^\ell ||_\infty < \varepsilon_{\ell}.$$
For $\ell = 0$ this statement holds, since $\bar{y}^\ell_v = \bar{x}^\ell_v$ and $W_2^0$ is the identity matrix.
For the induction step, we observe that
\begin{align*}
    \bar{y}^{\ell}_vW_2^\ell&\\
    =&f(\bar{y}_v^{\ell-1}W_2^{\ell-1}C^\ell+\AGG\{\!\{\bar{y}^{\ell-1}_u\}\!\}W_2^{\ell-1}A^\ell+\bar{b}^\ell)W_2^\ell \\
    =&f(\bar{y}_v^{\ell-1}W_2^{\ell-1}C^\ell+\AGG\{\!\{\bar{y}^{\ell-1}_uW_2^{\ell-1}\}\!\}A^\ell+\bar{b}^\ell)W_2^\ell\\
    =&f(((\bar{y}_v^{\ell-1}W_2^{\ell-1})\circ \AGG\{\!\{\bar{y}^{\ell-1}_uW_2^{\ell-1}\}\!\})W_1^\ell+\bar{b}^\ell)W_2^\ell\\ 
    =&\APR^\ell(\bar{y}_v^{\ell-1}W_2^{\ell-1},\AGG\{\!\{\bar{y}^{\ell-1}_uW_2^{\ell-1}\}\!\})
\end{align*}
where the second equation exploits that $\AGG$ commutes with matrix multiplication.
%
%

By induction hypothesis, $||\bar{x}_w^{\ell-1}- \bar{y}_w^{\ell-1}W_2^{\ell-1}||_\infty < \varepsilon_{\ell-1}$ for all vertices $w\in V^G$, and therefore
\begin{align*}
        ||&\COM^\ell(\bar{x}_v^{\ell-1}, \AGG^{\ell}(\{\!\{ \bar{x}_u^{\ell-1}\}\!\})) \\
        &- \COM^\ell(\bar{y}_v^{\ell-1} W_2^{\ell-1}, \AGG^{\ell}(\{\!\{ \bar{y}_u^{\ell-1}W_2^{\ell-1}\}\!\}))||_\infty < \frac{\varepsilon_\ell}{2}.
    \end{align*}

Using $\APR^\ell$ instead of $\COM^\ell$ adds a second error 
 which is
bounded by  $\frac{\varepsilon_{\ell}}{2}$, since all $\bar{y}_wW_2^{\ell-1}$ are in $I_{\ell-1}$ and $\AGG^\ell(\{\!\{\bar{y}_u^{\ell-1}W_2^{\ell-1}\}\!\})\in J_{\ell-1}$.
Therefore, \begin{align*} 
    &||\bar{x}^{\ell}_v- \bar{y}^{\ell}_vW_2^{\ell} ||_\infty \\ 
    =& ||\bar{x}^\ell_v - \COM^\ell(\bar{y}_v^{\ell-1}W_2^{\ell-1}, \AGG^\ell(\{\!\{\bar{y}_u^{\ell-1}W_2^{\ell-1}\}\!\})) \\
    +& \COM^\ell(\bar{y}_v^{\ell-1}W_2^{\ell-1}, \AGG^\ell(\{\!\{\bar{y}_u^{\ell-1}W_2^{\ell-1}\}\!\}))- \bar{y}^{\ell}_vW_2^\ell ||_\infty\\
    \leq& ||\bar{x}^\ell_v - \COM^\ell(\bar{y}_v^{\ell-1}W_2^{\ell-1}, \AGG^\ell(\{\!\{\bar{y}_u^{\ell-1}W_2^{\ell-1}\}\!\}))||_\infty \\
    +&||\COM^\ell(\bar{y}_v^{\ell-1}W_2^{\ell-1}, \AGG^\ell(\{\!\{\bar{y}_u^{\ell-1}W_2^{\ell-1}\}\!\}))- \bar{y}^{\ell}_vW_2^\ell ||_\infty\\
    \leq&\ \varepsilon_{\ell}
\end{align*}
where the topmost inequality is due to the triangle inequality.

The last layer applies the last missing $W_2^{L+1}$ to $\bar{y}_v^{L+1}$.
If a pointed graph $(G,v)$ with size at most $n$ is accepted by $\mathcal{H}$, that is $x_v^{L+1} = b$, then $\bar{y}_v^{L+1}W_2^{L+1}\in(b-\varepsilon, b+\varepsilon)$, since $\varepsilon_{L+1} = \varepsilon$.
Therefore, $\CLS_2(\bar{y}^{L+2}_v) = 1$ and $\mathcal{G}'$ accepts $(G,v)$.
Likewise, if $(G,v)$ is not accepted by $\mathcal{H}$, then $x_v^{L+1} = a$ and $\bar{y}_v^{L+1}W_2^{L+1}\in(a-\varepsilon, a+\varepsilon)$.
Thus, $\CLS_2(\bar{y}^{L+2}_v) = 0$ and $\mathcal{G}'$ does not accept $(G,v)$.
Thus, the \ac{gnn} $\mathcal{G}'$ accepts the same pointed graphs as $\mathcal{H}$ and $\mathcal{G}$ up to size $n$.
\end{proof}

Theorem~\ref{theorem: rml-to-gnn-translation-nonuniform-univ-approx}
is now a straightforward consequence.

\lemmarmltognntranslationnonuniform*
\noindent
\begin{proof}\ 
    Let $n$ be a maximum graph size.
    Further let $f$ be a continuous, non-polynomial activation function. 
    By Lemma~\ref{lemma: continuous-non-poly-univ-approx}, $f$ has the universal approximation property.
    Since $f$ is not a polynomial, it is not constant, and since $f$ is continuous, there exist $a,b\in\mathbb{R}$ and $\varepsilon > 0$ such that for all $x\in(a-\varepsilon, a+\varepsilon)$ and $y\in(b-\varepsilon, b+\varepsilon)$ we have $f(x) < f(y)$.

    We now show Point~1 of Theorem~\ref{theorem: rml-to-gnn-translation-nonuniform-univ-approx}. 
    By Theorem~\ref{lemma: rml-to-gnn-translation-nonuniform}, for each
    formula $\varphi$ in \ac{rml} there exists a \simpleGNN
    $\mathcal{G}$ with truncated ReLU that is equivalent to $\varphi$
    on graphs of size at most $n$.
    Truncated ReLU is continuous, thus the combination functions in $\mathcal{G}$ are continuous.
    Since $\MEAN$ is bounded continuous and commutes with matrix multiplication,
    by Lemma~\ref{lemma: nonuniform-approximation-of-gnns} there is a \meanGNN that uses $f$ as its activation function and is equivalent to $\mathcal{G}$ on graphs up to size~$n$.

        Now for Point~2 of Theorem~\ref{theorem: rml-to-gnn-translation-nonuniform-univ-approx}. 
    It is shown in \cite{Barcelo2020} that for each formula $\psi$ in
    \ac{gml} there exists a simple \sumGNN $\mathcal{G}$ with
    truncated ReLU that is equivalent to $\psi$ (on all graphs).
   Since $\SUM$ is bounded continuous and commutes with matrix multiplication,
Lemma~\ref{lemma: nonuniform-approximation-of-gnns} yields a simple
\sumGNN that uses $f$ as its activation function and is equivalent to $\mathcal{G}$ on graphs up to size $n$.
\end{proof}

\subsection{Proof of Theorems~\ref{lem:MLtomaxGNN}, \ref{lemma:
    mean-gnn-to-rml-non-uniform-translation}, and~\ref{lem:frommaxgnntoML}}

\lemMLtomaxGNN*

\noindent
\begin{proof}\
   The proof is a minor variation of that of Theorem~\ref{lemma: rml-to-gnn-translation-nonuniform}, but simpler. In particular, 
   in the case $\varphi_k=\Diamond \varphi_i$,
   we can simply set $A_{i,k}=1$ and $\bar b_k=0$.
   Note that in contrast to Case~4 in the proof
   of Theorem~\ref{lemma: rml-to-gnn-translation-nonuniform} and in analogy with the proof in \cite{Barcelo2020}, the graph size is not used.
   One can then easily prove an analogue of 
   Claim~1 in the proof of Theorem~\ref{lemma: rml-to-gnn-translation-nonuniform}, without using the fact
   that the graph size is bounded. 
\end{proof}

\lemmameangnntormlnonuniformtranslation*
\noindent
\begin{proof}\ 
We prove Points~1 and~2 of Theorem~\ref{lemma: mean-gnn-to-rml-non-uniform-translation} simultaneously.
Let $$\mathcal{G}=(L,\{\AGG^{\ell}\}_{\ell \in [L]},\{\COM^{\ell}\}_{\ell \in [L]},\CLS)$$ be a GNN
on $\Pi$-labeled graphs, with $\Pi=\{P_1,\dots,P_r\}$, and let $n \geq 1$ be an upper bound on the size of
input graphs.
Let $\chi_n^\ell$ be the set of feature vectors $\bar{x}$ such that for some input graph $G$ of size at most $n$ and some vertex $v\in V^G$, $\mathcal{G}$ generates $\bar{x}$ at $v$ in layer $\ell$.
By Lemma~\ref{lemma: nonuniform-finitely-many-feature-vectors}, each $\chi_n^\ell$ is finite.
We construct one modal logic formula
$\varphi^\ell_{\bar x}$ for every layer $\ell$ and 
every $\bar x \in \chi^\ell_n$, proceeding by induction on $\ell$.

    For $\ell=0$, we can derive the desired formulas from the initial feature vector. For all $\bar x \in \chi^\ell_n$, set
    $$\varphi_{\bar{x}}^0 = \bigwedge_{\substack{1 \leq i \leq r \\[0.5mm] x_i = 0}} \lnot P_i\land\bigwedge_{\substack{1 \leq i \leq r \\[0.5mm] x_i = 1}}P_i.$$ 
    In the induction step, we enumerate all possibilities for $\mathcal{G}$ to generate $\bar{x}$. This is different depending on the
    aggregation function and logic we work with.
    We start with the case of \RML.
    Let $\chi^{\ell-1}_n=\{\bar{y}_1,\dots,\bar{y}_p \}$ and
     $$F= \left\{\frac{\ell}{m}\mid 1\leq \ell\leq m\leq n,\ \ell,m\in\mathbb{N}^+\right\}\cup \{0\}.$$
    For all $\bar y \in \chi^{\ell-1}_n$ and
    all $f_1,\dots,f_p \in F$ with $\sum_{i=1}^p f_i = 1$, define 
    $$\varphi_{\bar{y}, f_1,\ldots,f_p}^{\ell} = \varphi_{\bar{y}}^{\ell-1}\land \bigwedge_{i=1}^p \markDiamond^{={f_i}}\varphi_{\bar{y}_i}^{\ell-1}.$$ 
    For each $\bar x \in \chi^{\ell}_n$, we can now define the formula $\varphi_{\bar{x}}^\ell$ to be the disjunction of all formulas $\varphi_{\bar{y}, f_1,\ldots,f_p}^{\ell}$ such that $\bar{x} = \COM^\ell(\bar{y},\mn{MEAN}\, M)$ where $M$ is any finite multiset over $\chi^{\ell-1}_n$ that realizes the fractions  $f_1,\dots,f_p$, that is, $\frac{M(\bar y_i)}{|M|}=f_i$ for $1 \leq i \leq p$.
   
    %
    We can now show the following:
\\[2mm]
{\bf Claim 1.} For all $\Pi$-labeled graphs of size at most $n$, all \mbox{$v \in V^G$}, all layers $\ell$ of \Gmc, and each feature vector $\bar{x}\in\chi^\ell_n$:
$\mathcal{G}$ assigns $\bar{x}$ to $v$ in $G$ in layer $\ell$ if and only if $G,v \models \varphi_{\bar{x}}^{\ell}$.
\\[2mm]
This is in fact easy to prove by induction on $\ell$,
using the fact that the result of mean aggregation only depends on the fraction of successors at which each 
possible feature vector was computed by the previous layer (and not on the exact number of such successors).

\smallskip

The case of {\sumGNN}s is very similar, except that for sum aggregation we need
to know the exact number of times that each
feature vector has been computed at some successor by the previous layer. Since the number of successor is bounded by the constant $n$, we can express all possibilities in GML. 
For all $\bar y \in \chi^{\ell-1}_n$ and
    all $m_1,\dots,m_p \in [n]\cup\{0\}$, define 
    $$\varphi_{\bar{y}, m_1,\ldots,m_p}^{\ell} = \varphi_{\bar{y}}^{\ell-1}\land \bigwedge_{i=1}^p \Diamond^{={m_i}}\varphi_{\bar{y}_i}^{\ell-1}.$$ 
    For each $\bar x \in \chi^{\ell}_n$, we then define the formula $\varphi_{\bar{x}}^\ell$ to be the disjunction of all formulas $\varphi_{\bar{y}, m_1,\ldots,m_p}^{\ell}$ such that $\bar{x} = \COM^\ell(\bar{y},\SUM\, M)$ where $M$ is the multiset over $\chi^{\ell-1}_n$ defined by 
    $M(\bar y_i)=m_i$ for $1 \leq i \leq p$. The rest of the proof remains unchanged.
 %
%
\end{proof}

\lemfrommaxgnntoML*
\noindent
\begin{proof}\
  The proof is analogous to that of
  Lemma~\ref{lemma: mean-gnn-to-rml-non-uniform-translation}, with two differences. First, finiteness of
  the relevant sets of feature vectors 
  holds already without imposing a constant bound on the graph size. And
  second, we once more have to adapt the
  formulas $\varphi_{\bar{y}, f_1,\ldots,f_p}^{\ell}$.
\\[2mm]
{\bf Claim.}
    Let $\mathcal{G}$ be a \maxGNN with $L$ layers over some finite set of vertex labels $\Pi$.
    For each layer $\ell$ of output dimension $\delta^\ell$ let 
    $\chi^\ell\subseteq \mathbb{R}^{\delta^\ell}$ denote the set of feature vectors $\bar x$ such that
    for some input graph $G$ and some vertex $v$
    in $G$, \Gmc generates $\bar x$ at $v$ in layer $\ell$.
    Then 
    $\chi^{\ell}$ is finite.    
\\[2mm]
\emph{Proof of claim}. The proof is straightforward by induction on~$\ell$.
$\chi^0$ is the set of all vectors over the set $\{0,1\}$. Then the set $\chi^\ell$ 
consists of all vectors $\bar x$ that can be
obtained by choosing a vector $\bar y \in 
\chi^{\ell-1}$ and a set $S \subseteq \chi^{\ell-1}$ and setting $\bar x = \COM^\ell(\bar{y},\MAX\, S)$. Clearly,
this implies that $\chi^\ell$ is finite. The crucial difference to the sum and mean aggregation cases is that we can work with a set here rather than with multisets since max aggregation
is oblivious to multiplicities.

\medskip
  
  Now for the construction of the \MoL formulas $\varphi_{\bar{x}}^\ell$.
  In the case of {\maxGNN}s, the result of aggregation only depends on which feature
vectors have been computed at some successor by the previous layer, and which have not.
We can thus use the exact same arguments as in the proof of Lemma~\ref{lemma: mean-gnn-to-rml-non-uniform-translation}, except that the formulas 
$\varphi_{\bar{y}, f_1,\ldots,f_p}^{\ell}$ are replaced as follows. For all 
$\bar y \in \chi^{\ell-1}$ and
    all subsets $S \subseteq \chi^{\ell-1}$, define 
    $$\varphi_{\bar{y},S}  = \varphi_{\bar{y}}^{\ell-1}\land \bigwedge_{\bar z \in S} \Diamond \varphi_{\bar{z}}^{\ell-1} \wedge
    \bigwedge_{\bar z \in \chi^{\ell-1} \setminus S} \Box \neg \varphi_{\bar{z}}^{\ell-1}.$$ 
 For each $\bar x \in \chi^{\ell}$, we then define the formula $\varphi_{\bar{x}}^\ell$ to be the disjunction of all formulas $\varphi_{\bar{y},S}$ such that $\bar{x} = \COM^\ell(\bar{y},\MAX\, S)$. The rest of the proof remains unchanged.
\end{proof}

\section{Proofs for Section~\ref{sect:uniform}}

\lemtoAFMLstepone*
\noindent
\begin{proof}\
   It is well-known
that any property $P$ definable by a GNN,
independently of the aggregation,
combination, and classification function used, is invariant
under graded bisimulation, that is,
if $(G_1,v_1) \in P$ and 
$(G_1,v_1)$ and $(G_2,v_2)$ are graded
bisimilar, then \mbox{$(G_2,v_2) \in P$}. We
refer to \cite{Barcelo2020} for a proof 
 and also for a detailed definition of graded bisimulations. It was further proved 
in \cite{Otto2019} that every FO formula in one free variable that is invariant under graded bisimulation (on finite models) is equivalent to a \GML formula.
Together with Lemma~\ref{lem:FromFOtoMSO}, this clearly implies the statement in the theorem.
\end{proof}

\lemmameangnntomlwinningstrategies*
\noindent
\begin{proof}\ 
%
%
    The proof is by induction on $\ell$.
    The statement is obviously true for $\ell=0$ 
    because $(v_i,k_i)$ satisfies the same propositional variables as $v_i$. 
    Therefore, because the winning condition for Spoiler is not satisfied in $\EF{}{0}{\MoL}{G_1,v_1,G_2,v_2}$, it is also not satisfied in $$\EF{}{0}{\GML[c]}{c\cdot G_1, (v_1,k_1), c\cdot G_2, (v_2,k_2)}.$$ 
%
    For the induction step, let the statement hold for $\ell-1$ and assume that $D$ has a winning strategy in $\EF{}{\ell}{\MoL}{G_1,v_1,G_2,v_2}$.
    Then we find a winning strategy for $D$ in $$\EF{}{\ell}{\GML[c]}{c\cdot G_1, (v_1,k_1), c\cdot G_2, (v_2,k_2)},$$ as follows.
    In the first round:
    \begin{itemize}
    
        \item Assume that in Step~1 of the first round $S$ selects $i \in \{1,2\}$ and $U_i=\{(u_i^1,a_1),\ldots,(u_i^n,a_n)\}\subseteq \mathcal{N}_{G_i}(v_i,k_i)$, with \mbox{$n\leq c$}.
        In the game $\EF{}{\ell}{\MoL}{G_1,v_1,G_2,v_2}$, $S$ may choose
        the same $i$ and any of the vertices $u_i^1,\dots,u_i^n$
        (or other elements of $\mathcal{N}_{G_i}(v_i)$) and for each  choice $u_i^j$
        the winning strategy for $D$ provides an answer
         $u_{3-i}^j\in \mathcal{N}_{G_{3-i}}(v_{3-i})$.
        In Step~2, we let $D$ 
        choose $U_{3-i}=\{(u_{3-i}^1,1),\ldots (u_{3-i}^n,n)\}\subseteq \mathcal{N}_{G_{3-i}}(v_{3-i},k_{3-i})$, which satisfies $|U_1| = |U_2|$.
        
        \item Assume that $S$ chooses $(u_{3-i}^j,j)$ in Step~3.
        Then $D$  chooses $(u_i^j,a_j)$. 
        
    \end{itemize}
    Since $D$ has a winning strategy
    in $\EF{}{\ell}{\MoL}{G_1,v_1,G_2,v_2}$, $D$~also
     has a winning strategy in $\EF{}{\ell-1}{\MoL}{G_1,u_1^j,G_2,u_2^j}$.  
    By the induction hypothesis, $D$ thus has a winning strategy in $$\EF{}{\ell-1}{\GML[c]}{c\cdot G_1, (u_1^j, a_1^j), c\cdot G_2, (u_2^j, a_2^j)},$$ 
    where $a_i^j = a_j$ and $a_{3-i}^j = j$.
    The remaining rounds are played according to that strategy.
\end{proof}
\lemmarmltognntranslation*
\noindent
\begin{proof}\ 
%
  We use the
    following step function as the activation function:
    $$f(x) = \begin{cases}
        1&\text{ if } x > 0\\
        0&\text{ otherwise}.
    \end{cases}$$


 The proof is then a minor variation of that of Theorem~\ref{lemma: rml-to-gnn-translation-nonuniform}, but simpler.
    We can assume that each diamond uses $>$ since $\markDiamond^{\geq t} \varphi\equiv \neg \markDiamond^{> (1-t)}\neg\varphi$.
 We only give the relevant weights:
    \begin{enumerate}[label=\textit{Case }\arabic*:, ref=\arabic*, leftmargin=*]
        \item $\varphi_k = P_k$. Set $C_{k,k} = 1$.
        \item \label{case: lnot-non-continuous}$\varphi_k = \lnot \varphi_i$. Set $C_{i,k} = -1$ and let $b_k = 1$.
        \item $\varphi_k = \varphi_i\lor\varphi_j$. Set $C_{i,k} = C_{j,k} = 1$.
        \item \label{case: diamond-non-continuous}$\varphi_k = \markDiamond^{> t}\varphi_i$. Set $A_{i,k} = 1$ and let $b_k = -t$.
    \end{enumerate}
%
    %
As the classification function (as in  Theorem~\ref{lemma: rml-to-gnn-translation-nonuniform}) we use
    $$\CLS(\bar{x}) = \begin{cases}
        1 & \text{ if } x_L>0\\
        0 & \text{otherwise}.
    \end{cases}$$
   It is not difficult to show correctness of the translation:
\\[2mm]
{\bf Claim.} For all $\varphi_k$, $1 \leq k \leq L$, the following holds:
if $v \in V^G$ and $k\leq k' \leq L$, then $$(\bar{x}_{G,v}^{k'})_{k} = \begin{cases}
    1&\text{ if } G,v\models \varphi_k,\\
    0&\text{ otherwise.}
\end{cases}$$
We leave details to the reader.
\end{proof}

\section{Proofs for Section \ref{sect:unif_reloaded}}

\subsection{Proof of Theorem~\ref{thm:crazyclass}}

We show that for each \ac{ml} formula there exists an
equivalent \meanGNN with continuous combination functions
and classification function.
$$\CLS(x) = \begin{cases} 1&\text{ if }
  x\in\mathbb{R}\setminus\mathbb{Q}\\0 &\text{otherwise.}
\end{cases}$$ In this translation, we represent truth of subformulas by
irrational numbers and falsity by rational numbers. We choose
$a\in\mathbb{Q}$ and $b\in\mathbb{R}\setminus\mathbb{Q}$ arbitrarily
to represent the falsity and truth of atomic formulas. For non-atomic
formulas, we are not able to maintain exactly this
representation. Instead
we represent falsity by rational numbers from the interval $I$ between
$a$ and $b$ and truth by values from the set 
$$\{ p+qb\mid p,q\in\mathbb{Q}, q>0\}\cap I.$$
We will see that this
encoding supports the implementation of disjunction and modal diamonds
in a natural way. Implementing negation, however, is less
straightforward.
We will rely on the following lemma.
%
\begin{lemma}\label{lemma: exchange_rational_irrational}
    Let $a\in\mathbb{Q}$ and $b\in\mathbb{R}\setminus\mathbb{Q}$ with $a<b$.\footnote{The condition $a<b$ is not necessary, but it simplifies our exposition.} Let $I = [a,b]$ be the interval between $a$ and $b$. 
    Let $F = I\cap \mathbb{Q}$ and $T = \{p+qb\mid p,q\in\mathbb{Q},\ q>0\}\cap I $.
   Then there exists a continuous function $f:\mathbb{R}\to\mathbb{R}$ such that $f(F) = T$ and $f(T) = F$.
\end{lemma}

\noindent
\begin{proof}\ We first construct a function with domain and range
  $F\cup T$ that swaps $F$ and $T$. After defining the image of $a$ and $b$,
  this proof extends Cantor's isomorphism theorem, stating that all countable, dense,
  linearly ordered sets without maximum and minimum are isomorphic
  \cite{Cantor1895}.  We then show that this function is continuous on
  its domain and that it can be extended to a continuous function on
  $\mathbb{R}$.

  Since $F\setminus\{a\}$ and $T\setminus\{b\}$ are countable, there exist enumerations
  $a_1,a_2,\ldots$ and $b_1,b_2,\ldots$ of $F\setminus\{a\}$ and $T\setminus\{b\}$,
  respectively. Consider the enumeration
  $x_1 = a_1,x_2 = b_1,x_3 = a_2, x_4=b_2, \ldots$ of $F\cup T\setminus\{a,b\}$.  We
  define a sequence $f_0,f_1,\dots$ of partial functions from $F\cup
  T$ and obtain the desired function $f$ in the limit.
  Let
  $f_0 = \{(a,b),(b,a)\}$.  For $f_{i+1}$, let $k$ be the
  smallest number such that $x_k$ is not in the domain of $f_i$.  If
  $x_k = a_{k'}$, then choose the smallest $j$ such that
    \begin{itemize}
        \item $b_j$ is not in the domain of $f_i$, and
        \item for all $x$ in the domain of $f_i$:
          \begin{itemize}
            \item if $a_{k'} < x$, then $b_j > f_i(x)$ and
            \item if $a_{k'} > x$, then $b_j < f_i(x)$.
      \end{itemize}
    \end{itemize}
    Since $T$ is dense, such $b_j$ exists.
    We then define $$f_{i+1} = f_i \cup \{(a_{k'}, b_j), (b_j, a_{k'})\}$$
    and proceed analogously if $x_k = b_{k'}$.
    Finally define $$f = \bigcup_{i\in\mathbb{N}} f_i.$$
    We observe the following. Note that Point~2 is the crucial
    property
    that we want to attain.

    \smallskip
    \noindent\textbf{Claim 1. }
    \begin{enumerate}
    \item $f$ is a bijection from $F\cup T$ to itself;
     \item $f(F) = T$ and $f(T) = F$. 
    \end{enumerate}
    We start with Point~1. It is clear from the definition that $f$ is a function. Also by
    definition, the domain of each $f_i$ is identical to its range.
    It follows that each $f_i$ is injective, and thus so is $f$. It is
    also surjective: we eventually choose any element of
    $F\cup T$, and if this happens during the definition
    of $f_i$, then the element is in the range of $f_i$, thus of~$f$.

    Point~2 easily follows from Point~1 and the definition of the
    functions
    $f_i$, which only map elements of $F$ to elements
    of $T$, and vice versa.
This finishes the proof of Claim~1.

    \medskip

    To define the extension of $f$ to domain $\mathbb{R}$, we first analyze the images of convergent sequences.
\\[2mm]
    \textbf{Claim 2.} 
        \begin{enumerate} 

        \item $f$ reverses order: if $x,y\in F\cup T$  with $x<y$, then $f(y)<f(x)$;

        \item if $(x_i)_{i\in\mathbb{N}^+}$ and
          $(y_i)_{i\in\mathbb{N}^+}$ are sequences over $F\cup T$ that converge in $\mathbb{R}$ to the same limit and
          such that $(f(x_i))_{i\in\mathbb{N}^+}$ and
          $(f(y_i))_{i\in\mathbb{N}^+}$ converge in $\mathbb{R}$, then
          $(f(x_i))_{i\in\mathbb{N}^+}$ and $(f(y_i))_{i\in\mathbb{N}^+}$ converge to the same limit.

        \item if $(x_i)_{i\in\mathbb{N}^+}$ is a sequence over $F\cup T$ that converges in $\mathbb{R}$, then
          $(f(x_i))_{i\in \mathbb{N}^+}$  also converges in
          $\mathbb{R}$;

          
    \end{enumerate}

    We prove Point~1 by induction on $i$.
    It holds for $f_0$, since $f_0(a) = b$ and $f_0(b) = a$.
    Assume that Point~1 holds for $f_{i-1}$ and let $f_{i} = f_{i-1}\cup \{(c,d),(d,c)\}$, where $c$ the first element in $x_1,x_2,\ldots$ that is not in the domain of $f_{i-1}$.
    For all $x$ in the domain of $f_{i-1}$ with $c< x$, we have $f_{i} (x)= f_{i-1}(x) < d = f_{i}(c)$ by choice of $d$.
    The analogous statement hold if $c > x$.
    We also have to show this point for $d$. So let $x$ be any element in the domain of $f_{i-1}$.
    Since the domain of $f_{i-1}$ is equal to its range, there exists a $y$ such that $x = f_{i-1}(y)$.
    If $d < x = f_{i-1}(y)$, then $c\geq y$ by definition of $f_{i}$. 
    Moreover, $c > y$ since $c$ is a fresh element.
    Therefore, $f_{i}(d) = c > y = f_{i}(x)$ 
    The analogous statement hold if $d > x$. 
    This point also holds if $x,y\in \{c,d\}$ and if $x,y$ are both in the domain of $f_{i-1}$.

    For Point~2, let $(x_i)_{i\in\mathbb{N}^+}$ and $(y_i)_{i\in\mathbb{N}^+}$, $x_i,y_i\in F\cup T$, be two sequences that converge to the same limit. 
    Let ${x = \lim_{i\to\infty} f(x_i)}$ and $y = \lim_{i\to\infty} f(y_i)$.
    Assume to the contrary that $x\neq y$. W.l.o.g. let $x < y$.
    Let $d_1,d_2\in F\cup T$ such that $x < d_1<d_2<y$. Such $d_1$ and $d_2$ exists since $F\cup T$ is dense in the interval between $a$ and $b$.
    Let $c_1$ and $c_2$ such that $f(c_i) = d_i$ for $i\in \{1,2\}$.
    Then there exists an index $N$ such that for all $n\geq N$ we have $f(x_n) <d_1<d_2< f(y_n)$.
    By Point~1, $y_n <c_1 < c_2 <  x_n$ for each $n\geq N$.
    Hence, $\lim_{i\to\infty} y_i \leq c_1 < c_2 \leq \lim_{i\to\infty} x_i$.
    This is a contradiction to $\lim_{i\to\infty} y_i = \lim_{i\to\infty} x_i$.
    Thus, $f(x_i)$ and $f(y_i)$ converge to the same limit.

    To prove Point~3, let $(x_i)_{i\in\mathbb{N}^+}$, $x_i\in F\cup T$, be a sequence that converges to some value $x\in \mathbb{R}$.
    We split $(x_i)_{i\in\mathbb{N}^+}$ into the subsequence $(z_i^>)$ containing all elements greater than $x$, the subsequence $(z^<_{i})$ containing all elements less than $x$ and the subsequence $(z_{i}^=)$ containing all elements equal to $x$.
    At least one of these is infinite and the limit of the infinite ones is $x$.
    We now show for all $\sim\ \in \{>,<,=\}$ that if $(z_{i}^\sim)$ is infinite, then $(f(z_{i}^\sim))_{i\in\mathbb{N}^+}$ converges.

    If $(z_{i}^=)$ is infinite, then $f(z_{i}^=)$ converges to $f(x)$ since it is constant.
    We now show the statement for $(z_{i}^<)$. The proof for $(z_i^>)$ is analogous.
    Since $(z_i^<)_{i\in\mathbb{N}^+}$ is bounded by $a$ and $b$ and by the Bolzano-Weierstrass theorem, there exists a monotone subsequence $(z_{\ell_i}^<)_{i\in\mathbb{N}}$ which is also convergent to $x\in\mathbb{R}$.
    Since all $z_{\ell_i}^< < x$, this sequence has to be monotone increasing.
    By Point~1, $(f(z_{\ell_i}^<))_{i\in\mathbb{N}^+}$ is a monotone decreasing subsequence of $(f(z_{i}^<))_{i\in\mathbb{N}^+}$ which is bounded by $a$ and $b$.
    Therefore, this subsequence is convergent to some $y\in\mathbb{R}$.
    We now show that $(f(z_i^<))_{i\in\mathbb{N}^+} $ also converges to $y$.
    Let $\varepsilon > 0$.
    Then there exists an $N$ such that $|f(z_{\ell_n}^<)-y| < \varepsilon$ for all $n\geq N$.
    Since $(z_i^<)_{i\in\mathbb{N}^+}$ converges to $x$ there has to be an $M$ such that $z_{\ell_N}^< \leq z_m^< < x$ for all $m\geq M$.
    For each $m \geq M$ we can find an index $m'$ such that $z_{m'}^<$ is part of $(z_{\ell_i}^<)_{i\in\mathbb{N}^+}$ and is closer to $x$, that is $z_{\ell_N}^<\leq z_m^<\leq z_{m'}^< < x$.
    By Point~1, $f(z_{m'}^<) < f(z_m^<) \leq f(z_{\ell_N})$.
    And since $(f(z_i^<))_{i\in\mathbb{N}^+}$ is monotone decreasing, we have $y\leq f(z_{m'}^<) $ and thus $|f(z_m^<)-y| \leq |f(z_{\ell_N}^<)-y| <\varepsilon$.
    Thus, $(f(z_i^<))_{i\in\mathbb{N}^+}$ converges to $y$.

   It follows from Point~2 that if two or all three of $(f(z_i^<))$, $(f(z_i^>))$ and $(f(z_i^=))$ are infinite, then they converge to the same limit~$y$.
    We show that if all three are infinite, then $(f(x_i))_{i\in\mathbb{N}^+}$ also converges.
    If only two of them converge, we can argue analogously.
    If all three are infinite, then for each $\varepsilon > 0$ we find indices $N^<, N^>, N^=$ and $N = \max(N^<,N^>,N^=)$ such that for all $n\geq N$ and $\sim\ \in \{<,>,=\}$, $|f(z_n^\sim)-y| < \varepsilon$.
    Hence, we can find an $N'$ such that for all $n\geq N'$ we have
    $|f(x_n)-y| < \varepsilon$.
    This finishes the proof of Claim~2.

    \medskip

   We now define a continuous extension $\hat{f}$ of $f$ to 
    domain~$\mathbb{R}$. We remind the reader that $a < b$. 
    
    Start with setting 

    \begin{itemize}
      
    \item $\hat{f}(x) = b$ for all $x < a$;

    \item $\hat{f}(x) = a$ for all $x > b$;

    \end{itemize}
    Since $f(a) = b$ and $f(b) = a$, it immediately follows that $\hat{f}$ is continuous at all those $x$.

    It remains to define $\hat{f}$ for all $x \in I = (a,b)$.  Since
    $(F\cup T)\setminus\{a,b\}$ is dense in $I$, there exists a
    sequence $(x_i)_{i\in\mathbb{N}^+}$ with limit~$x$ such that
    $x_i\in (F\cup T)\setminus\{a,b\}$ for all $i\in\mathbb{N}^+$.  We
    define $$\hat{f}(x) = \lim_{i\to\infty} f(x_i).$$ This limit
    exists by Claim~2.3 and is well-defined by Claim~2.2. 
    $\hat{f}$ is an extension of $f$, that is $\hat{f}(x) = f(x)$ for all $x\in F\cup T$, because $f(x) = \lim_{i\to\infty} f(x) = \hat{f}(x)$.


    We have to
    show that $\hat f$ is continuous at all elements of
    $[a,b]$.



    
    We first observe that the following is an easy consequence of
    Claim~2.1 and the definition of $\hat{f}$.

    \smallskip
    \noindent
    \textbf{Claim 3.} $\hat{f}$ reverses order: if $x<
    y$ then $\hat{f}(y)\leq \hat{f}(x)$. 

    \smallskip 
    \noindent\textbf{Claim 4.} $\hat{f}$ is continuous on $[a,b]$.

    \smallskip 
    \noindent
    We show the claim for all $x\in I$ first.
    Let $(x_i)_{i\in\mathbb{N}^+}$ be a sequence with limit $x$.
    We have to show that $\hat{f}(x) = \lim_{i\to\infty}\hat{f}(x_i)$.

    Since $x\in I$ and $I$ is an open interval, there exists an index
    $N$ such that  $x_n\in I$ for all $n\geq N$. 
    Thus, we can assume w.l.o.g that every $x_i$ is in $I$.
    Since $I$ is an open interval and $(F\cup T)\setminus\{a,b\}$ is dense in $I$, there exist upper and lower bounds of $(x_i)_{i\in\mathbb{N}^+}$ in $(F\cup T)\setminus\{a,b\}$.
    That is, there exist sequences $(y_i)_{i\in\mathbb{N}^+}$ and
    $(z_i)_{i\in\mathbb{N}^+}$ such that
 $y_i, z_i\in (F\cup T)\setminus\{a,b\}$, both converge to $x$, and $y_i \leq x_i\leq z_i$ for all $i\in\mathbb{N}^+$.
    By Claim~3, the image of $(x_i)_{i\in\mathbb{N}^+}$ is also bounded by the images of $(y_i)_{i\in\mathbb{N}^+}$ and $(z_i)_{i\in\mathbb{N}^+}$.
    Since the order is reversed, $f(z_i) = \hat{f}(z_i) \leq \hat{f}(x_i)\leq \hat{f}(y_i) = f(y_i)$.
    By definition of $\hat{f}$, we have $\hat{f}(x) = \lim_{i\to\infty} f(z_i) = \lim_{i\to\infty} f(y_i)$, and by the squeeze theorem   
    we also have $\hat{f}(x) = \lim_{i\to\infty} \hat{f}(x_i)$.
    It remains to argue that $\hat{f}$ is continuous at $a$ and $b$.
    We show that $\hat{f}$ is continuous at $a$; the argument that $\hat{f}$ is continuous at $b$ is analogous.
    Let $(x_i)_{i\in\mathbb{N}^+}$ be a sequence with limit $a$.
    We have to show that $\lim_{i\to\infty} f(x_i) = b$.
    There can only be finitely many $i$ such that $x_i> b$, hence these indices can be ignored without changing convergence.
    We can also safely ignore all $x_i < a$, since they satisfy $\hat{f}(x_i) = b$.
    If there are infinitely many $i$ left such that $x_i\in I$, we can argue analogously to the previous case $x\in I$ to prove $\lim_{i\to\infty} \hat{f}(x_i) = b$.
    If there are only finitely many $i$ such that $x_i\in I$, then all $x_i < a$ define the limit of $(f(x_i))_{i\in\mathbb{N}^+}$ which then has to be $b$.
%

\end{proof}

\begin{theorem}
  \label{thm:sglfahgdsgf}
    In the uniform setting when we allow arbitrary classification functions:
    $$\MoL\subseteq \conmeanGNN.$$ 
\end{theorem}

\noindent
\begin{proof}\
    Let $\varphi$ be an  \ac{ml}  formula over a finite set $\Pi = \{P_1,\ldots, P_r\}$ of vertex labels.
    Let $\varphi_1,\ldots,\varphi_L$ be an enumeration of the subformulas of $\varphi$ such that (i)~$\varphi_i = P_i$ for $1\leq i\leq r$ and (ii)~if $\varphi_\ell$ is a subformula of $\varphi_k$ then $\ell < k$.

    We construct a GNN \Gmc with $L+1$ layers, all of output dimension
    $L$.     Choose a rational number $a$ and an irrational number $b$ such that $a<b$, and let 
    $f$ be the function constructed for these choices of $a$ and $b$ in Lemma~\ref{lemma: 
      exchange_rational_irrational}. 
    The purpose of the first
    layer is to convert the initial feature vector into the intended
    format where truth is represented by irrational numbers and
    falsity
    by rational numbers, as explained at the beginning of this
    section.    For the first layer, the $k$-th entry of
    $\COM^1(\bar{x}, \bar{y})$ is thus defined as follows:

    \begin{enumerate}[label=\textit{Case }\arabic*:, ref=\arabic*, leftmargin=*]
        \item $\varphi_k = P_k$. Return $(b-a)x_k+a$
       \item Otherwise return $a$.
    \end{enumerate} 
    Since $(b-a)x_k+a$ is continuous, $\COM^1$ is continuous. 

    For all layers $\ell \geq 2$, we use a different combination function. 
    The $k$-th entry of $\COM^\ell(\bar{x},\bar{y})$ is defined as follows:
    \begin{enumerate}[label=\textit{Case }\arabic*:, ref=\arabic*, leftmargin=*]
        \item $\varphi_k = P_k$. Return $x_k$.
        \item $\varphi_k = \neg\varphi_i$. Return $f(x_i)$.
        \item $\varphi_k = \varphi_i\lor\varphi_j$. Return $\frac{x_i+x_j}{2}$.
        \item $\varphi_k = \Diamond \varphi_i$. Return $y_i$.
    \end{enumerate}
    Since each of these functions is continuous, $\COM^\ell$ is continuous.
    $\mathcal{G}$ uses the following classification function:
    $$\CLS(x) = \begin{cases}
        1&\text{ if } x\in\mathbb{R}\setminus \mathbb{Q},\\
        0&\text{ otherwise.}
    \end{cases}$$

Let $I = [a,b]$, $F = \mathbb{Q}\cap I$ and $T =
\{p+qb\mid p,q\in\mathbb{Q}, q> 0\}\cap I$.     We now show the correctness of our construction. 
    \\[2mm]
    \textbf{Claim.} 
    For all $\varphi_k$, $1\leq k\leq L$, the following holds: if $v\in V^G$ and $k\leq k'\leq L$, then 
    \begin{align*}
    (\bar{x}_{G,v}^{k'})_k&\in T \text{ if } G,v\models\varphi_k, \text{ and }\\
    (\bar{x}_{G,v}^{k'})_k&\in F \text{ otherwise.}
    \end{align*}
    The proof of the claim is by induction on $k$.

    \smallskip
    \noindent 
    \emph{Case~1}. 
    For \mbox{$1\leq k\leq r$}, $\varphi_k = P_k$ is an atomic
    formula. If $G,v\models \varphi_k$, then
    \mbox{$(\bar{x}_{G,v}^0)_k = 1$}
    and the first layer returns $$(\bar{x}_{G,v}^1)_k = (b-a)\cdot 1+ a = b\in T.$$
    All subsequent layers $\ell$ return $(\bar{x}_{G,v}^\ell)_k = b\in T$.
    If $G,v\not\models\varphi_k$, then $(\bar{x}_{G,v}^0)_k = 0$,
    the first layer returns
    $$(\bar{x}_{G,v}^1)_k = (b-a)\cdot 0 + a = a\in F,$$
    and for all subsequent layers $\ell$, we have $(\bar{x}_{G,v}^\ell)_k = a\in F$.

      \smallskip 
    \noindent 
    \emph{Case~2}. Let $\varphi_k = \neg\varphi_i$. If
    $G,v\models\neg\varphi_{i}$, then since \mbox{$i < k$} the induction
    hypothesis yields $(\bar{x}_{G,v}^{k'-1})_i \in F$ and thus $(\bar{x}_{G,v}^{k'})_k\in T$ by Lemma~\ref{lemma: exchange_rational_irrational}.
    If $G,v\not\models\neg\varphi_i$, then $(\bar{x}_{G,v}^{k'-1})_i\in T$ and thus $(\bar{x}_{G,v}^{k'})_k\in F$.

      \smallskip 
    \noindent 
    \emph{Case~3}. Let $\varphi_k = \varphi_i \lor\varphi_j$. If $G,v\models\varphi_k$, then at least one of $(\bar{x}_{G,v}^{k'-1})_i$ and $(\bar{x}_{G,v}^{k'-1})_j$ is in $T$. 
    Since the definition of $T$ requires
    $q>0$,
   \begin{equation*}
    \frac{(\bar{x}_{G,v}^{k'-1})_i +
      (\bar{x}_{G,v}^{k'-1})_j}{2}
    \tag{$*$}
    \end{equation*}
    can be written as $p'+q'b$ where $p',q'\in\mathbb{Q}$ and $q' > 0$.
    If $G,v\not\models\varphi_k$, then both $(\bar{x}_{G,v}^{k'-1})_i$ and $(\bar{x}_{G,v}^{k'-1})_j$ are in $F$, thus both are rational numbers.
    Therefore, $(*)$ is also rational, and it is easy to verify that it is in $F$.

      \smallskip 
    \noindent 
    \emph{Case~4}. Let $\varphi_k = \Diamond\varphi_i$. First assume that $G,v\models \varphi_k$.
    Let $\mathcal{N}(G,v) = \{u_1,\ldots,u_n\}$.
    For $1\leq j\leq n$, there exist $p_j,q_j\in\mathbb{Q}$ such that $(\bar{x}_{G,u_j}^{k'-1})_i =p_j + q_jb$.
    Then
    \begin{align*}
        &\MEAN\{\!\{ (\bar{x}_{G,u_j}^{k'-1})_i \mid 1\leq j\leq n\}\!\} \\
        = \; & \MEAN\{\!\{p_j \mid 1\leq j\leq n\}\!\} + \MEAN\{\!\{
               q_j \mid 1\leq j\leq n\}\!\}b.
    \end{align*}
    By induction hypothesis, $q_j \geq 0$ for all $j$ and $q_j > 0$
    for at least one $j$.
    Thus, the above can be written as $p+qb$ where $p,q\in\mathbb{Q}$ and $q > 0$.
    Thus  $(\bar{x}_{G,v}^{k'})_k\in T$, as required.
    If $G,v\not\models\varphi_k$, then $q_j = 0$ for all $j$, and thus
    $(\bar{x}_{G,u_j}^{k'-1})_i$ is a rational number.
    It is easy to verify that it is also in $F$.
\end{proof}

\subsection{Relations of AFML to \ac{ef} games and \ac{ml}}
\begin{theorem} \label{lemma: ef-afml-equivalence}
Let $P$ be a vertex property and let \(k\in\{1,2\}\).
The following  are equivalent for all $\ell\geq 0$:
    \begin{enumerate}
        \item there exists an \(\AFML[k]\) formula $\varphi$ of modal depth at most $\ell$ such that 
        for all pointed graphs  $(G,v)$: $G,v\models \varphi$ if and only if $(G,v) \in P$.
        \item Spoiler has a winning strategy in $\EF{}{\ell}{\AFML[k]}{G_1,v_1,G_2,v_2}$ for all
        pointed graphs $(G_1,v_1), (G_2,v_2)$ with
        $(G_1,v_1)\in P$ and $(G_2,v_2)\notin P$.
    \end{enumerate}
  \end{theorem}
We start with proving a basic connection between AFML logic and AFML
games. Note that the subsequent theorem implies Theorem~\ref{lemma: ef-afml-equivalence} because
up to equivalence, there are only finitely many \ac{afml} formulas of any fixed modal depth $\ell$.

\begin{restatable}{theorem}{lemGMLgames}
\label{lem:GMLgames}
  Let $k\in \{1,2\}$.
    The following  are equivalent for all possibly infinite pointed graphs $(G_1,v_1)$ and $(G_2,v_2)$
    and all $\ell \geq 0$:
    \begin{enumerate}
        \item there is an \(\AFML[k]\) formula $\varphi$ of modal depth at most $\ell$ such that $G_1,v_1\models \varphi$ and $G_2,v_2\not\models \varphi$,
        \item Spoiler has a winning strategy in $\EF{}{\ell}{\AFML[k]}{G_1,v_1,G_2,v_2}$. 
    \end{enumerate}
\end{restatable}
\noindent
\begin{proof}\
We start with the proof for $\AFML[1]$.

    \medskip
    ``$1 \Rightarrow 2$''.  The proof is by induction
    on $\varphi$.
    For the induction start, there are three cases. First let $\varphi = P$ or $\varphi = \neg P$ and assume that ${G_1,v_1\models \varphi}$ and ${G_2,v_2\not\models \varphi}$.
    Thus, $\pi_1(v_1) \neq \pi_2(v_2)$ and  Spoiler wins $\EF{}{0}{\AFML[1]}{G_1,v_1,G_2,v_2}$.
    The third case is $\varphi = \Box\bot$. Assume that ${G_1,v_1\models\varphi}$ and ${G_2,v_2\not\models \varphi}$. Then $v_1$ has no successors and by the definition of $\AFML[1]$-games Spoiler is allowed to choose a $u_2\in \mathcal{N}(v_2)$, which exists because $v_2$ does not satisfy $\varphi$.
    Duplicator has no response in $\mathcal{N}(v_1) = \emptyset$, thus $S$ has a winning strategy for $\EF{}{1}{\AFML[1]}{G_1,v_1,G_2,v_2}$---note that the modal depth of $\Box\bot$ is $1$.

    For the induction step, assume that the statement holds for $\psi_1$ and $\psi_2$. 
    We have to show that the statement also holds for $\psi_1\land \psi_2$, $\psi_1\lor\psi_2$ and $\Diamond\psi_1$.
    Let $\varphi = \psi_1\land\psi_2$ and assume that ${G_1,v_1\models\varphi}$ and ${G_2,v_2\not\models \varphi}$.
    Then ${G_2,v_2\not\models \psi_k}$ for some $k \in \{1,2\}$ and by induction hypothesis, Spoiler has a winning strategy for $\EF{}{\ell_k}{\AFML[1]}{G_1,v_1,G_2,v_2}$ where $\ell_k$ is the modal depth of $\psi_k$.
    Since the modal depth $\ell$ of $\varphi$ satisfies $\ell\geq \ell_k$, $S$ also has a winning strategy for $\EF{}{\ell}{\AFML[1]}{G_1,v_1,G_2,v_2}$.
    The case $\varphi = \psi_1\lor \psi_2$ is analogous.

    Now let $\varphi = \Diamond\psi$ with $\psi$ of modal depth $\ell$ and assume that  ${G_1,v_1\models \varphi}$ and ${G_2,v_2\not\models \varphi}$. Then there exists a $u_1\in\mathcal{N}(v_1)$ such that $G_1,u_1\models \psi$, and for all $u_2\in\mathcal{N}(v_2)$ we have ${G_2,u_2\not\models\psi}$.
    By induction hypothesis, Spoiler has a winning strategy for $\EF{}{\ell}{\AFML[1]}{G_1,u_1,G_2,u_2}$ for all $u_2\in\mathcal{N}(v_2)$.
    We can extend this to  a winning strategy for $\EF{}{\ell+1}{\AFML[1]}{G_1,v_1,G_2,v_2}$, by letting $S$ choose $u_1$ in the first round.

\medskip
``$2 \Rightarrow 1$''. The proof is  by induction on $\ell$.
    For the induction start,
    assume that 
     Spoiler wins $\EF{}{0}{\AFML[1]}{G_1,v_1,G_2,v_2}$. Then the vertex labels of $v_1$ and $v_2$ differ. We thus
    find an $\AFML[1]$ formula $\varphi$ of
    the form $P$ or $\neg P$
    such that 
     $G_1,v_1\models \varphi
     $ and $G_2,v_2\not\models\varphi$. 

    For the induction step, let the statement hold for ${\ell-1 \geq 0}$ and assume that Spoiler wins $\EF{}{\ell}{\AFML[1]}{G_1,v_1,G_2,v_2}$.
  There are two ways in which this may happen.
    In case $S$ wins because they choose  $u_2\in\mathcal{N}(v_2)$ in $G_2$ and $D$ cannot respond with a $u_1\in\mathcal{N}(v_1)$, we have ${G_1,v_1\models \Box\bot}$ while ${G_2,v_2\not\models \Box\bot}$.
    Therefore, $\Box\bot$ is an $\AFML[1]$ formula that distinguishes $v_1$ and $v_2$. It has modal depth $1 \leq \ell$.


    The second case is that   $S$ chooses $u_1 \in \Nmc(v_1)$ in $G_1$ and 
    for every response $u_2\in\mathcal{N}(v_2)$ that $D$ may choose, $S$ has a winning strategy
    for the game  $\EF{}{\ell-1}{\AFML[1]}{G_1,u_1,G_2,u_2}$.
    By induction hypothesis, for each $u_2\in\mathcal{N}(v_2)$ there exists a formula $\psi_{u_2}$ of modal depth at most $\ell-1$ such that ${G_1,u_1\models \psi_{u_2}}$ and ${G_2,u_2\not\models\psi_{u_2}}$.
    But then $\varphi = \Diamond(\bigwedge_{u_2\in\mathcal{N}(v_2)}\psi_{u_2})$ is an $\AFML[1]$ formula of modal depth $\ell$ such that   $G_1,v_1\models \varphi
     $ and $G_2,v_2\not\models\varphi$ (we take the empty conjunction to be $\top$). 
    %
%

    \medskip
    For $\AFML[2]$, the result 
    follows from the $\AFML[1]$
    case and the observations that (i)~an EF-game for $\AFML[2]$ is exactly an
    EF-game for $\AFML[1]$ with
    the roles of $(G_1,v_1)$ and
    $(G_2,v_2)$ swapped and (ii)~every
    $\AFML[1]$ formula   is equivalent
    to the complement of an $\AFML[2]$ formula and vice versa. 
\end{proof}

\begin{restatable}{lemma}{lemunifAFMLsubsetML}
\label{lem:unifAFMLsubsetML}
  $\AFML \subsetneq \MoL$.
\end{restatable}
\noindent
\begin{proof}\ 
    We show that $\varphi = \Diamond P\land \Box Q$ is not equivalent to an \ac{afml} formula.
    We use the following three graphs:
    \begin{itemize}
        \item $V^A = \{a,b,c\}$,\\
        $E^A = \{(a,b), (a,c)\}$,\\
        $\pi^A(a) =\emptyset$, $\pi^A(b) = \{P,Q\}$ and $\pi^A(c) = \{Q\}$.
        \item $V^{B_1} = \{a,b,c,d\}$,\\
        $E^{B_1} = \{(a,b), (a,c) , (a,d)\}$,\\
        $\pi^{B_1}(a) = \pi^{B_1}(d)=\emptyset$, $\pi^{B_1}(b) = \{P,Q\}$ and $\pi^{B_1}(c) = \{Q\}$.
        \item $V^{B_2} = \{a,c\}$,\\
        $E^{B_2}= \{(a,c)\}$,\\
        $\pi^{B_2}(a) =\emptyset$  and $\pi^{B_2}(c) = \{Q\}$.
    \end{itemize}

    We have $A,a\models \varphi$, $B_1,a\not\models\varphi$ and  $B_2,a\not\models\varphi$.

    We now show that for each $\ell\in\mathbb{N}^+$ Duplicator wins $\EF{}{\ell}{\AFML[1]}{A,a,B_1,a}$ as well as $\EF{}{\ell}{\AFML[2]}{A,a,B_2,a}$.

    In $\EF{}{\ell}{\AFML[1]}{A,a,B_1,a}$, Spoiler does not win with zero rounds, because both vertices $a$ satisfy $\pi^A(a) = \pi^{B_1}(a) = \emptyset$.
    In the first round $S$ can choose  $b\in V^A$ or $c\in V^A$. Duplicator can answer with $b\in V^{B_1}$ and $c\in V^{B_1}$ respectively. 
    Both $b$ satisfy $\pi(b) = \{P,Q\}$, and both $c$ satisfy $\pi(c) = \{Q\}$, so Spoiler does not win in the first round.
    Additionally, each of these vertices does not have successors, hence $S$ has no vertex they can play in the second round.
    Hence, $D$ will win in the next round.
    This implies, that $D$ also has a winning strategy for all $\ell \geq 2$.

    In $\EF{}{\ell}{\AFML[2]}{A,a,B_2,a}$, Spoiler does not win with zero rounds, because both vertices $a$ satisfy $\pi^A(a) = \pi^{B_2}(a) = \emptyset$.
    In the first round, $S$ has to choose $c\in V^{B_2}$ and Duplicator can respond with $c$ in $V^A$, which has the same labeling.
    Since both vertices $c$ do not have successors, $D$ wins in the second round. This also means that $D$ will win the \ac{ef}-game for all $\ell \geq 2$.
\end{proof}
\subsection{Proof of Theorem~\ref{lem:conmeangnnsubseteqafml}}

\begin{lemma}
\label{lem:meangnnbounds}
    For each \conmeanGNN $\mathcal{G}$ with $L$ layers, there exist $m_-,\ m_+\in\mathbb{R}$ such that for all input graph $G$, vertices $v \in V^G$, and $\ell \in [L]$, the vector $\bar{x}^\ell_{G,v}$ only contains  values from
    the range $[m_-,m_+]$.
\end{lemma}
\noindent
\begin{proof}\ 
We identify sequences $m^0_-,m^1_-,\dots,m^L_-$ and $m^0_+,m^1_+,\dots,m^L_+$ such that for all $\ell \in [L]$, the vector $\bar{x}^\ell_{G,v}$ only contains  values from
    the range $[m^\ell_-,m^\ell_+]$. We may then set $m_-$ to the minimum of all $m^\ell_-$
    and $m_+$ to the maximum of all $m^\ell_+$.
    
    We proceed by induction on $\ell$.
    To start, we may set
     $m^0_- = 0$ and $m^0_+ = 1$ because
    the initial feature vectors only contain the values $0$ and $1$.

    Now let $\ell > 0$ and
       let $\delta^{\ell-1}$ 
       and $\delta^{\ell}$ be the output dimensions of layers $\ell-1$ and $\ell$. 
    Then the feature vectors that get computed by $\MEAN$ in layer~$\ell$ are in $[m_-^{\ell-1}, m_+^{\ell-1}]^{\delta^{\ell-1}}$ and $\COM$ gets as input a vector in $[m_-^{\ell-1},m_+^{\ell-1}]^{2\delta^{\ell-1}}$. 
    By the Heine-Borel theorem, $[m_-^{\ell-1},m_+^{\ell-1}]^{2\delta^{\ell-1}}$ is a compact subset of $\mathbb{R}^{\delta^\ell}$. Since  $\COM$ is continuous and it is well-known that the image of a compact set under a continuous function is compact, $\COM([m_-^{\ell-1},m_+^{\ell-1}]^{2\delta^{\ell-1}})$ is also compact.
    Once more applying Heine-Borel,  
%
  %
   we obtain $m^\ell_-$ and $m^\ell_+$ such that $\COM([m_-^{\ell-1},m_+^{\ell-1}]^{2\delta^{\ell-1}})\subseteq [m^\ell_-,m^\ell_+]^{\delta^{\ell}}$.
\end{proof}
%

\smallskip

\lemmagnninvarianceundersmallchanges*

\noindent
\begin{proof}\ 
The proof is  by induction on $\ell$.
Let 
$$
\mathcal{G}=(L,\{\AGG^{\ell}\}_{\ell \in [L]},\{\COM^{\ell}\}_{\ell \in [L]},\CLS)
$$
be a \conmeanGNN with $L$
layers, let $\varepsilon >0$, $n \geq 1$, and $\ell \in [L]$.

In the induction start, where $\ell=0$,
we can
    choose  \mbox{$c=1$} no matter what 
    $\varepsilon$ is because each vertex
    $(v,i)$ in a graph $H$ has the same vertex labels in $H$ and any $n$-extension $H'$ of $H$, and thus $\bar{x}_{H,(v,i)}^{0} = \bar{x}_{H',(v,i)}^0.$

    For the induction step, where $\ell>0$, let
    $m_-,m_+$ be the values from 
    Lemma~\ref{lem:meangnnbounds}. 
    Let $\gamma$ be the output dimension of layer $\ell-1$ which is the input dimension of layer $\ell$.
        By the Heine-Cantor theorem and since $[m_-, m_+]^{2\gamma}\subseteq \mathbb{R}^{2\gamma}$ is compact and  $\COM^{\ell}$ is continuous, the restriction of  $\COM^{\ell}$ to domain $[m_-,m_+]^{2\gamma}$ is uniformly continuous; note that for convenience here we view $\COM^{\ell}$ as a function with a single input vector of dimension $2\gamma$ rather than with two input vectors of dimension $\gamma$.
    There is thus a value 
     $\delta$ such that for all $\bar{x},\bar{y}\in [m_-,m_+]^{2\gamma}$ with $||\bar{x}-\bar{y}||_\infty< \delta$ we have $||\COM^{\ell}(\bar{x})-\COM^{\ell}(\bar{y})||_\infty <\varepsilon$. 
   Let $c^{\ell-1}$ be the constant obtained in the induction hypothesis for $\varepsilon=\delta/5$ and the same value of~$n$. 
   
    Let $m = \max(|m_-|, |m_+|)$ 
    and choose $c^*$ such that $$\frac{mn}{c^*}\leq \frac{\delta}{5} \text{\quad and\quad } c^*\geq n.$$ 
    Set $c = \max(c^{\ell-1}, c^*)$. 


    Now consider any $c' \geq c$, graph $H=c' \cdot G$,
    and $n$-extension $H'$ of $H$.
    Let $(v,i)$ be a vertex in $H$. Then
    $$\bar{x}_{H',(v,i)}^{\ell} = \COM^\ell(\bar{x}_{H',(v,i)}^{\ell-1}, \bar z_{H',(v,i)})$$
    where $\bar z_{H',(v,i)}$ is the result of  mean aggregation,
    and likewise for $\bar{x}_{H,(v,i)}^{\ell}$.      

    Since we are working with the maximum metric and by choice of
    $\delta$, to prove that 
    $$||\bar{x}_{H',(v,i)}^{\ell}-\bar{x}_{H,(v,i)}^{\ell}||_\infty<\varepsilon,$$
    we have to show that $||\bar{x}^{\ell-1}_{H',(v,i)}-\bar{x}^{\ell-1}_{H,(v,i)}||_\infty < \delta$ and 
    $||\bar{z}_{H',(v,i)}-\bar{z}_{H,(v,i)}||_\infty<\delta$.

    The first inequality holds by induction hypothesis since $c'$ is larger or equals to $c^{\ell-1}$. 
    %
   
    We now show the second inequality.
    First assume that $(v,i)$ has at least one successor in $H$. Because $H$ is a $c'$ scaled graph, $(v,i)$ has at least $c'$ and thus at least $c$ successors.
    Let $N=|\mathcal{N}(H, (v,i))|\geq c$. Moreover,
    \begin{align*}
        \bar{z}_{H',(v,i)} &=\MEAN(\{\!\{\bar{x}_{H',(u,j)}^{\ell-1}\mid (u,j)\in\mathcal{N}(H',(v,i))\}\!\}) \\
        &= \frac{1}{|\mathcal{N}(H',(v,i))|}\left(\sum_{(u,j)\in\mathcal{N}(H',(v,i))} \bar{x}_{H', (u,j)}^{\ell-1}\right).
    \end{align*}

    Let \begin{align*}
        \Delta_1 &= \frac{1}{|\mathcal{N}(H',(v,i))|} - \frac{1}{|\mathcal{N}(H,(v,i))|}\\&= \frac{1}{|\mathcal{N}(H',(v,i))|}-\frac{1}{N}
    \end{align*}
    and 
    $$\Delta_2 =\sum_{(u,j)\in\mathcal{N}(H',(v,i))}\bar{x}_{H', (u,j)}^{\ell-1}\ -\sum_{(u,j)\in\mathcal{N}(H,(v,i))}\bar{x}_{H, (u,j)}^{\ell-1}.$$
    We thus can write $\bar{z}_{H',(v,i)}$ as 
    $$ \left(\Delta_1 + \frac{1}{N}\right)\left(\Delta_2 + \sum_{(u,j)\in\mathcal{N}(H,(v,i))} \bar{x}_{H, (u,j)}^{\ell-1}\right).$$    
    The difference between $\bar{z}_{H',(v,i)}$ and $\bar{z}_{H,(v,i)}$ is thus
    $$\Delta_1\left(\sum_{(u,j)\in\mathcal{N}(H,(v,i))} \bar{x}_{H, (u,j)}^{\ell-1}\right)+ \frac{\Delta_2}{N} + \Delta_1\Delta_2.$$
    By the triangle inequality follows that ${||\bar{z}_{H',(v,i)}-\bar{z}_{H,(v,i)}||_\infty}$ is bounded by
    $$|\Delta_1|\!\left|\left|\sum_{(u,j)\in\mathcal{N}(H,(v,i))}\!\bar{x}_{H, (u,j)}^{\ell-1}\right|\right|_\infty\!+ \frac{||\Delta_2||_\infty}{N}  + |\Delta_1|\cdot||\Delta_2||_\infty.$$

    We bound each term as follows:
    \begin{itemize}
        \item $|\Delta_1|$ is bounded by $\frac{1}{N}-\frac{1}{N+n}$ since $H'$ is an $n$-extension of $H$. We have $\frac{1}{N}-\frac{1}{N+n} = \frac{n}{N(N+n)}\leq \frac{n}{N^2}.$
        \item $||\Delta_2||_\infty$ is bounded by $nm + N\frac{\delta}{5}$. In $H'$ there can be up to $n$ new neighbors which add a difference of up to $m$ each. And for each of the $N$ successors in $H$, ${||\bar{x}^{\ell-1}_{H',(u,j)} - \bar{x}^{\ell-1}_{H,(u,j)}||_\infty}$ is bounded by $\frac{\delta}{5}$ by choice of $c^{\ell-1}$. 
        \item $\left|\left|\sum_{(u,j)\in\mathcal{N}(H,(v,i))} \bar{x}_{H, (u,j)}^{\ell-1}\right|\right|_\infty$ is bounded by $Nm$ since each entry is a sum that adds $N$ terms which are all bounded by $m$. 
    \end{itemize}
    Thus, $||\bar{z}_{H',(v,i)}-\bar{z}_{H,(v,i)}||_\infty$ can be bounded by $$\frac{Nmn}{N^2}+ \frac{1}{N}(nm+N\frac{\delta}{5}) + \frac{n}{N^2}\cdot(nm+N\frac{\delta}{5}).$$
    It can now be verified easily that this sum is bounded by $\delta$ since $N\geq c^*\geq n$ and $$\frac{mn}{N}\leq \frac{mn}{c^*}\leq \frac{\delta}{5}.$$

    Now assume that $(v,i)$ has no successors in $H$. Then $\mathcal{N}(H',(v,i)) = \emptyset= \mathcal{N}(H, (v,i))$.
    Thus, $$\bar{z}_{H',(v,i)} = \bar{z}_{H,(v,i)}.$$

    Therefore, for all vertices $(v,i)$ in $H$, we have ${||\bar{x}^{\ell-1}_{H',(v,i)}-\bar{x}^{\ell-1}_{H,(v,i)}||_\infty} < \delta$ and $||\bar{z}^{\ell-1}_{H',(v,i)}-\bar{z}^{\ell-1}_{H,(v,i)}||_\infty<\delta$.
    And by choice of $\delta$ we have 
    \begin{align*}
        &||\bar{x}^\ell_{H',(v,i)} - \bar{x}^\ell_{H,(v,i)}||  = \\
        &  ||\COM^\ell(\bar{x}^{\ell-1}_{H',(v,i)}, \bar{z}^{\ell-1}_{H',(v,i)})
        - \COM^\ell(\bar{x}^{\ell-1}_{H,(v,i)}, \bar{z}^{\ell-1}_{H,(v,i)})||_\infty\\
        & < \varepsilon.
    \end{align*}
\end{proof}

At this point, we have everything we need to prove
Theorem~\ref{lem:conmeangnnsubseteqafml}.

\lemconmeangnnsubseteqafml*
\noindent
\begin{proof}\
    Assume to the contrary that there exists a \contresmeanGNN $\mathcal{G}$ with $L$ layers that is equivalent to an MSO formula, but  not  to an \ac{afml} formula. By Corollary \ref{corollary: meangnn-to-ml-translation}, \Gmc is equivalent to an \MoL formula $\varphi$.
    Let $$\CLS(\bar{x}) = \begin{cases}
    1&\text{ if }x_i \sim c_0,\\
    0&\text{ otherwise}
\end{cases}$$ be the classification function of $\mathcal{G}$, where ${\sim} \in \{ {>},{\geq} \}$.

\medskip
First assume that 
$\sim$ is $>$.    Because $\varphi$ is not expressible in \ac{afml}, for each $\ell\in\mathbb{N}$ there exist pointed graphs $(G,v)$ and $(G',v')$ with $G,v\models \varphi$ and $G',v'\not\models\varphi$ such that $D$ has a winning strategy in $\EF{}{\ell}{\AFML[1]}{G,v,G',v'}$.
    We show below how to transform
    $(G,v)$ into a pointed graph $(H,u)$ such that $H,u\models\varphi$ and $D$ has a winning strategy in $\EF{}{\ell}{\MoL}{H,u,G',v'}$.
    Thus, by Theorem~\ref{lemma: ef-gml-equivalence}, $\varphi$ is not expressible in \ac{ml}. A contradiction.
    
    The transformation
    of  $(G,v)$ into $(H,v)$ 
   crucially relies on the observation that, due to the
equivalence of
$\varphi$ to \Gmc and since $\sim$ is $>$, the following holds:
\begin{description}
    
    \item[$(\ast_1)$] for each pointed graph $(G,v)$ accepted by \Gmc, there is an $\varepsilon>0$ such that
     all pointed graphs $(H,u)$ with ${||\bar{x}^L_v-\bar{x}^L_u||_\infty} < \varepsilon$ are also accepted by \Gmc.
    
\end{description} 
In fact, we may simply choose $\varepsilon = (\bar x^L_v)_i - c_0$.

\medskip

Assume that $D$ has a winning strategy in $\EF{}{\ell}{\AFML[1]}{G,v,G',v'}$. Then for all $K \geq \ell$,  $D$ also has a winning strategy in $\EF{}{\ell}{\AFML[1]}{\Unr^K(G,v), v, G',v'}$,
as a consequence of  Theorem~\ref{lemma: ef-afml-equivalence} and since \AFML formulas of modal depth
$\ell$ are invariant under unraveling up to depth $K \geq \ell$.
Since $\Unr^K(G,v)$ is tree-shaped,
the only way for Spoiler to play
a vertex $u=v_1\cdots v_n$ is to
play exactly the vertices $v_1,v_1v_2,\dots,u$ on the unique
path from the root $v=v_1$ to $u$.
Once this has happened, $S$ can 
never play $u$ again. As a consequence,
the winning strategy of $D$ in $\EF{}{\ell}{\AFML[1]}{\Unr^K(G,v), v, G',v'}$
may be viewed as a function $\ws:V^{\Unr^K(G,v)}\to V^{G'}$  such that if $S$ plays vertex $u$ in
$\Unr^K(G,v)$, then $D$ 
answers with $\ws(u)$. We also 
set $\ws(v)=v'$.

Let $m=|V^{G'}|$ and $K = \max(\ell, L)$.
By Lemma \ref{lemma: gnn-invariance-under-small-changes}, there exists a $c$ such that in each $m$-extension $X$ of ${G''=c\cdot \Unr^K(G,v)}$,
we have $||\bar{x}_{G'',v}^L - \bar{x}_{X,v}^L||_\infty<\varepsilon$
with $\varepsilon$ the constant from $(*_1)$.


The pointed graph $(H,u)$ is defined
as follows:
%
%
\begin{enumerate}

    \item start with $G''=c\cdot \Unr^K(G,v)$;

    \item take the disjoint union with all $\Unr^K(G',v')$,  $v'\in V^{G'}$;

    \item for each vertex $(u,i) \in V^{G''}$ that has at least one successor, let $\mathcal{N}(G',\ws(u)) =\{u_1',\ldots, u_m'\}$.
    Add to $(u,i)$ the fresh successors $u_1',\ldots,u_m'$ (which are all roots of unravelings added in Step~2);

    \item $u=(v,1)$.
\end{enumerate}
{\bf Claim~1.}
$H,u \models \varphi$.
\\[1mm]
Since $G,v \models \varphi$, $(G,v)$ is 
accepted by \Gmc. 
By Lemmas~\ref{lemma: gnn-invariance-under-unraveling} and~\ref{lemma: gnn-invariance-under-scaling}, $\bar{x}_{G,v}^L =\bar{x}_{G'', v}^L$.
By
choice of
$c$ and since $H$ is an $m$-extension of $G''$,
we have $||\bar{x}_{G,v}^L - \bar{x}_{H,u}^L||_\infty<\varepsilon$.
Consequently, it follows from
$(*_1)$ that $(H,u)$ is also accepted by \Gmc and therefore
$H,u \models \varphi$, as desired.
\\[3mm]
{\bf Claim~2.} $D$ has a winning strategy in $\EF{}{\ell}{\MoL}{H,u,G',v'}$.
\\[2mm]
To prove Claim~2, we show that for each $k\leq \ell$, the following holds:
    \begin{itemize}
        \item[(i)] $D$ has a winning strategy in $\EF{}{K-k}{\MoL}{H,v_1'\cdots v_k',G',v_k'}$, for each $v_1'\cdots v_k'\in V^{\Unr^K(G',v_1')}$.
        
        \item[(ii)] If $D$ has a winning strategy in $$\EF{}{k}{\AFML[1]}{\Unr^K(G,v_1), v_1\cdots v_n, G',\ws(v_1\cdots v_n)},$$ then for all $i$, $D$ has a winning strategy in $$\EF{}{k}{\MoL}{H, (v_1\cdots v_n, i), G', \ws(v_1\cdots v_n)}.$$
        
    \end{itemize}
    Claim~2 follows from Point~(ii) since  $D$ has a winning strategy in  $\EF{}{\ell}{\AFML[1]}{\Unr^K(G,v),v,G',v'}$.

Point~(i) is almost immediate.
By Theorem~\ref{lemma: ef-gml-equivalence} and since modal
formulas of depth $K-k$ are invariant under unraveling up to
depth $K$, 
$D$ has a winning strategy in $\EF{}{K-k}{\MoL}{\Unr^K(G',v_1'), v_1'\cdots v_k',G',v_k'}$. Point~(i) follows since in $H$ there is no edge that leaves the disconnected component $\Unr^K(G',v_1')$.

It thus remains to prove Point~(ii),
which we do  by induction on $k$. 
    For $k=0$ notice that 
    $v_1\cdots v_n$ and $\ws(v_1\cdots v_n)$ have the same vertex labels because $D$ wins the $\AFML[1]$-game. By definition of $H$, $(v_1\cdots v_n, i)$ also has the same labels and thus $D$ wins the 0-round $\MoL$-game.

    For the induction step, let (ii) hold for $k-1\geq 0$ and assume that $D$ wins $$\EF{}{k}{\AFML[1]}{\Unr^K(G,v_1), v_1\cdots v_n, G',\ws(v_1\cdots v_n)}.$$
    Therefore, $\mathcal{N}(\Unr^K(G,v_1),v_1\cdots v_n) = \emptyset$ if and only if $\mathcal{N}(G', \ws(v_1\cdots v_n)) = \emptyset$, because otherwise Spoiler could win in the first round.

    If $\mathcal{N}(\Unr^K(G,v_1), v_1\cdots v_n) = \emptyset$, then   by definition of~$H$ also $\mathcal{N}(H, (v_1\cdots v_n, i)) = \emptyset$
    Hence, $D$ wins $\EF{}{k}{\MoL}{H, (v_1\cdots v_n, i), G', \ws(v_1\cdots v_n)}$ since $S$ cannot choose a successor.
    Otherwise, Spoiler has three choices in this game:
    \begin{itemize}
        \item $S$ can choose a vertex $(v_1\cdots v_n u, j)\in \Nmc(H,v_1 \cdots v_n)$ such that $v_1\cdots v_n u\in V^{\Unr^K(G,v_1)}$. 
        
        Then $D$ may choose
        $$\ws(v_1\cdots v_nu)\in \Nmc(G',\ws(v_1 \cdots v_n)).$$ But $\ws$ represents a winning strategy for $D$ in  $$\EF{}{k-1}{\AFML[1]}{\Unr^K(G,v_1), v_1\cdots v_nu, G',\ws(v_1\cdots v_nu)}.$$
        By induction hypothesis, $D$ thus has a winning strategy for the remaining game
        $$\EF{}{k-1}{\MoL}{H, (v_1\cdots v_nu, j), G', \ws(v_1\cdots v_nu)}.$$

        \item $S$ can choose a vertex $u'\in\Nmc(H,v_1\cdots v_n)$ such that $u' \in V^{\Unr^K(G', u')}$.
        
        Then by definition of $H$, $u' \in \Nmc(G',\ws(v_1,\cdots v_n))$.  Thus $D$ can choose $u'$.
        By Point~(i), $D$ has a winning strategy in the remaining game $\EF{}{k-1}{\MoL}{H,u',G',u'}$.
     
        \item 
         $S$ can  choose a successor $u'$ of  $\ws(v_1,\ldots,v_n)$ in $G'$.
                
        Since $\ws(v_1\cdots v_n)$ has a successor in $G'$, 
        $v_1\cdots v_n$ also has at least one successor in
        $\Unr^K(G, v_1)$.
        By construction of $H$,
        this implies that $(v_1\cdots v_n,i)$ has $u'$ as a successor in $H$.
        Thus, $D$ can choose $u'$.
        By Point~(i),  $D$ has a winning strategy in $\EF{}{k-1}{\MoL}{H,u',G',u'}$.
    \end{itemize}
    This finishes the case 
    where $\sim$ is $>$.

\medskip

  The case where $\sim$ is $\geq$ is analogous. 
  Because $\varphi$ is not expressible in AFML, for each $\ell \in \mathbb{N}$ there exist pointed graphs $(G,v), (G',v')$ with $G,v\models\varphi, G',v'\not\models \varphi$ and $D$ has a winning strategy for $\EF{}{\ell}{\AFML[2]}{G,v,G',v'}$. It is easy to see
  that any such winning strategy is
  also a winning strategy for $\EF{}{\ell}{\AFML[1]}{G',v',G,v}$. In fact, these games are exactly identical. 
  Moreover, since $\sim$ is $\geq$, the following holds:
  \begin{itemize}

  \item[$(\ast_2)$] for each pointed graph $(G,v)$ rejected by \Gmc, there is an $\varepsilon>0$ such that
     all pointed graphs $(H,u)$ with ${||\bar{x}^L_v-\bar{x}^L_u||_\infty} < \varepsilon$ are also rejected by \Gmc.

  \end{itemize}
We may  use exactly the same arguments as in the case where $\sim$ is $>$, except that the roles of 
$(G,v)$ and $(G',v')$ are swapped.
This includes the exact same construction of $(H,u)$, now from $(G',v')$. Instead of showing that
$H,u \models \varphi$, we then have to show that $H,u \not\models \varphi$. This is exactly analogous since $(*_2)$ is about rejection while $(*_1)$ is about acceptance.
\end{proof}

\subsection{Proof of Theorems~\ref{lemma: afml-to-cmgnn-translation} and~\ref{thm:sfjlherogahfga}}

\lemmaafmltocmgnntranslation*
\noindent
\begin{proof}\ 
Recall that $\AFML=\AFML[1] \cup \AFML[2]$.
    We start with $\AFML[1]$ formulas.
    Let $\varphi$ be an $\AFML[1]$ formula over a finite set
    $\Pi=\{P_1,\dots,P_r\}$ of vertex labels, 
    and let $\varphi_1,\ldots,\varphi_L$ be an enumeration of the subformulas of $\varphi$ such that (i)~$\varphi_i=P_i$ for $1 \leq i \leq r$ and
(ii)~if $\varphi_\ell$ is a subformula of $\varphi_k$,
then $\ell < k$.
We construct a simple \ac{gnn} \Gmc with $2L$ layers, all of input and output dimension $3L+1$ and with all combination functions identical. Each subformula gets computed within two layers. The truth value of the
$i$-th subformula is stored in the $i$-th component of feature vectors. Additionally, we use up to two positions per subformula for bookkeeping purposes, which explains why we need at least $3L$ components in
feature vectors. The additional $3L+1$-st 
component is needed for technical purposes
and will be constant~1.



We define the matrices $A,\ C\in \mathbb{R}^{3L+1\times 3L+1}$ and bias vector $\bar{b}\in \mathbb{R}^{3L+1}$ that define the combination function. All entries that are not mentioned explicitly have value~$0$. Set $b_{3L+1} = 1$ to achieve that
the $3L+1$-st component is constant $1$.\footnote{With the exception of the initial feature vectors, for which this is irrelevant.} Let $k \in [L]$. We make a case 
distinction to set certain entries:
    \begin{enumerate}[label=\textit{Case }\arabic*:, ref=\arabic*, leftmargin=*]

        \item $\varphi_k = P_k$: Set $C_{k,k} = 1$.
        \item $\varphi_k = \lnot P_i$: Set $C_{i,k} = -1$ and  $b_k = 1$.
        \item\label{case: box-bottom} $\varphi_k = \Box\bot$: Set $A_{3n+1,k} = -1$ and $b_k = 1$.
        \item $\varphi_k = \varphi_i\lor\varphi_j$: Set $C_{i,k} = C_{j,k} = 1$.
        \item\label{case: diamond} $\varphi_k = \Diamond\varphi_i$: Set $A_{i,k} = 1$.
        \item\label{case: land} $\varphi_k = \varphi_i\land\varphi_j$: 
        Set 
        $$
          \begin{array}{rcl}
          C_{i,k+n} &=& 1 \\[1mm]
          C_{j, k+n} &=& -1 \\[1mm] C_{i,k+2n} &=& -1 \\[1mm] C_{j,k+2n} &=& 1 \\[1mm]
          C_{i,k} &=& C_{j,k} = 1 \\[1mm]
          C_{k+n,k} &=& C_{k+2n,k} = -1.
          \end{array}
        $$
    \end{enumerate}
    Note that the use of bookkeeping
    values in  components $L+1,\dots,3L$ of feature vectors happens (only) in Case~6.
    We use the classification function
    $$\CLS(\bar{x}) = \begin{cases}
        1 & \text{ if } x_L>0\\
        0 & \text{otherwise}.
    \end{cases}$$
We next show correctness of the translation.
\\[2mm]
{\bf Claim 1.} For all $\varphi_k$, $1 \leq k \leq L$, the following holds:
if $v \in V^G$ and $2k\leq k' \leq L$, then $(\bar{x}_{G,v}^{k'})_{k} \in (0,1]$ if $G,v\models \varphi_k$ and  $(\bar{x}_{G,v}^{k'})_{k}=0$ otherwise.
\\[2mm]
    We prove Claim~1 by induction on $k$.
%
    The correctness for Cases $1$, $2$, and $4$ is straightforward to  verify.

    For Case \ref{case: box-bottom}, where $\varphi_{k} = \Box\bot$, we exploit
    that the $3L+1$-st component of every feature vector is constant $1$. This implies that if a vertex has at least one successor, then 
     $$\MEAN(\{\!\{\bar{x}^{k'-1}_{G,u}\mid u\in\mathcal{N}(v)\}\!\})$$
    returns a vector with $1$ in the $3L+1$-st component and thus the $k$-th component in $\bar{x}_{G,v}^{k'}$ will be set to $0$.
    When a vertex has no successors, the above mean returns a vector with $0$ in the $3L+1$-st component and the $k$-th component in $\bar{x}_{G,v}^{k'}$ will be set to $1$.

    For Case \ref{case: diamond}, where $\varphi_k=\Diamond \varphi_i$, we know from the induction hypothesis that for each $u\in V^G$, the value stored in the $i$-th component
    of feature vector $\bar{x}^{k'-1}_{G,u}$ is from $(0,1]$ if $G,u\models \varphi_i$ and $0$ otherwise.
    Therefore,
          $$\MEAN(\{\!\{\bar{x}^{k'-1}_{G,u}\mid u\in\mathcal{N}(v)\}\!\})$$
    returns a vector that has $0$ in the
    $i$-th component if $v$ has no successors that satisfies $\varphi_i$ and some value from $(0,1]$ otherwise. 
    This value is then stored in the $k$-th entry of $\bar{x}^{k'}_{G,v}$.
    Note that we do not have
    much control of that value except
    that it is  from the stated range. This is the intuitive reason why there is no obvious way to encode the truth of formulas using value~1.
    
    For Case \ref{case: land}, we first observe the following.
    \\[2mm]
    {\bf Claim 2.}
    Let $x,y \in [0,1]$. Then the following are equivalent:
    \begin{enumerate}
        \item $\trRELU(x+y-\trRELU(x-y)-\trRELU(y-x)) > 0$;
        \item $x > 0$ and $y > 0$.
    \end{enumerate}
    %
    %
    \emph{Proof of claim.}
    When $x = 0$ then $\trRELU(x-y) = 0$  and $\trRELU(y-x) = \trRELU(y) = y$, both because $y \in [0,1]$.
    Hence, $\trRELU(x+y-\trRELU(x-y)-\trRELU(y-x))$ becomes $\trRELU(0+y-0-y) = 0$.
    The case $y = 0$ is analogous.

    For the other direction, let $0<x\leq y$.
    Then $\trRELU(y-x) = y-x$ and $\trRELU(x-y) = 0$, both because $x,y \in [0,1]$.
    Thus,  $\trRELU(x+y-\trRELU(x-y)-\trRELU(y-x))$ becomes $\trRELU(x+y-(y-x)) = \trRELU(2x) \in (0,1]$. The case  $0<y\leq x$ is analogous.
This finishes the proof of Claim~2.

    %
    \medskip

    Now let $\varphi_{k} = \varphi_i\land\varphi_j$. Analyzing the
    coefficient set in Case~6, it can be
    verified that  after two layers, $(\bar{x}_v^{k'})_{k}$ contains $\trRELU((\bar{x}_v^{k'-1})_i+(\bar{x}_v^{k'-1})_j-\trRELU((\bar{x}_v^{k'-2})_i-(\bar{x}_v^{k'-2})_j)-\trRELU((\bar{x}_v^{k'-2})_j-(\bar{x}_v^{k'-2})_i))$.
    By Claim~2, this is the desired value.

\medskip

    This finishes the proof for $\AFML[1]$ formulas.
    We now show how to realize $\AFML[2]$ formulas.
    Let $\varphi\in\AFML[2]$. Then there exists a $\psi\in\AFML[1]$ such that $\varphi\equiv \neg \psi$. We have just seen that there if a GNN \Gmc that is
    equivalent to $\psi$.
   We add a layer to $\mathcal{G}$ that computes the function $f(\bar{x}) = 1-x_L$, where $L$ is the index used by the classification function in $\mathcal{G}$.
   This is easily realizable by a simple \conmeanGNN.
    We also replace the classification function with $$\CLS(x) = \begin{cases}
    1&\text{ if } x\geq 1\\
     0&\text{ otherwise.}
\end{cases}$$
Now  $\mathcal{G}$ accepts a pointed graph $(G,v)$ if and only if the constructed \ac{gnn} rejects $(G,v)$.
Thus, the latter is equivalent to $\neg\psi \equiv\varphi$.%
\end{proof}

\lemAFMLinsimplemeanGNNwithsigmoid*

\noindent
\begin{proof}\ 
 %
As in Lemma~\ref{lemma: afml-to-cmgnn-translation}, we first concentrate on $\AFML[1]$ formulas.
      Let $\varphi$ be such a formula  and  let $\varphi_1,\ldots,\varphi_L$ be an enumeration of the subformulas of $\varphi$ such that (i)~$\varphi_i=P_i$ for $1 \leq i \leq r$ and
(ii)~if $\varphi_\ell$ is a subformula of $\varphi_k$,
then $\ell < k$.
We construct a simple \ac{gnn} \Gmc with $5L+1$ layers, all of input and output dimension $13L+1$. In contrast to previous reductions, the combination function of the first layer is different to that of the other layers, which share the same combination function. This helps us to deal with formulas of the form $P_k$ and $\neg P_k$ in a technically simple way and  could be avoided at the expense of a slightly more technical translation. Each subformula gets computed within five layers. As before, the truth value of the
$i$-th subformula is stored in the $i$-th component of feature vectors. The additional positions are used for bookkeeping purposes, with the  $13L+1$-st 
component being constant~$\frac{1}{2}$.

We first introduce a suitable helper function that will assist us to deal with conjunction.
\\[2mm]
{\bf Claim 1.}
    Define 
    $$\tau(x) = -\sigma(\sigma(x))-\sigma(\sigma(-x)).$$ Then the following properties hold:
    \begin{enumerate}
        \item $\tau$ has a strict global minimum at $x=0$, with $e_{\min}=\tau(0) = -\frac{2}{1+\frac{1}{\sqrt{e}}}$,
        \item $\tau$ is symmetric in the sense that $\tau(x) = \tau(-x)$ for all $x \in \mathbb{R}$, and
        \item $\tau$ is injective on the interval $[0,\infty)$.
    \end{enumerate}
    The function
    $$\alpha(x,y) = \tau(\tau(x-y)-\tau(x+y-1)) - e_{\min}$$
    is $0$ if $x = \frac{1}{2}$ or $y = \frac{1}{2}$ and is greater than $0$ otherwise.
\\[2mm]
    Properties~1-3 of $\tau$ are straightforward to verify.
By definition and Property~2, the function $\alpha$ has a strict global minimum if $\tau(x-y) = \tau(x+y-1)$ which by Properties~2 and~3 is the case if and only if $x-y = x+y-1$ or $x-y = -x-y+1$.
The first case holds if $y = \frac{1}{2}$ and the second case holds if $x = \frac{1}{2}$. This finishes the proof of Claim~1. 

\medskip

We define the matrices $A,\ C\in \mathbb{R}^{13L+1\times 13L+1}$ and bias vector $\bar{b}\in \mathbb{R}^{13L+1}$ that define the combination function for the first layer. All entries that are not mentioned explicitly have value~$0$. Set $b_{13L+1} = 0$ to achieve that
the $13L+1$-st component is  $\sigma(0) = \frac{1}{2}$. Let $k \in [L]$. We make a case 
distinction as follows:
    \begin{enumerate}[label=\textit{Case }\arabic*:, ref=\arabic*, leftmargin=*]
        \item  \label{case: sigmoid-variable-first-layer}$\varphi_k = P_k$. Set $C_{k,k} = 1$ and $b_k = 0$.        \item  \label{case: sigmoid-negvariable-first-layer} $\varphi_k = \lnot P_i$. Set $C_{i,k} = -1$ and $b_k = 1$.
\end{enumerate}
What this achieves is that, for formulas $\varphi_k$ of the form 
$P_k$ or $\neg P_k$, $(\bar{x}_v^1)_k \in (\frac{1}{2},1)$
if  $G \models \varphi_k$
and $(\bar{x}_v^1)_k =\frac{1}{2}$ if $G \not\models \varphi_k$. The first layer
thus treats all these formulas
simultaneously.\footnote{Which actually means that $5L-5m$ of the remaining layers suffice, where $m$ is the number of subformulas of $\varphi$ that are of the form $P_k$ or $\neg P_k$; however, additional layers are not harmful.}

The remaining $5L$ layers all use the same combination function. 
We next give the defining matrices $A,\ C\in \mathbb{R}^{13L+1\times 13L+1}$ and bias vector $\bar{b}\in \mathbb{R}^{13L+1}$, again only
explicitly giving entries that are not 0. We set $b_{13L+1} = 0$ to achieve that
the $13L+1$-st component is  $\sigma(0) = \frac{1}{2}$. Let $k \in [L]$. We make a case 
distinction as follows:
    \begin{enumerate}[label=\textit{Case }\arabic*:, ref=\arabic*, leftmargin=*]
        \item  \label{case: sigmoid-variable}$\varphi_k = P_k$. Set $C_{k,k} = 1$ and $b_k = -\frac{1}{2}$.        \item  \label{case: sigmoid-negvariable} $\varphi_k = \lnot P_i$. Set $C_{k,k} = 1$ and $b_k = -\frac{1}{2}$.
        \item $\varphi_k = \Box\bot$. Set $A_{13L+1,k} = -1$ and $b_k = \frac{1}{2}$.
        \item $\varphi_k = \varphi_i\lor\varphi_j$. Set $C_{i,k} = C_{j,k} = 1$ and $b_k = -1$.
        \item \label{case: sigmoid-diamond}$\varphi_k = \Diamond\varphi_i$. Set $A_{i,k} = 1$ and $A_{13L+1,k} = -1$. 
        \item $\varphi_k = \varphi_i\land \varphi_j$. Using Claim~1, it is easy to see that $$\sigma(\alpha((\bar{x}_v^{k'-1})_i,(\bar{x}_v^{k'-1})_j))$$ gives the desired result.
        Computing $\alpha$ requires
        three applications of $\tau$, each in turn requiring four temporary values, namely for $\sigma(x), \sigma(\sigma(x)), \sigma(-x)$ and $\sigma(\sigma(-x))$, and $2$ layers. In $\alpha$, the two inner computations of $\tau$ can be computed in parallel, thus $\alpha$ needs $12$ temporary values and $4$ layers. It is straightforward to work out the corresponding entries in $C$ and $\bar b$, we omit details.
    \end{enumerate}
    Note that the use of bookkeeping
    values in  components $L+1,\dots,13L$ of feature vectors happens (only) in Case~6.
     We use the classification function
    $$\CLS(\bar{x}) = \begin{cases}
        1&\text{ if } x_L > \frac{1}{2}\\
        0&\text{ otherwise.}
    \end{cases}$$
We next show correctness of the translation.
\\[2mm]
{\bf Claim 2.} For all $\varphi_k$, $1 \leq k \leq L$, the following holds:
if $v \in V^G$ and $5k+1\leq k' \leq L$, then $(\bar{x}_v^{k'})_k \in (\frac{1}{2},1)$
if  $G \models \varphi_k$
and $(\bar{x}_v^{k'})_k = \frac{1}{2}$ if $G \not\models \varphi_k$. 
\\[2mm]
    We prove Claim~2 by induction on $k$.
%

    Cases~1 and~2 rely on the fact that formulas of the form $P_k$ and $\neg P_k$ have already been treated in the first layer. They simply implement the `identity function' in the sense that 
    $(\bar{x}_v^{k'})_k \in (\frac{1}{2},1)$ if  
     $(\bar{x}_v^{k'-1})_k \in (\frac{1}{2},1)$ and
     $(\bar{x}_v^{k'})_k = \frac{1}{2}$ if  
     $(\bar{x}_v^{k'-1})_k = \frac{1}{2}$
     for all $k' \geq 1$.

  For Case \ref{case: box-bottom}, where $\varphi_{k} = \Box\bot$, we exploit
    that the $13L+1$-st component of every feature vector is constant $\frac{1}{2}$. This implies that if a vertex has at least one successor, then 
     $$\MEAN(\{\!\{\bar{x}^{k'-1}_{G,u}\mid u\in\mathcal{N}(v)\}\!\})$$
    returns a vector with $\frac{1}{2}$ in the $13L+1$-st component and thus the $k$-th component in $\bar{x}_{G,v}^{k'}$ will be set to $\sigma(0)=\frac{1}{2}$.
    When a vertex has no successors, the above mean returns a vector with $0$ in the $13L+1$-st component and the $k$-th component in $\bar{x}_{G,v}^{k'}$ will be set to $\sigma(\frac{1}{2}) \in (\frac{1}{2},1)$. 

    Case~4  is straightforward to  verify.

    For Case~5, first assume that a vertex $v$ has a successor $u$ that satisfies $\varphi_i$. Then $(\bar{x}_u^{k'-1})_i > \frac{1}{2}$  and therefore $\MEAN(\{\!\{ 
    (\bar{x}_u^{k'-1})_i \mid 
    u\in\mathcal{N}(v) \}\!\})  > \frac{1}{2}.$
    Additionally, $\MEAN(\{\!\{(\bar{x}_u^{k'-1})_{13L+1}
    \mid u\in\mathcal{N}(v) \}\!\}) = \frac{1}{2}$, thus $(\bar{x}_v^{k'})_k > \frac{1}{2}$.
Now assume that $v$ has at least one successor, but none of its successors satisfies $\varphi_i$. Then $\MEAN(\{\!\{(\bar{x}^{k'-1}_u)_i
\mid u\in\mathcal{N}(v) \}\!\}) =\frac{1}{2}$,
implying $(\bar{x}_v^{k'})_k = \frac{1}{2}$.
    Finally assume that $v$ has no successors. Then $\MEAN(\{\!\{ (\bar{x}^{k'-1}_u)_{i} \mid {u\in\mathcal{N}(v)} \}\!\}) =\MEAN(\{\!\{ (\bar{x}^{k'-1}_u)_{13L+1} \mid
    {u\in\mathcal{N}(v)} \}\!\}) = 0$ and thus $(\bar{x}_v^{k'})_k = \sigma(0) = \frac{1}{2}$.

    For Case~6, it suffices to invoke Claim~1 and observe that
    $\sigma(0)=\frac{1}{2}$ and
    $\sigma(x)\in(\frac{1}{2},1)$
    for all $x >0$.
    
    \medskip

    If $\varphi$ is a formula in $\AFML[2]$ we again use that there exists a formula $\psi\in\AFML[1]$ such that $\varphi\equiv \neg \psi$.
    Let $\mathcal{G}$ be the \ac{gnn} that is equivalent to $\psi$.
    We add a layer to $\mathcal{G}$ that computes $\sigma(\frac{1}{2}-x_L)$.
    The classification function is replaced with 
    $$
    \CLS(x)=
    \begin{cases}
    1&\text{ if } x\geq \frac{1}{2}\\
    0&\text{ otherwise.}
\end{cases}$$
Now  $\mathcal{G}$ accepts a vertex $v$ if and only if the extended \ac{gnn} rejects $v$.
Thus, the latter is equivalent to $\neg\psi \equiv \varphi$.
\end{proof}

\section{Proof of Theorem~\ref{thm:summarymeanabsoluteuniform}}

To prove Theorem~\ref{thm:summarymeanabsoluteuniform}, we state as
separate theorem the items stated there. These are
Theorems~\ref{thm:pointone}
and~\ref{theorem: rml_not_expr_by_meanc}.


\begin{theorem}
  \label{thm:pointone}
  The property $\mathcal{P}$ `there exist more successors that satisfy~$P_1$ than successors that satisfy $P_2$' is not expressible in \ac{rml}, but  by a simple \contresmeanGNN.
\end{theorem}

\noindent
\begin{proof}\ 
    Assume that $\varphi$ is a formula in \ac{rml} that expresses this property.
    Let $R$ be the set of all rational numbers $r$ such that $\markDiamond^{>r}$ or $\markDiamond^{\geq r}$ occurs in $\varphi$.
    Because $R$ is finite, there exists $x = \min (R\cup
    (1-R))\setminus\{0\}$. 
    Clearly, $(0,x)\cap R = (1-x, 1)\cap R = \emptyset$.
    It is easy to see that $\varphi$ cannot distinguish between different fractions in $(0,x)$.

    Let $y\in \mathbb{Q}$ such that $0 < 3y < x$.
    We now consider a graph $G_1$ where a vertex $v_1$ has $y$
    successors that satisfy $P_1$, $2y$ successors that satisfy $P_2$,
    and $1-3y$ successors that satisfy neither $P_1$ nor $P_2$.
    Graph $G_2$ with vertex $v_2$ is defined analogously, but with
    the roles of $P_1$ and $P_2$ swapped.
    
    In both graphs all fractions are in the intervals $(0,3y]\subseteq (0,x)$ and $[(1-3y), 1)\subseteq (1-x,1)$, thus $\varphi$ will classify $v_1$ and $v_2$ identically.
    But $v_2$ satisfies property $\mathcal{P}$ while $v_1$ does not.
    Thus, there exists no \ac{rml} formula that expresses $\mathcal{P}$.

    The one-layer, simple \ac{gnn} with $\trRELU$ activation, $C$ the all zero matrix and $$A = \begin{pmatrix}
        1\\-1
    \end{pmatrix}\text{ and } \CLS(x) = \begin{cases}
        1&\text{ if } x>0\\
        0&\text{ otherwise}
    \end{cases}$$
    computes the fraction of successors satisfying $P_1$ and subtracts the fraction of successors satisfying $P_2$.
    This difference is greater than $0$ if and only if there are more successors satisfying $P_1$ than $P_2$.%
\end{proof}

We recall the notions of convex sets and affine spaces, as well as 
some lemmas regarding the interior and closure of convex sets.
See \cite{Rockafellar1970} for more in-depth information.

\begin{definition}
    The \emph{affine hull} $\aff(S)$ of a set $S\subseteq
    \mathbb{R}^n$ is defined
    as $$\{\lambda_1x_1+\cdots+\lambda_mx_m\mid m\in\mathbb{N},
    x_i\in S, \lambda_i \in \mathbb{R}, \sum_{i=1}^m\lambda_i = 1\}.$$
    It is the smallest superset of $S$ that contains all lines through two of its points.
    That is, for $x,y\in\aff(S)$, $$\{(1-\lambda)x+\lambda y\mid\lambda\in\mathbb{R}\}\subseteq \aff(S).$$
\end{definition}
%
%
\begin{definition}
    The \emph{convex hull} of a set $S\subseteq \mathbb{R}^n$ is defined as \begin{align*}
        \{&\lambda_1x_1+\cdots+\lambda_mx_m\mid\\&\qquad m\in\mathbb{N},\ x_i\in S,\ 0\leq \lambda_i\leq 1,\ \sum_{i=1}^m\lambda_i = 1\}.
    \end{align*}
    We denote the convex hull by $\cvh(S)$.
    It is the smallest superset of $S$ that contains all line segments between two of its points.
    That is, for $x,y$ in the convex hull of $S$, $$\{\lambda x+(1-\lambda) y\mid 0\leq\lambda\leq 1\}$$ is a subset of the convex hull.
    A set $S$ is called convex if it is identical to its convex hull.
\end{definition}

The convex hull of $S$ and the means of multisets over $S$ are tightly connected.
\begin{lemma}
    Let $S\subseteq \mathbb{R}^n$ be a set of vectors and let $\mathcal{M}(S)$ be the set of all finite multisets over $S$.
    Then $$\MEAN(\mathcal{M}(S)) = \{\MEAN(M) \mid M\in \mathcal{M}(S)\}\subseteq \cvh(S).$$
    Moreover, the means of elements in $\mathcal{M}(S)$ define exactly the elements in the convex hull where the parameters $\lambda$ are rational numbers.
    That is, \begin{align*}
        \{\MEAN(M)\mid M\in\mathcal{M}(S)\} =
         \{\lambda_1x_1+\cdots +\lambda_mx_m&\\\mid m\in\mathbb{N}, x_i\in S, \lambda_i\in [0,1]\cap\mathbb{Q}, \sum_{i=1}^m\lambda_i = 1\}&.
    \end{align*}
\end{lemma}

\begin{lemma}\label{lemma:mean-cvh-dense-preserving}
    Let $T\subseteq S\subseteq \mathbb{R}^n$ and let $T$ be dense in $S$. Then $\{\MEAN(M)\mid M\in\mathcal{M}(T)\}$ is dense in $\cvh(S)$.
\end{lemma}
\noindent\begin{proof}\ 
    We can rewrite $\MEAN(\mathcal{M}(T))$ as
    \begin{align*}
        \MEAN(&\mathcal{M}(T)) = \{\lambda_1x_1+\cdots+\lambda_mx_m\\
        &\mid m\in\mathbb{N}, x_i\in T, \lambda_i\in[0,1]\cap \mathbb{Q}, \sum_{i=1}^m\lambda_i = 1\}.
    \end{align*}
    Since $\mathbb{Q}$ is dense in $\mathbb{R}$ and $T$ is dense in $S$, it is easy to verify that each vector in $\cvh(S)$ can be approximated by vectors in $\MEAN(\mathcal{M}(T))$.
    Thus, $\MEAN(\mathcal{M}(T))$ is dense in $\cvh(S)$.
\end{proof}

\begin{definition}
    The \emph{relative interior} of a convex set $S\subseteq\mathbb{R}^n$ is the (topological) interior of $S$ relative to its affine hull.
    That is \begin{align*}
    \ri(S) = \{x\in S\mid\ &\exists \varepsilon>0\ \forall y\in\aff(S)\\&||x-y||_\infty < \varepsilon \implies y\in S\}.
    \end{align*}
\end{definition}

\begin{lemma}
For all convex sets $S$,
   \begin{itemize}
    \item the relative interior of $S$ is empty if and only if $S$ is the empty set, and
    \item the closure of $\ri(S)$ is the same as the closure of $S$. Furthermore, the closure is contained in the affine hull of~$S$:
    $$\clo(\ri(S)) = \clo(S) \subseteq \aff(S).$$
   \end{itemize} 
\end{lemma}

\begin{lemma}[Accessibility Lemma]\label{lemma: accessibility lemma}
   Let $S\subseteq \mathbb{R}^n$ be a convex set.
   Let $x\in\ri(S)$ and $y\in\clo(S)$. Then $$\{\lambda x+(1-\lambda) y\mid 0\leq \lambda<1\}\subseteq \ri(S).$$ 
\end{lemma}
%

We also introduce some lemmas regarding dense sets in $\mathbb{R}^n$ with respect to the maximum metric $||\cdot||_\infty$.

 \begin{definition}
    Let $X\subseteq Y\subseteq \mathbb{R}^{\delta}$. Then $X$ is dense in $Y$ if for all $y\in Y$ and $\varepsilon > 0$ we have
     $$B_\varepsilon(y) = \{y'\in Y\mid ||y'-y||_\infty < \varepsilon\}\cap X \neq \emptyset.$$
\end{definition}

\begin{lemma}\label{lemma: continuous-preserve-dense-sets}
    Let $S\subseteq X$ be dense in $X$ and let $f$ be a continuous function. Then $f(S)$ is dense in $f(X)$.
\end{lemma}

\begin{lemma}\label{lemma: dense-sets-open-set-intersection}
   Let $X,Y\subseteq Z\subseteq \mathbb{R}^n$ such that $X$ is dense in $Z$ and $Y$ is open in $Z$.
   Then $X\cap Y$ is dense in $Y$.  
 \end{lemma}
 
\noindent
\begin{proof}\ 
  Let $y\in Y$. Further let  $\varepsilon>0$. Define the open
  $\varepsilon$-ball around $y$ to be
  $$
  B_\varepsilon(y) = \{z\in Y\mid ||z-y||_\infty < \varepsilon\}.
  $$
  We have to show that $B_\varepsilon(y)\cap X\cap Y\neq \emptyset$.

    As $Y$ is open in $Z$, there is a $\delta$
    with  $0 < \delta<\varepsilon$ such that the open
    $\delta$-ball around $y$ is a subset of $Y$, that is
    $$B_\delta(y) = \{z\in Z\mid ||z-y||_\infty < \delta\}\subseteq Y.
    $$
    Therefore, $B_\delta(y)\subseteq B_\varepsilon(y)$.
    Since $y\in Z$ and $X$ is dense in $Z$, we have
  $B_\delta(y)\cap X\neq\emptyset$.
    Thus $B_\varepsilon(y)\cap X\cap Y\supseteq B_\delta(y)\cap X\cap Y\neq \emptyset$.
\end{proof}

\begin{definition}
    A path in $\mathbb{R}^n$ is a set $X\subseteq \mathbb{R}^n$ such that there exists a continuous surjective function $f:[0,1]\to X$.
\end{definition}
In the rest of the proof, we will consider the class \Tmc of all
graphs that take the form of a directed tree of depth two,
with vertex labels from $\Pi = \{P\}$, and where $P$ may only
label leaves.

\begin{theorem}\label{lemma: two-layer-cmgnn-afrml-inexpressiveness}
    There is no two-layer \contresmeanGNN that classifies the roots of trees in $\mathcal{T}$ according to the \ac{rml} formula $\varphi = \markDiamond^{>\frac{1}{2}}\markDiamond^{>\frac{1}{2}}P$.
  \end{theorem}

  \noindent
\begin{proof}\ 
  Let $$\mathcal{G}=(L,\{\AGG^{\ell}\}_{\ell \in [2]},\{\COM^{\ell}\}_{\ell \in [2]},\CLS)$$
be a two-layer \contresmeanGNN with input dimensions
$\delta^0=1$ and $\delta^1$. When \Gmc is run on  a graph $G\in \mathcal{T}$,
then the feature vectors computed at a leaf $v \in V(G)$, by layer~1 and~2, only
depend on whether $v$ is labeled with $P$ or not, but not on any
other aspects of $G$. Moreover, the feature vector computed by the
first layer at the root is entirely independent of $G$. 
Let $\bar{x}^1$ be the feature vector computed by the first layer at
the root, $\bar{y}^0$ the initial feature vector of the intermediate
vertices, and $\bar{z}_+^\ell$ and $\bar{z}_-^\ell$ the feature
vectors of the leaves that are and are not labeled with $P$,  respectively.


For an intermediate vertex with a fraction of  $r\in[0,1]\cap\mathbb{Q}$
successors labeled $P$, the vector computed in layer~1 is
$$\bar{y}^1_r = \COM^1(\bar{y}^0, r\bar{z}_+^0+(1-r)\bar{z}_-^0).$$
We define the set of all such vectors by $Y^1$.
This set is dense in $Y^1_\mathbb{R} = \{\COM^1(\bar{y}^0, r\bar{z}_+^0 + (1-r)\bar{z}_-^0\mid r\in [0,1])\}$.
As an intuition, the set
$$\{(\bar{y}^0, r\bar{z}_+^0+(1-r)\bar{z}_-^0)\mid r\in [0,1]\}$$
forms a line segment in $\mathbb{R}^{2\delta^0}$ and $Y^1_\mathbb{R}$
forms a path in $\mathbb{R}^{\delta^1}$.  
The set $Y^1$ can be partitioned into two sets $Y^1_+ =
\{\bar{y}_r^1\mid r\in (\frac{1}{2},1]\cap \mathbb{Q}\}$ and $Y^1_- = \{\bar{y}_r^1\mid
r\in [0,\frac{1}{2}]\cap \mathbb{Q}\}.$
We can split $Y^1_\mathbb{R}$ analogously into $Y^1_{+,\mathbb{R}}$ and $Y^1_{-,\mathbb{R}}$.

For each root vertex $v$ with
$$s = \frac{|\{u\in\mathcal{N}(v)\mid u\models
  \markDiamond^{>\frac{1}{2}}P\}|}{|\mathcal{N}(v)|},$$
the $\MEAN$ computed in the second layer can be reformulated as follows:
\begin{align*}
    &\MEAN\{\!\{\bar{y}_u^1\mid u\in\mathcal{N}(v)\}\!\} \\
    &= s\cdot\MEAN\{\!\{\bar{y}_u^1\mid u\in\mathcal{N}(v),u\models\markDiamond^{>\frac{1}{2}}P\}\!\}\\
    &\phantom{=}+ (1-s)\cdot\MEAN\{\!\{\bar{y}_u^1\mid u\in\mathcal{N}(v),u\not\models\markDiamond^{>\frac{1}{2}}P\}\!\}.
\end{align*}
Recall that to classify the trees in \Tmc according to the \ac{rml}
formula $\markDiamond^{>\frac{1}{2}}\markDiamond^{>\frac{1}{2}}P$, the \ac{gnn} has to decide whether $s>\frac{1}{2}$.
Clearly, the first $\MEAN$ in the above sum processes a multiset in $\mathcal{M}(Y_+^1)$, while the second $\MEAN$ processes a multiset in $\mathcal{M}(Y_-^1)$.
We define

\begin{alignat*}{2}
M_+ = \{s\bar{y}_++(1-s)\bar{y}_-\mid\ & s\in (\frac{1}{2},1]\cap \mathbb{Q},&\\
      &\bar{y}_+\in \MEAN(\mathcal{M}(Y_+^1)),&\\
 &\bar{y}_-\in\MEAN(\mathcal{M}(Y_-^1))\}&\\
  M_{+,\mathbb{R}} = \{s\bar{y}_++(1-s)\bar{y}_-\mid\ & s\in (\frac{1}{2},1],\qquad\qquad&\\
   \bar{y}_+\in\cvh(&Y^1_{+,\mathbb{R}}),\ \bar{y}_-\in\cvh(Y^1_{-,\mathbb{R}})\}.&
\end{alignat*}

We define $M_-$ and $M_{-,\mathbb{R}}$ analogously, but require that $s\in [0,\frac{1}{2}]\cap \mathbb{Q}$ and $s\in [0,\frac{1}{2}]$ respectively.
Now for each $\bar{m}_+\in M_{+}$, the vector $\COM^2(\bar{x}^1, \bar{m}_+)$ is a feature vector computed at the root of a tree in $\mathcal{T}$ that satisfies $\varphi$.
And likewise, for each $\bar{m}_-\in M_{-}$, the vector $\COM^2(\bar{x}^1, \bar{m}_-)$ belongs to a tree in $\mathcal{T}$ that does not satisfy $\varphi$.

By Lemma~\ref{lemma:mean-cvh-dense-preserving}, $\MEAN(\mathcal{M}(Y_+^1))$ is dense in $\cvh(Y_{+,\mathbb{R}}^1)$ and $\MEAN(\mathcal{M}(Y_-^1))$ is dense in $\cvh(Y_{-,\mathbb{R}}^1)$.
Therefore, $M_+$ is dense in $M_{+,\mathbb{R}}$. Analogously $M_-$ is dense in $M_{-,\mathbb{R}}$.
We now show that $M_{+,\mathbb{R}}$ and $M_{-,\mathbb{R}}$ have a non-empty
intersection. To be more precise, we show there exists a set in $M_{+,\mathbb{R}}$ that is open in $\cvh(Y_{+,\mathbb{R}}^1)$ that is also contained in $M_{-,\mathbb{R}}$.
Let $\bar{y}\in \ri(\cvh(Y_{+,\mathbb{R}}^1))$ and let $\varepsilon>0$ such that
 \begin{align*}
    B_{\mathbb{R}}=\{y' \mid y'\in \aff(\cvh(Y_{+,\mathbb{R}}^1)), ||y'-y||_\infty < \varepsilon\}\\\subseteq\ri(\cvh(Y^1_{+,\mathbb{R}})).
 \end{align*}
Consider the point $$\bar{y}^1_{0.5} = \COM^1(\bar{y}^0, 0.5\bar{z}_+^0+0.5\bar{z}_-^0)\in Y_-^1.$$
Because $\COM^1$ is continuous, we can approximate $\bar{y}^1_{0.5}$ by vectors in $Y_{+,\mathbb{R}}^1$, hence $\bar{y}^1_{0.5}\in \clo(\cvh(Y_{+,\mathbb{R}}^1))$.
By the accessibility lemma, $$B_{?,\mathbb{R}} = \frac{1}{4}B_{\mathbb{R}} + \frac{3}{4}\bar{y}^1_{0.5}\subseteq\ri(\cvh(Y_{+,\mathbb{R}}^1)).$$
Since $\ri(\cvh(Y_{+,\mathbb{R}}^1))\subseteq M_{+,\mathbb{R}}$, it follows that $B_{?,\mathbb{R}}\subseteq M_{+,\mathbb{R}}$.
But we also have $B_{?,\mathbb{R}}\subseteq M_{-,\mathbb{R}}$ since for each vector $\bar{v} \in B_{?,\mathbb{R}}$ we can find a $\bar{y}_+\in \cvh(Y_{+,\mathbb{R}}^1)$ and can choose $\bar{y}_- = \bar{y}_{0.5}^1\in \cvh(Y_{-,\mathbb{R}}^1)$ and $s=\frac{1}{4}\in [0,\frac{1}{2}]$ such that $\bar{v} = s\bar{y}_+ + (1-s)\bar{y}_-$.

\smallskip
It remains to argue that $B_{?,+} = M_+\cap B_{?,\mathbb{R}} $ and $B_{?,-} = M_- \cap B_{?,\mathbb{R}}$ are dense in $B_{?, \mathbb{R}}$.

We show both statements by using Lemma~\ref{lemma: dense-sets-open-set-intersection} by considering these sets relative to $\cvh(Y_{+,\mathbb{R}}^1)$.
For the first statement, consider \(X = M_+\cap \cvh(Y_{+,\mathbb{R}}^1)\), \(Y = B_{?,\mathbb{R}}\), \(Z = \cvh(Y_{+,\mathbb{R}}^1)\).
The set $B_{?,\mathbb{R}}$ is open in $\cvh(Y_{+,\mathbb{R}})$ since $B_{\mathbb{R}}$ is open in $\aff(\cvh(Y_{+,\mathbb{R}}^1))$, $\bar{y}_{0.5}^1\in\aff(\cvh(Y_{+,\mathbb{R}}^1))$ and $B_{?,\mathbb{R}}$ is just a translation and scaling within $\aff(\cvh(Y_{+,\mathbb{R}}^1))$.
Thus, $B_{?,\mathbb{R}}$ is open in $\aff(\cvh(Y_{+,\mathbb{R}}^1))$ and thus also in $\cvh(Y_{+,\mathbb{R}}^1)$.
Since $\MEAN(\mathcal{M}(Y_+^1))\subseteq M_+\cap \cvh(Y_{+,\mathbb{R}}^1)$ and \(\MEAN(\mathcal{M}(Y_+^1))\) is dense in \(\cvh(Y_{+,\mathbb{R}}^1)\), it follows that $M_+\cap \cvh(Y_{+,\mathbb{R}}^1)$ is dense in $\cvh(Y_{+,\mathbb{R}}^1)$.
We can rewrite $B_{?,+}$ as $(M_+\cap \cvh(Y_{+,\mathbb{R}}^1))\cap B_{?,\mathbb{R}}$, since $B_{?,\mathbb{R}}\subseteq \cvh(Y_{+,\mathbb{R}}^1)$. 

Before showing that $B_{?,-}$ is dense in $B_{?,\mathbb{R}}$ we first consider the sets \({X = \MEAN(\mathcal{M}(Y_+^1))}\), \(Y = B_{\mathbb{R}}\), as well as \(Z = \cvh(Y_{+,\mathbb{R}})\).
Using Lemma~\ref{lemma: dense-sets-open-set-intersection}, we conclude that \(B = B_{\mathbb{R}}\cap \MEAN(\mathcal{M}(Y_+^1))\) is dense in \(B_\mathbb{R}\).
We now transform these sets according to the definition of \(B_{?,\mathbb{R}}\):
\(X = \frac{1}{4}\MEAN(\mathcal{M}(Y_+^1))+\frac{3}{4}\bar{y}_{0.5}^1\), \({Y = B_{?, \mathbb{R}}}\), and \({Z = \frac{1}{4}\cvh(Y_{+,\mathbb{R}})+\frac{3}{4}\bar{y}_{0.5}^1}\).
\(X\) remains dense and \(Y\) remains open in \(Z\), therefore \({B_{?,\mathbb{R}}\cap (\frac{1}{4}\MEAN(\mathcal{M}(Y_{+}^1))+\frac{3}{4}\bar{y}_{0.5}^1)}\) is dense in \(B_{?, \mathbb{R}}\).
The desired statement now follows from \({\frac{1}{4}\MEAN(\mathcal{M}(Y_{+}^1))+\frac{3}{4}\bar{y}_{0.5}^1\subseteq M_-}\).

We can now finish our proof that \(\mathcal{G}\) has to misclassify some graphs in \(\mathcal{T}\).
W.l.o.g we can assume that $\COM^2$ has output dimension $1$ since $\CLS(\bar{x})$ only depends on one entry in $\bar{x}$. Since projections are continuous, this does not affect the continuity of $\COM^2$.
By Lemma~\ref{lemma: continuous-preserve-dense-sets} and since $\COM^2$ is continuous, $\COM^2(\bar{x}^1, B_{?,+})$ and $\COM^2(\bar{x}^1, B_{?,-})$ are dense in $\COM^2(\bar{x}^1, B_{?,\mathbb{R}})$ and the step function $\CLS$ cannot separate these two dense sets.
\end{proof}

We still have to show that the restriction to two-layer GNNs does not
change the expressiveness on
the class of graphs~\Tmc.


\begin{theorem}\label{theorem: rml_not_expr_by_meanc}
The \ac{rml} formula $\markDiamond^{>\frac{1}{2}}\markDiamond^{>\frac{1}{2}}P$  is not expressible by a \contresmeanGNN.
\end{theorem}

\noindent
\begin{proof}\ 
We show that for each \contresmeanGNN $\mathcal{G}$ with $L$ layers
there exists a \ac{gnn} with two layers that classifies the root of each tree in $\mathcal{T}$ identically.
The construction preserves the continuity of the combination functions.

Let $\bar{a}_+^0,\ldots,\bar{a}_+^L$ and
$\bar{a}_-^0,\ldots,\bar{a}_-^L$ be the feature vectors assigned by
$\mathcal{G}$ to a vertex without successors that does and does not satisfy $P$, respectively.
Notice that all initial feature vectors are either $\bar{a}_+^0$ or $\bar{a}_-^0$.

The first layer of the two-layer \ac{gnn} works as follows: 
${\COM'}^1(x,y)$ is the concatenation of the vectors
\begin{align*}
    \bar{v}_1 &= \COM^1(\bar{a}_-^0, y\bar{a}_+^0+(1-y)\bar{a}_-^0)\\
    \bar{v}_i &= \COM^i(\bar{v}_{i-1}, y\bar{a}_+^{i-1}+(1-y)\bar{a}_-^{i-1})
\end{align*}
for all $1 < i\leq L$.
Intuitively, $y$ is the fraction of successors satisfying $P$ and ${\COM'}^1$ computes all feature vectors of a vertex at height one that does not satisfy $P$.
Since all $\COM^i$ are continuous, ${\COM'}^1$ is again continuous because continuous functions are closed under addition, function composition and vector concatenation.

The second layer is defined similarly: ${\COM'}^2((\bar{x}_1,\ldots,\bar{x}_L), (\bar{y}_1,\ldots,\bar{y}_L))$  returns the vector $\bar{u}_L$, where
\begin{align*}
    \bar{u}_1 &= \COM^1(\bar{a}_-^0, \bar{a}_-^0)\\
    \bar{u}_i &= \COM^i(\bar{u}_{i-1}, \bar{y}_{i-1})
\end{align*}
Intuitively, these vectors are the feature vectors of a vertex of height two in a tree where the vertex itself and all immediate successors do not satisfy $P$.
Therefore, this \ac{gnn} computes the same feature vector as \(\mathcal{G}\) at the root of each tree in $\mathcal{T}$.
Analogously to ${\COM'}^1$, because all $\COM^i$ are continuous, ${\COM'}^2$ is continuous.

By using the same classification function as $\mathcal{G}$, the
constructed two-layer \ac{gnn} classifies trees in $\mathcal{T}$ in
the same way as $\mathcal{G}$.
It thus follows from Theorem~\ref{lemma:
  two-layer-cmgnn-afrml-inexpressiveness} that $\mathcal{G}$ does not express $\markDiamond^{>\frac{1}{2}}\markDiamond^{>\frac{1}{2}}P$.%
\end{proof}

\end{document}